\documentclass[pdflatex,sn-mathphys-num]{sn-jnl}

\usepackage[edges]{forest}
\usepackage{lmodern}
\usepackage{bbding} 
\usepackage{pifont}
\usepackage{adjustbox}
\usepackage{lineno} 
\definecolor{lightcoral}{rgb}{0.94, 0.5, 0.5}
\definecolor{lightgreen}{rgb}{0.56, 0.93, 0.56}
\definecolor{harvestgold}{rgb}{0.98, 0.85, 0.40}
\definecolor{brightlavender}{rgb}{0.75, 0.58, 0.89}
\definecolor{capri}{rgb}{0.0, 0.75, 1.0}
\definecolor{carminepink}{rgb}{0.92, 0.3, 0.56}
\definecolor{celadon}{rgb}{0.67, 0.88, 0.69}
\definecolor{darkpastelgreen}{rgb}{0.01, 0.75, 0.24}

\definecolor{hidden-draw}{RGB}{205, 44, 36}
\definecolor{hidden-blue}{RGB}{194,232,247}
\definecolor{hidden-orange}{RGB}{243,202,120}
\definecolor{hidden-yellow}{RGB}{242,244,193}
\definecolor{tree-level-1}{RGB}{245,20,85}
\definecolor{tree-level-2}{RGB}{246,86,118}
\definecolor{tree-level-3}{RGB}{248,177,193}
\definecolor{tree-leaf}{RGB}{176,230,198}

\definecolor{Self}{RGB}{255,0,128}
\definecolor{Ensemble}{RGB}{0,127,255}
\definecolor{Iterative}{RGB}{153,51,255}

\definecolor{exemplar1}{RGB}{136,98,148}
\definecolor{exemplar2}{RGB}{148,210,242}
\definecolor{knowledge1}{RGB}{249,219,152}
\definecolor{knowledge2}{RGB}{255,245,220}
\usepackage{tikz}
\usetikzlibrary{shapes.geometric, arrows.meta, positioning, fit}

\usepackage{booktabs}
\usepackage{multirow}
\usepackage{caption}
\usepackage{array}
\usepackage{adjustbox}
\usepackage{graphicx}%
\usepackage{multirow}%
\usepackage{amsmath,amssymb,amsfonts}%
\usepackage{amsthm}%
\usepackage[title]{appendix}%
\usepackage{xcolor}%
\usepackage{textcomp}%
\usepackage{manyfoot}%
\usepackage{booktabs}%
\usepackage{algorithm}%
\usepackage{algorithmicx}%
\usepackage{algpseudocode}%
\usepackage{listings}%

\usepackage{capt-of}

\begin{document}

\title[Article Title]{Decoding Linguistic Representations of Human Brain}


\author[1,2]{\fnm{Yu} \sur{Wang}}\email{yuwangsjtu@sjtu.edu.cn}
\equalcont{These authors contributed equally to this work.}

\author[1]{\fnm{Heyang} \sur{Liu}}\email{liuheyang@sjtu.edu.cn}
\equalcont{These authors contributed equally to this work.}

\author[1]{\fnm{Yuhao} \sur{Wang}}\email{colane@sjtu.edu.cn}

\author[1]{\fnm{Chuan} \sur{Xuan}}\email{handmasterxuan@gmail.com}

\author[1]{\fnm{Yixuan} \sur{Hou}}\email{yixuanhou.sjtu@gmail.com}

\author[1]{\fnm{Sheng} \sur{Feng}}\email{fs0015@sjtu.edu.cn}

\author[1]{\fnm{Hongcheng} \sur{Liu}}\email{hongcheng\_liu@sjtu.edu.cn}

\author[1,2]{\fnm{Yusheng} \sur{Liao}}\email{liao20160907@sjtu.edu.cn}

\author*[1,2]{\fnm{Yanfeng} \sur{Wang}}\email{wangyanfeng622@sjtu.edu.cn}

\affil[1]{\orgname{Shanghai Jiao Tong
University}, \orgaddress{\country{China}}}

\affil[2]{\orgdiv{Shanghai Artificial Intelligence Laboratory}, \orgaddress{\country{China}}}


\abstract{Language, as an information medium created by advanced organisms, has always been a concern of neuroscience regarding how it is represented in the brain. Decoding linguistic representations in the evoked brain has shown groundbreaking achievements, thanks to the rapid improvement of neuroimaging, medical technology, life sciences and artificial intelligence. In this work, we present a taxonomy of brain-to-language decoding of both textual and speech formats. This work integrates two types of research: neuroscience focusing on language understanding and deep learning-based brain decoding. Generating discernible language information from brain activity could not only help those with limited articulation, especially amyotrophic lateral sclerosis (ALS) patients but also open up a new way for the next generation's brain-computer interface (BCI). This article will help brain scientists and deep-learning researchers to gain a bird's eye view of fine-grained language perception, and thus facilitate their further investigation and research of neural process and language decoding.}




\maketitle

\tikzstyle{my-box}=[
    rectangle,
    draw=hidden-draw,
    rounded corners,
    text opacity=1,
    minimum height=1.5em,
    minimum width=5em,
    inner sep=2pt,
    align=center,
    fill opacity=.5,
]
\tikzstyle{cause_leaf}=[my-box, minimum height=1.5em,
    fill=harvestgold!20, text=black, align=left,font=\scriptsize,
    inner xsep=2pt,
    inner ysep=4pt,
]

\tikzstyle{class_leaf}=[my-box, minimum height=1.5em,
    fill=cyan!20, text=black, align=left,font=\scriptsize,
    inner xsep=2pt,
    inner ysep=4pt,
]

\tikzstyle{detect_leaf}=[my-box, minimum height=1.5em,
    fill=carminepink!20, text=black, align=left,font=\scriptsize,
    inner xsep=2pt,
    inner ysep=4pt,
]
\tikzstyle{mitigate_leaf}=[my-box, minimum height=1.5em,
    fill=lightgreen!20, text=black, align=left,font=\scriptsize,
    inner xsep=2pt,
    inner ysep=4pt,
]
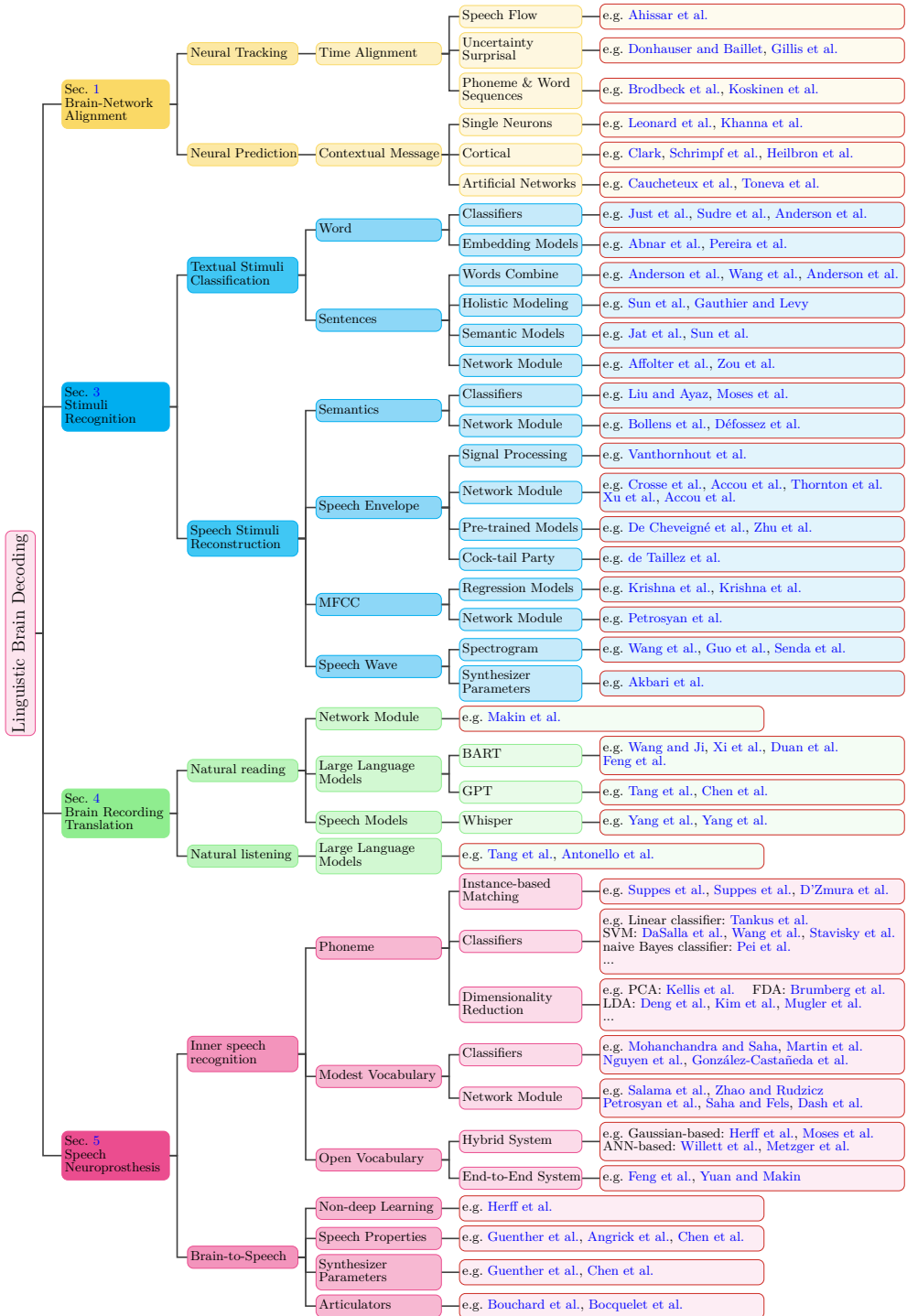
\begin{figure*}[tp]
    \centering
    \resizebox{\textwidth}{!}{
        \begin{forest}
            forked edges,
            for tree={
                grow=east,
                reversed=true,
                anchor=base west,
                parent anchor=east,
                child anchor=west,
                base=left,
                font=\small,
                rectangle,
                draw=hidden-draw,
                rounded corners,
                align=left,
                minimum width=4em,
                edge+={darkgray, line width=1pt},
                s sep=3pt,
                inner xsep=2pt,
                inner ysep=3pt,
                ver/.style={rotate=90, child anchor=north, parent anchor=south, anchor=center,font=\large},
            },
            where level=1{text width=6.8em,font=\scriptsize,}{},
            where level=2{text width=7.0em,font=\scriptsize,}{},
            where level=3{text width=8.0em,font=\scriptsize,}{},
            where level=4{text width=7.8em,font=\scriptsize,}{},
            [
                Linguistic Brain Decoding, ver, color=carminepink!100, fill=carminepink!15,
                text=black
                [
                    Sec.~\ref{sec1} \\Brain-Network \\ Alignment , color=harvestgold!100, fill=harvestgold!100, text=black
                    [
                        Neural Tracking, color=harvestgold!100, fill=harvestgold!60,  text=black
                        [
                            Time Alignment, color=harvestgold!100, fill=harvestgold!40, text=black
                            	[
                            		Speech Flow, color=harvestgold!100, fill=harvestgold!20, text=black
                            		[
                                		{e.g. \citeauthor{ahissar2001speech}}
                                		, cause_leaf, text width=20em
                            		]
                            ]
                            [
                            		Uncertainty \\ Surprisal, color=harvestgold!100, fill=harvestgold!20, text=black
                            		[
                                		{e.g. \citeauthor{donhauser2020two,gillis2021neural}}
                                		, cause_leaf, text width=20em
                            		]
                            ] 
                            [
                            		Phoneme \& Word \\ Sequences, color=harvestgold!100, fill=harvestgold!20, text=black
                            		[
                                		{e.g. \citeauthor{brodbeck2018rapid, koskinen2020brain}}
                                		, cause_leaf, text width=20em
                            		]
                            ]
                        ]
                    ]
                    [
                        Neural Prediction, color=harvestgold!100, fill=harvestgold!60, text=black
                        [
                            Contextual Message, color=harvestgold!100, fill=harvestgold!40, text=black
                            [
                            		Single Neurons, color=harvestgold!100, fill=harvestgold!20, text=black
                            		[
                                		{e.g. \citeauthor{leonard2024large, khanna2024single}}
                                		, cause_leaf, text width=20em
                            		]
                            ]
                            [
                            		Cortical, color=harvestgold!100, fill=harvestgold!20, text=black
                            		[
                                		{e.g. \citeauthor{clark2013whatever, schrimpf2021neural,heilbron2022hierarchy}}
                                		, cause_leaf, text width=20em
                            		]
                            ]
                            [
                            		Artificial Networks, color=harvestgold!100, fill=harvestgold!20, text=black
                            		[
                                		{e.g. \citeauthor{ caucheteux2021disentangling, toneva2022combining}}
                                		, cause_leaf, text width=20em
                            		]
                            ]
                        ]
                    ]
                ]
                [
                    Sec.~\ref{sec3} \\ Stimuli \\ Recognition , color=cyan!100, fill=cyan!100, text=black
                    [
                        Textual Stimuli \\ Classification, color=cyan!100, fill=cyan!60,  text=black
                        [
                           Word, color=cyan!100, fill=cyan!40, text=black
                            	[
                            		Classifiers, color=cyan!100, fill=cyan!20, text=black
                            		[
                                		{e.g. \citeauthor{just2010neurosemantic, sudre2012tracking, anderson2017visually}}
                                		, class_leaf, text width=20em
                            		]
                            ]
                            [
                            		Embedding Models, color=cyan!100, fill=cyan!20, text=black
                            		[
                                		{ e.g. \citeauthor{abnar2018experiential,pereira2018toward}}
                                		, class_leaf, text width=20em
                            		]
                            ] 
                        ]
                        [
                           Sentences, color=cyan!100, fill=cyan!40, text=black
                            	[
                            		Words Combine, color=cyan!100, fill=cyan!20, text=black
                            		[
                                		{e.g. \citeauthor{anderson2017predicting, wang2017predicting, anderson2019integrated}}
                                		, class_leaf, text width=20em
                            		]
                            ]
                            [
                            		Holistic Modeling, color=cyan!100, fill=cyan!20, text=black
                            		[
                                		{e.g. \citeauthor{sun2019towards, gauthier2019linking}}
                                		, class_leaf, text width=20em
                            		]
                            ] 
                            [
                            		Semantic Models, color=cyan!100, fill=cyan!20, text=black
                            		[
                                		{e.g. \citeauthor{jat2019relating,sun2020neural}}
                                		, class_leaf, text width=20em
                            		]
                            ]
                            [
                            		Network Module, color=cyan!100, fill=cyan!20, text=black
                            		[
                                		{e.g. \citeauthor{affolter2020brain2word, zou2021towards}}
                                		, class_leaf, text width=20em
                            		]
                            ]
                        ]
                    ]
                    [
                        Speech Stimuli \\ Reconstruction, color=cyan!100, fill=cyan!60, text=black
                        [
                            Semantics, color=cyan!100, fill=cyan!40, text=black
                            [
                            		Classifiers, color=cyan!100, fill=cyan!20, text=black
                            		[
                                		{e.g. \citeauthor{liu2018speech, moses2019real}}
                                		, class_leaf, text width=20em
                            		]
                            ]
                            [
                            		Network Module, color=cyan!100, fill=cyan!20, text=black
                            		[
                                		{e.g. \citeauthor{bollens2022learning,defossez2023decoding}}
                                		, class_leaf, text width=20em
                            		]
                            ]
                        ]
                        [
                            Speech Envelope, color=cyan!100, fill=cyan!40, text=black
                            [
                            		Signal Processing, color=cyan!100, fill=cyan!20, text=black
                            		[
                                		{e.g. \citeauthor{vanthornhout2018speech}}
                                		, class_leaf, text width=20em
                            		]
                            ]
                            [
                            		Network Module, color=cyan!100, fill=cyan!20, text=black
                            		[
                                		{e.g. \citeauthor{crosse2016multivariate, accou2021modeling,thornton2022robust}\\ \citeauthor{xu2022decoding,accou2023decoding}}
                                		, class_leaf, text width=20em
                            		]
                            ]
                            [
                            		Pre-trained Models, color=cyan!100, fill=cyan!20, text=black
                            		[
                                		{e.g. \citeauthor{de2018decoding, zhu2023eeg2vec}}
                                		, class_leaf, text width=20em
                            		]
                            ]
                            [
                            		Cock-tail Party, color=cyan!100, fill=cyan!20, text=black
                            		[
                                		{e.g. \citeauthor{de2020machine}}
                                		, class_leaf, text width=20em
                            		]
                            ]
                        ]
                        [
                            MFCC, color=cyan!100, fill=cyan!40, text=black
                            [
                            		Regression Models, color=cyan!100, fill=cyan!20, text=black
                            		[
                                		{e.g. \citeauthor{krishna2019state, krishna2021advancing}}
                                		, class_leaf, text width=20em
                            		]
                            ]
                            [
                            		Network Module, color=cyan!100, fill=cyan!20, text=black
                            		[
                                		{e.g. \citeauthor{petrosyan2021compact}}
                                		, class_leaf, text width=20em
                            		]
                            ]
                        ]
                        [
                            Speech Wave, color=cyan!100, fill=cyan!40, text=black
                            [
                            	Spectrogram, color=cyan!100, fill=cyan!20, text=black
                            		[
                                		{e.g. \citeauthor{wang2020stimulus,guo2023end,senda2024auditory}}
                                		, class_leaf, text width=20em
                            		]
                            ]
                            [
                            		Synthesizer \\ Parameters, color=cyan!100, fill=cyan!20, text=black
                            		[
                                		{e.g. \citeauthor{akbari2019towards}}
                                		, class_leaf, text width=20em
                            		]
                            ]
                        ]
                    ]
                ]
                [
                        Sec.~\ref{sec4} \\Brain Recording \\ Translation, color=lightgreen!100, fill=lightgreen!100, text=black
                        [
                        Natural reading, color=lightgreen!100, fill=lightgreen!60,  text=black
                        [
                            Network Module, color=lightgreen!100, fill=lightgreen!40, text=black
                            [
                                		{e.g. \citeauthor{makin2020machine}}
                                , mitigate_leaf, text width=20em
                            ]
                        ]
                        [
                            Large Language \\ Models, color=lightgreen!100, fill=lightgreen!40, text=black
                            [
                            	BART, color=lightgreen!100, fill=lightgreen!20, text=black
                            		[
                                		{e.g. \citeauthor{wang2022open, xi2023unicorn,duan2024dewave}\\ \citeauthor{feng2023aligning}}
                                , mitigate_leaf, text width=20em
                            		]
                            ]
                            [
                            		GPT, color=lightgreen!100, fill=lightgreen!20, text=black
                            		[
                                		{e.g. \citeauthor{tang2023semantic, chen2024openvocabulary}}
                                , mitigate_leaf, text width=20em
                            		]
                            ]
                        ]
                        [
                            Speech Models, color=lightgreen!100, fill=lightgreen!40, text=black
                            [
                                Whisper, color=lightgreen!100, fill=lightgreen!20, text=black
                                [
                                		{e.g. \citeauthor{yang2024decode,yang2024mad}}
                                , mitigate_leaf, text width=20em
                                ]
                            ]
                        ]
                        ]
                        [
                            Natural listening, color=lightgreen!100, fill=lightgreen!60, text=black
                            [
                            	Large Language \\ Models, color=lightgreen!100, fill=lightgreen!40, text=black
                            		[
                                		{e.g. \citeauthor{tang2023semantic, antonello2024many}}
                                , mitigate_leaf, text width=20em
                            		]
                            ]
                        ]
                    ]
                    [
                        Sec.~\ref{sec5} \\Speech \\ Neuroprosthesis, color=carminepink!100, fill=carminepink!100, text=black
                        [
                        Inner speech \\ recognition, color=carminepink!100, fill=carminepink!60,  text=black
                           [
                               Phoneme, color=carminepink!100, fill=carminepink!40, text=black
                            [
                                Instance-based \\ Matching, color=carminepink!100, fill=carminepink!20, text=black
                                [
                                    {e.g. \citeauthor{suppes1997brain, suppes1998brain, d2009toward}}
                                    , detect_leaf, text width=20em
                                ]
                            ]
                            [
                                Classifiers, color=carminepink!100, fill=carminepink!20, text=black
                                [
                                    {e.g. Linear classifier: \citeauthor{tankus2012structured} \\
                                    SVM: \citeauthor{dasalla2009single, wang2013analysis, stavisky2018decoding} \\
                                    naive Bayes classifier: \citeauthor{pei2011decoding} \\...}
                                    , detect_leaf, text width=20em
                                ]
                            ]
                            [
                               Dimensionality \\ Reduction, color=carminepink!100, fill=carminepink!20, text=black
                                [
                                    {e.g. PCA: \citeauthor{kellis2010decoding} \hspace{2mm} FDA: \citeauthor{brumberg2011classification} \\
                                    LDA: \citeauthor{deng2010eeg, kim2014eeg,mugler2014direct} \\...}
                                    , detect_leaf, text width=20em
                                ]
                            ]
                        ]
                        [
                           Modest Vocabulary, color=carminepink!100, fill=carminepink!40, text=black
                            [
                                Classifiers, color=carminepink!100, fill=carminepink!20, text=black
                                [
                                    {e.g.  \citeauthor{mohanchandra2016communication,martin2016word}\\ \citeauthor{nguyen2017inferring,gonzalez2017sonification}}
                                    , detect_leaf, text width=20em
                                ]
                            ]
                            [
                                Network Module, color=carminepink!100, fill=carminepink!20, text=black
                                [
                                    {e.g. \citeauthor{salama2014recognition,zhao2015classifying}\\
                                    \citeauthor{saha2019hierarchical,dash2020decoding,petrosyan2021compact}}
                                    , detect_leaf, text width=20em
                                ]
                            ]
                        ]
                        [
                           Open Vocabulary, color=carminepink!100, fill=carminepink!40, text=black
                            [
                                Hybrid System, color=carminepink!100, fill=carminepink!20, text=black
                                [
                                    {e.g. Gaussian-based: \citeauthor{herff2015brain,moses2016neural}\\
                                    ANN-based: \citeauthor{willett2023high, metzger2023high}}
                                    , detect_leaf, text width=20em
                                ]
                            ]
                            [
                                End-to-End System, color=carminepink!100, fill=carminepink!20, text=black
                                [
                                    {e.g. \citeauthor{feng2024towards,yuan2024improving}}
                                    , detect_leaf, text width=20em
                                ]
                            ]
                        ]
                        ]
                        [
                           Brain-to-Speech, color=carminepink!100, fill=carminepink!60, text=black
                            [
                               Non-deep Learning, color=carminepink!100, fill=carminepink!40, text=black
                                [
                                    {e.g. \citeauthor{herff2019generating}}
                                    , detect_leaf, text width=20em
                                ]
                            ]
                            [
                                Speech Properties, color=carminepink!100, fill=carminepink!40, text=black
                                [
                                    {e.g. \citeauthor{guenther2009wireless, angrick2019speech, chen2024neural}}
                                    , detect_leaf, text width=20em
                                ]
                            ]
                            [
                               Synthesizer \\ Parameters, color=carminepink!100, fill=carminepink!40, text=black
                                [
                                    {e.g. \citeauthor{guenther2009wireless, chen2024neural}}
                                    , detect_leaf, text width=20em
                                ]
                            ]
                            [
                               Articulators, color=carminepink!100, fill=carminepink!40, text=black
                                [
                                    {e.g. \citeauthor{bouchard2013functional,bocquelet2014robust}}
                                    , detect_leaf, text width=20em
                                ]
                            ]
                    ]
                ]
           ]
        \end{forest}
    }
    \caption{The main content flow and categorization of this survey.}
    \label{categorization_of_survey}
\end{figure*}
Language is a unique mode of communication exclusive to higher-order organisms, distinguishing them from other species. The human brain houses the nervous center for processing language information. Despite advancing insights into the human brain, the intricacies of linguistic representation and processing still largely resemble a black-box model, making it challenging to explain intuitively the internal mechanisms and patterns of information transfer \cite{abnar2019blackbox}. The language communication requires the involvement of a complete pronunciation loop, which is unavailable for patients with movement disorders, especially those with amyotrophic lateral sclerosis (ALS) \cite{goutman2023amyotrophic}. Deep learning presents a viable solution for understanding language in the brain by utilizing large-scale trainable parameters to map the correlation between external stimuli and neural activity. This paper summarizes a single direction of brain linguistic perception - from the evoked brain response to its language stimuli or intention, often called brain decoding. Progress in this field involves the joint efforts of neuroscientists and artificial intelligence (AI) researchers. We introduce the neurological foundations of brain decoding using deep networks and illustrate multiple model architectures for generating linguistic stimuli or imagined speech. We classify task forms into multiple standardized paradigms, facilitating researchers to further progress their work, and conclude by discussing the challenges faced by related fields and proposing directions for an ideal brain-computer interface (BCI).


Figure~\ref{categorization_of_survey} illustrates the main content of this survey. We begin by discussing the neurological basis of linguistic decoding in the human brain. Neural tracking ensures the temporal alignment of brain responses with linguistic properties, while continuous neural prediction supports the integration of contextual information. Neural prediction is observed at both the single-neuron and population levels, and its corresponding context has proven essential in artificial neural networks (ANNs). Stimuli recognition is the simplest form of brain decoding, involving the differentiation of linguistic stimuli by analyzing the subject's evoked brain responses. For text stimuli reconstruction, decoding is performed at the word or sentence level using classifiers, embedding models, and custom network modules to map the relationship between brain wave patterns and linguistic representations. Considering the dynamics of speech flow, restoring related semantics, the speech envelope, mel-frequency cepstral coefficients (MFCC), and speech waves present broader challenges. Brain recording translation paradigms are applied in natural reading or listening scenarios, where the decoding system generates the stimulus sequence in textual or speech form based on the evoked brain response. This task is analogous to machine translation, treating brain activity as the source language and translating it into human-understandable text. Speech neuroprosthesis focuses on decoding inner or vocal speech based on human intentions. The field has progressed from phoneme-level recognition to open-vocabulary sentence decoding. Brain-to-speech technology is a promising direction, with spectrograms generated through matching algorithms or by considering speech properties, synthesizer parameters, and articulator movements. Additionally, to assist neuroscientists and artificial intelligence researchers in better developing decoding systems, we provide evaluation metrics introduced from deep learning tasks before the introduction of brain decoding solutions (Section~\ref{sec2}) and summarize the representative datasets used for sequence decoding in the appendix (Appendix~\ref{secA1}).


\section{Brain-Network Alignment}\label{sec1}

Brain signal recordings measure and quantify the instinct biometric neural response from the human brain, whether passively or actively. Brain signals can be divided into two categories: invasive and non-invasive. The former, including functional magnetic resonance imaging (fMRI), electroencephalography (EEG), magnetoencephalography (MEG), etc., are affected by transcranial attenuation and have a lower signal-to-noise ratio (SNR). On the other hand, invasive methods such as electrocorticography (ECoG) are hampered by the limited public availability due to the necessity of neurosurgery. Deep learning provides new ideas for understanding the activity paradigm of the human brain, which allows modeling the correlation between evoked brain activity and linguistic information from a higher-dimensional perspective without step-by-step deduction. The previous work adopts classifiers that could distinguish coarse categories of stimuli, while convolutional and recurrent neural networks (CNNs, RNNs) make this classification more fine-grained. The latest achievement lies in the corporation of language models, especially the large language models (LLMs) that demonstrate the capability of inferring and reasoning about brain data \cite{feng2023aligning, feng2024towards, chen2024openvocabulary}. The brain-network alignment has been verified at multiple levels, which ensures the effectiveness of leveraging ANNs in studies of neural progress.

\begin{figure}[htp!]
\centering
\includegraphics[width=1\textwidth]{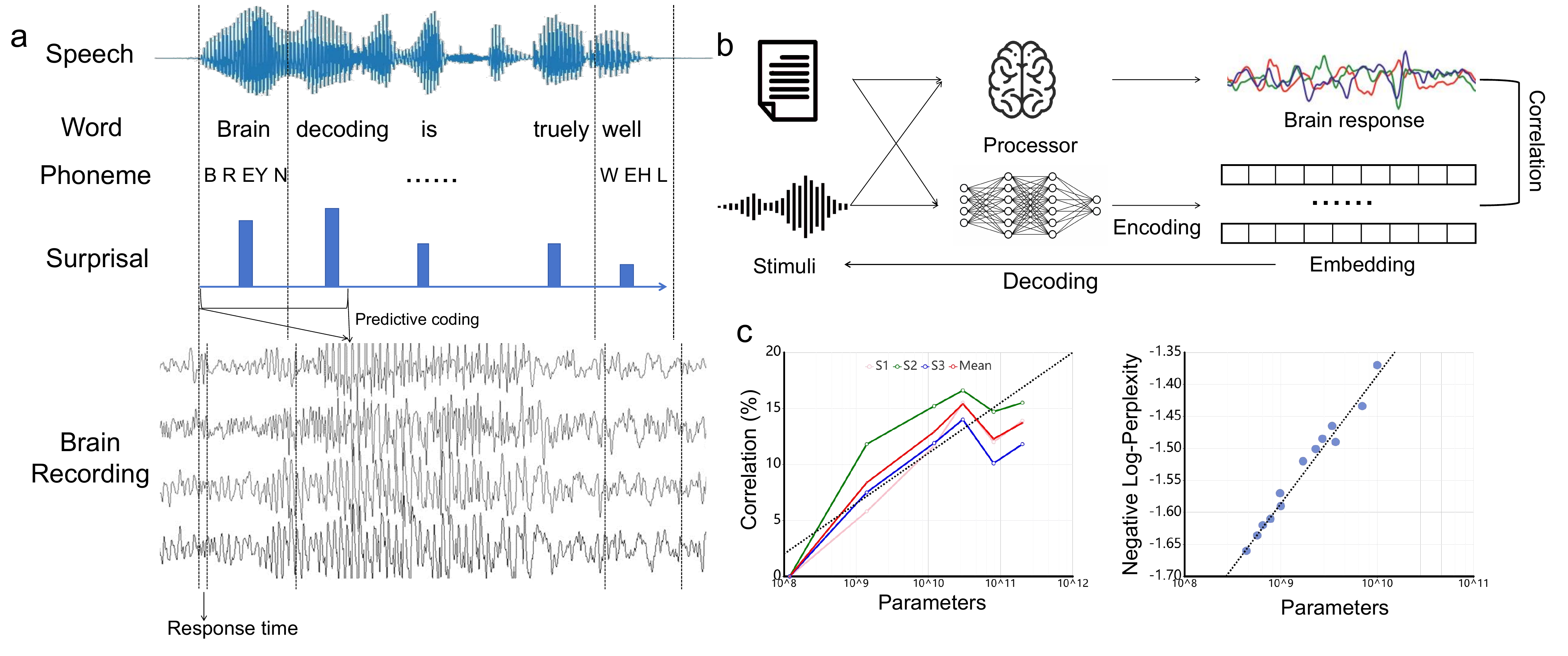}
\caption{\textbf{The formation of linguistic representation in the human brain.} a. The human brain tracks the dynamic flow of speech and linguistic properties with minor response delay, and the neural response is performed in a continuous predictive manner. b. The human brain and the neural networks can both encode textual or verbal-linguistic stimuli into specific representations, and the decoding process aims to generate its original form from the evoked response. c. The scaling laws for the brain encoding models and pre-trained LLMs respectively \cite{antonello2024scaling, lin2024selecting}. For brain encoding, $S_1$, $S_2$ and $S_3$ represent different experiment subjects. The performance grows as the model parameters increase.}\label{fig2}
\end{figure}

The neural tracking process during verbal communication \cite{ahissar2001speech} is shown in Fig.~\ref{fig2}a. The cortical activity automatically tracks the dynamics of speech as well as various linguistic properties, including surprisal, phonetic sequences, word sequences, and other linguistic representations \cite{brodbeck2018rapid, koskinen2020brain, donhauser2020two, gillis2021neural}. A minor time shift has been observed for information transfer and neural response. Neural tracking ensures the temporal alignment of brain recordings with linguistic representations, facilitating the serialized and temporal processing and modeling of cortical activities. As shown in Fig.~\ref{fig2}b, language stimuli are processed into regular evoked brain responses, often referred to as cortical encoding, which begins to emerge early in human life \cite{di2023emergence}. In contrast, brain linguistic decoding aims to reconstruct the stimuli perceived or the intention expressed from active representations with ANNs correlated with high-dimension brain response. In natural listening settings, the human brain encodes a wide range of acoustic features and processes external language stimuli temporally through prediction, highlighting the importance of contextual information in cortical perception, even at the level of single neurons \cite{leonard2024large, khanna2024single}. Predictive processing fundamentally forms the comprehension mechanisms in the human brain, occurring hierarchically in both acoustic and linguistic representations \cite{clark2013whatever, schrimpf2021neural, heilbron2022hierarchy}. This phenomenon underscores the profound impact of context on the forecasting and tracking of ongoing speech streams, necessitating the use of contextual representations to investigate cortical responses \cite{broderick2018electrophysiological,  caucheteux2021disentangling, toneva2022combining}. This characteristic is similar to language models constructed by neural networks, where the same stimuli presented in varying contexts are mapped onto diverse semantic features.

When processing natural language, ANNs and the cortical language network exhibit similar patterns of functional specialization. Research, particularly focused on Transformers and LLMs, shows that the computations of ANNs account for a significant portion of the variance observed in brain activity \cite{whittington2021relating, liu2023coupling}. Showcasing the structures and hierarchy of neural networks not only guides researchers in developing more human-aided language models but also offers neuroscientists new avenues for understanding the brain. Human understanding of the neural processes involved in the perception of linguistic stimuli by the brain is limited. Alternatively, the emergence of high-parameter ANNs has provided an effective method for modeling linguistic stimuli, especially the LLMs trained on large-scale, legitimate text. As shown in Fig.~\ref{fig2}c, it has been verified that the brain encoding models and pre-trained LLMs follow the scaling laws, where the model performance increases as the number of parameters grows, indicating the necessity to develop more complex architecture to bridge the brain activity patterns and human linguistic representations \cite{antonello2024scaling, lin2024selecting}. 


\section{Brain Decoding Evaluation}\label{sec2}


Brain decoding is to generate textual or spoken information from evoked brain activity. As an interdisciplinary field of neuroscience and artificial intelligence, early work on brain decoding mainly follows the paradigm of classification, recognition and sequence decoding. Similar experiments are closely related to machine translation (MT), text-to-speech (TTS) and automatic speech recognition (ASR), which inspire the evaluation of brain decoding.

\begin{table}[htp!]
\centering
\caption{Metrics for brain decoding evaluation. The column of the application shows the scope of these metrics for brain decoding, where TSC represents textual stimuli classification, BRT stands for brain recording translation, ISR refers to inner speech recognition and BTS indicates brain-to-speech tasks.}
\label{metrics}
\begin{tabular}{|c|c|c|c|c|}
\hline
\textbf{Target} & \textbf{Metric} & \textbf{Application} & \textbf{Origin} & \textbf{Methods}  \\ \hline
\multirow{7}{*}{\textbf{Text}} 
    & Accuracy & TSC & Classification & percentage of correct output  \\ \cline{2-5}
    & BLUE & \multirow{3}{*}{BRT}  & \multirow{3}{*}{MT} & precision of n-grams \\ \cline{2-2} \cline{5-5}
    & ROUGE & & & recall of n-grams \\ \cline{2-2} \cline{5-5}
    & BERTScore &  & & semantic similarity  \\ \cline{2-5}
    & PER  & \multirow{3}{*}{ISR} & \multirow{3}{*}{ASR} & phoneme accuracy  \\ \cline{2-2} \cline{5-5}
    & CER & & & character accuracy \\ \cline{2-2} \cline{5-5}
    & WER & & & word accuracy \\ \hline
\multirow{5}{*}{\textbf{Speech}}  & PCC & \multirow{5}{*}{BTS} & Statistics & linear correlation of variables  \\ \cline{2-2} \cline{4-5}
    & STOI & & SI & human intelligibility correlation  \\ \cline{2-2} \cline{4-5}
    & FFE & & & accuracy of pitch (F0)   \\ \cline{2-2} \cline{5-5}
    & MCD & & TTS & accuracy of MFCCs   \\ \cline{2-2} \cline{5-5}
    & MOS & & & subjective human evaluation  \\ \hline
\end{tabular}
\end{table}

Table~\ref{metrics} shows the evaluation metrics. For text decoding approaches, the generated text is typically evaluated by calculating the similarity with the reference transcription. In the textual stimuli classification (TSC) paradigm, accuracy is widely used to measure the percentage of correct instances. As for sequence decoding which aims to reconstruct the linguistic information at the sentence level, most approaches utilize metrics for MT, which focus on semantic similarity. BLEU (Bilingual Evaluation Understudy) \cite{papineni2002bleu} calculates the precision of n-grams compared to reference translations and ROUGE (Recall-Oriented Understudy for Gisting Evaluation) focuses more on recall. In brain decoding, they are usually used in the sequential decoding of non-invasive recordings. BERTScore \cite{zhang2019bertscore} is a recent metric leveraging deep contextualized embeddings from BERT \cite{devlin2018bert} to capture semantic similarity instead of matching exact n-grams. WER (word error rate) is a common metric to evaluate the performance of ASR systems. It measures the accuracy of decoded hypotheses compared to reference transcriptions. In brain decoding, WER is widely adopted for inner speech recognition, which aims to decode the intended speech of subjects without making distinguishable sounds given the invasive evoked brain recordings. It measures by calculating the proportion of insertion, substitution, and deletion errors. In addition to the word-level calculation, CER (character error rate) and PER (phoneme error rate) are carried out on character- and phoneme-level respectively. 

In natural listening and pronouncing scenarios, reconstructing the speech wave can be necessary. The simplest method is to calculate the statistical correlation between the generated and reference speech, with the PCC (Pearson correlation coefficient) showing the most preference. It is a measure of the linear relationship between two continuous variables. STOI (short-time objective intelligibility) \cite{taal2010short} is a metric evaluating speech intelligibility (SI). It is designed to provide an objective measure that correlates well with human subjective intelligibility ratings. FFE (F0 frame error) \cite{taal2010short} and MCD (mel-cepstral distortion) \cite{kubichek1993mel} are the metrics to evaluate the accuracy of pitch and MFCCs respectively, which have been widely used in TTS. MOS (mean opinion score) is commonly used for evaluating the perceived quality of audio, video, and multimedia content. It provides a subjective measure of quality based on human judgments and typically uses a five-point scale where participants rate the quality of the synthesized speech slices.

\section{Stimuli Recognition}\label{sec3}

\begin{figure}[htp]
\centering
\includegraphics[width=1\textwidth]{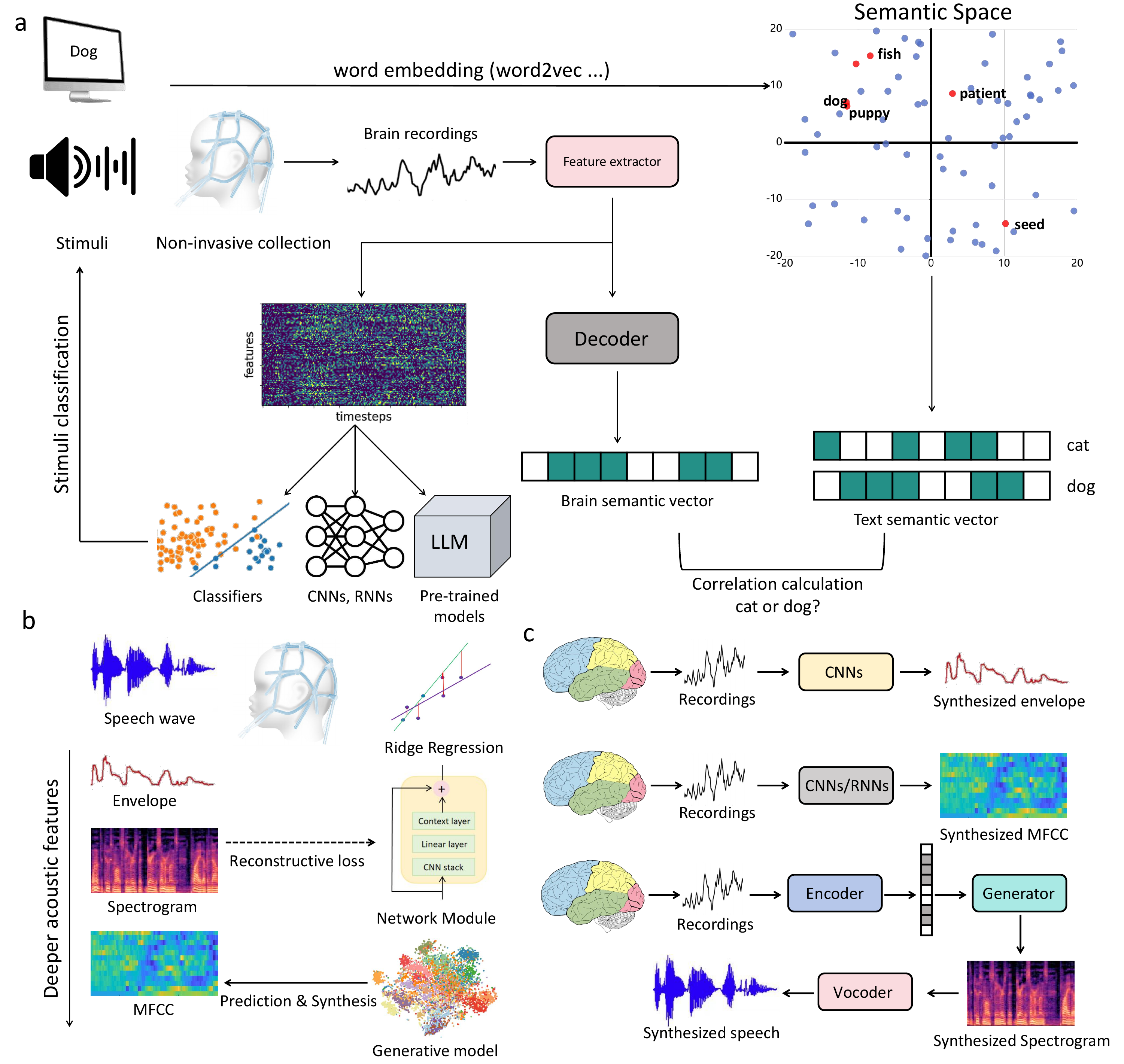}
\caption{\textbf{Stimuli recognition of evoked brain activity.} a. An overview of the stimuli recognition task. The subject receives textual or vocal information while the active brain signals are collected. The raw brain recordings are processed into feature space, followed by classifiers, networks or pre-trained models to distinguish the original stimuli based on the complexity and candidate size. Several approaches adopted Word embeddings (i.e. word2vec) to compare the decoded vector in a semantic space. b. In natural listening scenarios, restoring the original speech features and waveform is a more complex task. Regression models (i.e. ridge regression), CNN and RNN-based network modules, and paramount generation models (i.e. GAN) are widely used. c. The decoding architecture for various speech-related targets. The speech envelope can be easily reconstructed with CNNs while more complex networks are necessary for the decoding of MFCC \cite{accou2023decoding, petrosyan2021compact}. The most difficult task is to synthesize the stimuli wave, where an encoder-generator-vocoder architecture has been verified effective \cite{wang2020stimulus}.}\label{fig3}
\end{figure}

Fig.~\ref{fig3}a shows the experiment setting of stimuli recognition. Compared with fine-grained decoding, a moderate set of candidates is necessary. The subjects passively receive external stimuli, usually by reading text or listening to podcasts, and deep learning methods are adopted to classify the original stimuli based on evoked brain signals.

\subsection{Textual Stimuli Classification}\label{subsec3.1}

The major languages in the world have their textual form, and communication via text outperforms due to its efficiency and concision. When humans receive textual information through reading or other formats, the brain maps the language stimuli to specific corresponding activations. Textual stimuli classification distinguishes the original information provided to the subject within several candidates. The previous approach included investigating the perception methods of concrete nouns to avoid the abstract representations of other concepts \cite{just2010neurosemantic, sudre2012tracking}. Classifiers were adopted to distinguish which concrete noun had been perceived by the subject. Following this, other studies extended the approach to abstract nouns, proving the superiority of text-based models over visually grounded approaches \cite{anderson2017visually}, resulting in the evaluation of 8 different word embedding models for predicting another given either the neural activation patterns or word representations \cite{abnar2018experiential}. In \cite{pereira2018toward}, the researchers presented a brain decoding system based on a semantic space trained on massive text corpora. The decoded representations were detailed enough to differentiate between sentences with similar meanings.

Sentences are the most common unit for human perception of text. The following work treated the sentence-level brain responses as a combination of latent word effects, bridging the relationship of text perception in word and sentence levels \cite{anderson2017predicting, wang2017predicting, anderson2019integrated}. Following these approaches, the holistic encoding of sentence stimuli was proposed \cite{sun2019towards, gauthier2019linking}. Studies further evaluated various distributed semantic models to predict or decipher brain response to textual sentences, with the Transformer-based model achieving the best performance \cite{sun2020neural}. Another classification task is performed on the long-text level (or story level). The researchers predicted the evoked brain response during natural reading and classified the corresponding brain activity by distance to the synthesized brain image \cite{wehbe2014simultaneously}. Incorporating the pre-trained models is a promising solution. In \cite{jat2019relating}, the approach bridged the textual stimuli pattern and MEG recordings using multiple network architecture, with BERT showing the best performance. A much more difficult task is to accomplish text stimuli decoding on larger vocabularies. In \cite{affolter2020brain2word}, a network module with dense layers and a regression-based decoder was implemented to directly classify an fMRI scan over a 180-word vocabulary and evaluated on previously unseen subjects. Although the recognition effect far exceeded the chance probability (5.22\% Top-1 and 13.59\% Top-5 accuracy), it was still far from the application level. Following these achievements, in \cite{zou2021towards} the researchers introduced more difficult tasks, predicting masked words or phrases given the evoked brain images. The proposed approach utilized an encoder-decoder paradigm and achieved 18.20\% and 7.95\% top-1 accuracy over a 2,000-word vocabulary on the two tasks respectively, demonstrating the promise for using brain responses to predict textual stimuli.


Textual stimulus classification is greatly limited by the decoding range and almost performed on dozens or hundreds of candidates, which is separate from real-world applications. As an initial attempt, this task illustrates the possibility of obtaining textual information from the evoked cortex, gradually developing into open vocabulary sequence decoding, and is divided into two task formats in the subsequent introduction - brain recording translation and inner speech recognition.

\subsection{Speech Stimuli Reconstruction}\label{subsec3.2}

Speech perception entails processes that convert acoustic signals into neural representations. In neuroscience, this includes the complete pathway from the cochlear nerve to the auditory cortex areas. Previous research has demonstrated that the hierarchical structure in ANNs trained on speech representations aligns with that of the ascending auditory pathway, supporting the feasibility of using deep learning approaches to model auditory perception \cite{li2023dissecting}.

The speech stimuli reconstruction aims at forming semantic information, acoustic features and synthesized perceived speech from evoked brain activity (Fig.~\ref{fig3}). Classifiers had been used to distinguish perceived stimuli before the deep learning methods were applied. The logistic regression was applied to classify the speech stimuli perceived by an unseen subject during training \cite{liu2018speech}. Inspired by the ASR systems, the phoneme-level Viterbi decoding was introduced to recognize the heard utterance in a question-answering setting \cite{moses2019real}. Another work introduced a contrastive learning model inspired by CLIP \cite{radford2021learning} to predict the correct segment out of 1,000 possibilities \cite{defossez2023decoding}. It leveraged the correlation between speech waves and EEG/MEG time series with wav2vec 2.0 \cite{baevski2020wav2vec} and CNNs as the speech and brain modules respectively. The research of content and subject recognition are not separated, considering the speech flow can be identified in both spaces. One attempt was to adopt variational autoencoders to transform the EEG space into disentangled latent spaces, representing the content and subject distribution respectively \cite{bollens2022learning}.

The speech envelope refers to the variations in amplitude or intensity of a speech signal over time. It plays a crucial role in speech perception and understanding, for our brains are tuned to these variations, helping recognize speech sounds, syllables, and words, even in noisy environments \cite{aiken2008human, ding2014cortical}. Earlier work focused on the signal processing and linear model to align the envelope representation with brain activity \cite{vanthornhout2018speech}. After that, some other research implemented convolutional models \cite{crosse2016multivariate, accou2021modeling} or based on mutual information analysis \cite{de2023beyond}. In \cite{thornton2022robust}, the researchers evaluated the envelope construction performance of ridge regression, convolution and fully connected layers (FCs). The more in-depth research led to the development of the VLAAI deep neural network, a convolutional-based architecture to achieve more precise reconstruction \cite{accou2023decoding}. Considering the highly robust correlation between envelope and linguistic information, some extended to a cocktail party setting, where attended speech envelope was predicted with a context-aware neural network \cite{de2020machine}. A recent work adopted a transformer-based encoder-decoder architecture \cite{xu2022decoding}. The envelope reconstruction also served as a downstream task to evaluate the pre-trained models of brain signals \cite{de2018decoding, zhu2023eeg2vec}. Compared with the speech envelope, MFCC is a widely used feature in speech recognition that represents the short-term power spectrum of sound. The parallels between speech recognition and brain-to-text technologies inspired the prediction of MFCC from brain recordings using regression and generative models \cite{krishna2019state}, as well as RNN-based modules \cite{petrosyan2021compact}. Subsequent research extended this approach to various acoustic features, predicting 16 different types using an attention-based regression model \cite{krishna2021advancing}.

Instead of reconstructing the acoustic features, synthesizing speech directly from brain recordings is more challenging, yet it holds greater practical significance and application prospects. In \cite{pasley2012reconstructing}, the researchers opened up the possibility of speech restoration with evoked brain recordings. This approach implemented a linear spectrogram model with strict recording quality and word selection requirements. The following studies investigated the reconstruction performance of linear and non-linear models based on speech spectrogram and vocoder parameters of the synthesizer respectively \cite{akbari2019towards}. The result demonstrated the significance of non-linear neural networks directly estimating the synthesizer parameters. Other studies leveraged Wasserstein GAN (wGAN) \cite{arjovsky2017wasserstein} for generator pre-training to obtain the spectrogram representation \cite{wang2020stimulus} and dual generative adversarial network (DualGAN) \cite{yi2017dualgan} for cross-domain mapping between EEG signals and speech waves \cite{guo2023end}. In this field, network optimization contributes to performance improvement, with the self-attention module demonstrating its superiority to multi-layer perceptrons (MLPs) and CNNs to restore the spectrogram \cite{senda2024auditory}.

Compared with text, speech contains richer information related to the speaker's emotion and identification, etc., which brings more challenges to restoring the speech stimulation received by the subject. Brain stimuli reconstruction has confirmed the correspondence between speech features and brain representations. From the current perspective, reconstructing recognizable speech waveforms requires multiple rounds of iterations of recording quality and network architecture.

\section{Brain Recording Translation}\label{sec4}

\begin{figure}[htp]
\centering
\includegraphics[width=1\textwidth]{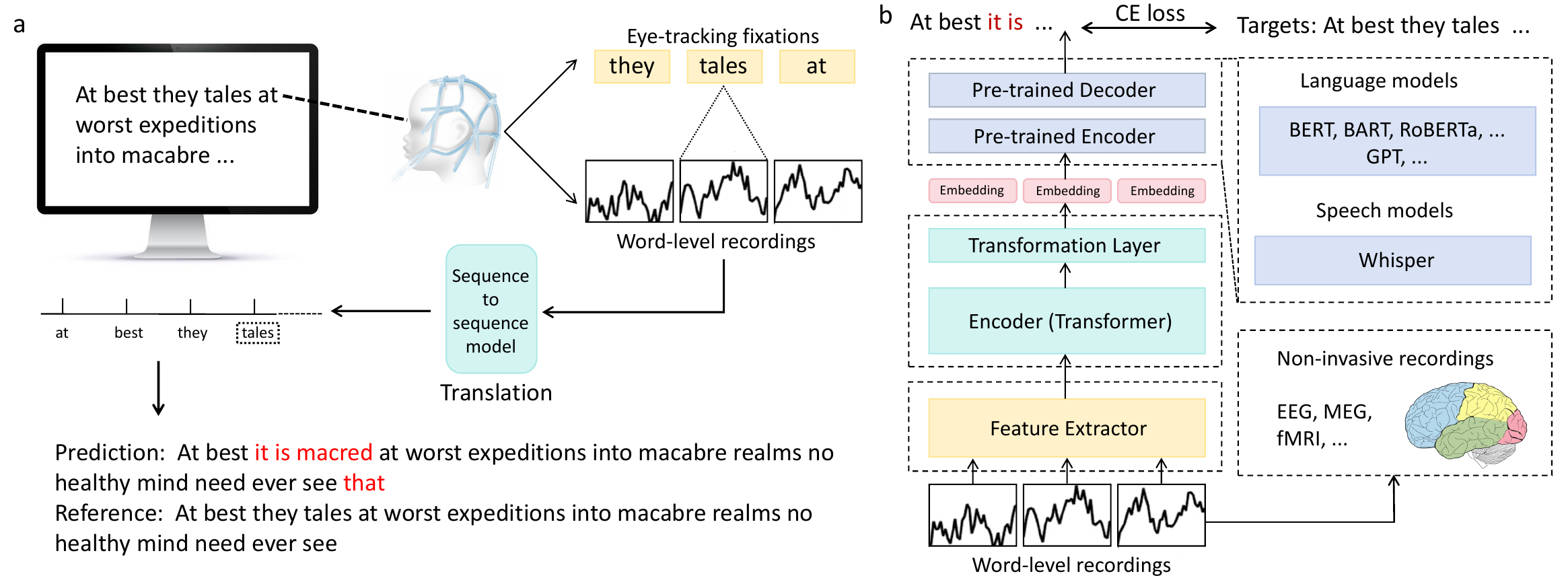}
\caption{\textbf{The experiment setting and model architecture of brain recording translation.} a. Taking the natural reading scenario as an example, the subject performs natural reading while the active brain signals are collected. The eye movements are typically recorded to determine the text transcription corresponding to the brain data at each time step. A sequence-to-sequence model processes the evoked brain recordings to determine the related word and then form the decoded sentence. b. A feasible translation model architecture, including feature extraction, feature transformation and a pre-trained encoder-decoder to generate the decoding sentence. Both the pre-trained language models (i.e. BART) and speech models (i.e. Whisper) have been verified effective.}\label{fig4}
\end{figure}

Decoding natural sentences from brain signals remains a significant challenge. Unlike simpler tasks that convert brain signals into categorical labels, brain recording translation directly decodes linguistic stimuli (Fig.~\ref{fig4}). This process borrows concepts and evaluation methods from machine translation, as both tasks aim to map representations between two different units of analysis. Brain recording translation involves decoding open vocabulary based on brain activity, which implies a vast search space. However, decoding linguistic stimuli from brain recordings fundamentally differs from machine translation, for the stimuli text or speech is deterministic, while in machine translation, the potential targets can be numerous. Given the complexity of language and the resolution limitations of neuroimaging, this task demonstrates the balance between the invasiveness of data collection and the precision of recognition.

The brain recordings are typically collected during natural reading and listening scenarios. The subjects read given textual information or listen to podcasts, while their brain activity is recorded (Fig~\ref{fig4}a). Researchers reconstruct text stimuli through deep learning solutions. In \cite{makin2020machine} the authors first introduced the concept of machine translation into brain decoding. Although this work aimed to generate the word sequence during attempted speech, the serialization of text generation provided new insights for subsequent work. The neural network architecture contained temporal convolution to model contextual relations and encoder-decoder RNNs to generate predicted text. The experiment was conducted with ECoG recordings and carried out on a vocabulary list of several hundred words. The following work turned to the BLEU and ROUGE scores \cite{wang2022open}. This work largely expanded the decoding vocabulary ($\sim$50,000) by fully leveraging the inference capabilities of pre-trained LLMs. Specifically, a multi-layer Transformer encoder is used to map non-invasive EEG features to the embedding space of the BART tokenizer \cite{lewis2019bart}, and the decoded sentence is generated through the decoder of BART. Following these achievements,  this paradigm was progressed by directly interpreting raw brain signals with contrastive learning methods borrowed from VQ-VAE \cite{van2017neural} and introducing discrete encoding into the EEG recording representation \cite{duan2024dewave}. However, their models were highly estimated with teacher-forcing schema during evaluation \cite{jo2024eegtotext}, preventing them from generating meaningful sentences in real-life applications.

Alternatively, a solution with implementation potential was proposed, which made the first attempt to generate text directly from MEG recordings without teacher forcing \cite{yang2024decode}. The proposed architecture, named NeuSpeech, is a modification of Whisper \cite{radford2023robust}, with the parameters in the encoder tuned for the translation task. Compared to other approaches, NeuSpeech is more straightforward and corroborates the association between brain activity and speech signals. It is worth mentioning that NeuSpeech was evaluated on previously seen text, while the advanced solution contributed to an open-vocabulary MEG-to-text translation model capable of generating unseen text \cite{yang2024mad}. In this novelty approach, multiple alignments were conducted between the MEG recordings and speech audio. The brain module was mapped to Whisper representations in three aspects, the Mel spectrogram, hidden state and decoded text. Another work proposed simultaneously leveraged the reasoning ability of LLMs and implemented an fMRI encoder to learn a suitable prompt in an auditory-decoding setting. The prompt of text and fMRI modalities are aligned through a contrastive loss \cite{chen2024openvocabulary}.

\begin{table*}[t!]
\caption{Brain recording translation performance of various model architecture. The column of TF refers to whether to implement teacher forcing during inference. The performance of MAD deteriorates for it is evaluated on previously unseen text.}
\begin{adjustbox}{center}
\centering
\label{BRT}
\begin{tabular}{|c|c|c|c|c|c|c|c|}
\hline
\multirow{2}{*}{Model} &\multirow{2}{*}{TF} & \multicolumn{2}{c|}{Data} & \multicolumn{4}{c|}{Performance(\%)} \\ \cline{3-8}
&  & Dataset & Type & \footnotesize BLUE-1 & \footnotesize ROUGE-1 & \footnotesize BERTScore & \footnotesize WER \\ \hline
Transformer+BART\cite{wang2022open} & \ding{52} & ZuCo\cite{hollenstein2018zuco, hollenstein2019zuco} & \footnotesize EEG & 40.1 & 30.1 & - & - \\ \hline
EncodingModel+GPT-1\cite{tang2023semantic} & \ding{55} & Self-collect & \footnotesize fMRI & 24.1 & - & 81.0 & 93.3 \\ \hline
BrainTranslator(BART)\cite{feng2023aligning} & \ding{52} &  ZuCo\cite{hollenstein2018zuco, hollenstein2019zuco} & \footnotesize EEG & 35.9 & 39.1 & - & 68.5 \\ \hline
DeWave(BART)\cite{duan2024dewave} & \ding{52} & ZuCo\cite{hollenstein2018zuco, hollenstein2019zuco} & \footnotesize EEG & 41.4 & 30.7 & - & - \\ \hline
\multirow{2}{*}{UniCoRN(BART)\cite{xi2023unicorn}} & \multirow{2}{*}{\ding{52}} & Narratives\cite{nastase2021narratives} & \footnotesize fMRI & 62.9 & 59.5 & - & - \\ \cline{3-8}
 & & ZuCo\cite{hollenstein2018zuco, hollenstein2019zuco} & \footnotesize EEG & 57.7 & 64.4 & - & - \\ \hline
NeuSpeech(Whisper)\cite{yang2024decode} & \ding{55} & \footnotesize MEG-MASC,MOUS\cite{gwilliams2023introducing,schoffelen2019204} & \footnotesize MEG & 57.7 & 60.2 & - & 58.9 \\ \hline
MAD(Whisper)\cite{yang2024mad} & \ding{55} &  MEG-MASC\cite{gwilliams2023introducing} & \footnotesize MEG & 10.4 & 6.9 & 83.4 & - \\ \hline
BP-GPT(GPT-2)\cite{chen2024openvocabulary} & \ding{55} &  Podcasts\cite{lebel2023natural, tang2023semantic} & \footnotesize fMRI & 26.0 & - & 84.3 & - \\ \hline
\end{tabular}
\end{adjustbox}
\end{table*}

The setting of brain recording translation is reasonable. Under this paradigm, more work emerged that implements LLMs to translate brain signals in large vocabularies, including schemes using contrastive learning and curriculum learning \cite{feng2023aligning}. A similar approach was also used for decoding fMRI signals, which used an encoder-decoder architecture with BART as a text generator \cite{xi2023unicorn}. The reconstruction loss of fMRI signals was used to train a better encoder, and the discretized EEG signal and the text vector after word2vec \cite{mikolov2013efficient} were fed to the contrastive learning module in EEG-text pairs, in which the EEG representation aligned with pre-trained language models inspired by the CLIP model \cite{radford2021learning}. Another method of experiment was to collect brain recordings while participants listened to narrative stories \cite{tang2023semantic}. The fMRI data was sent into GPT after the feature extractor to complete the sequence generation task. 

The models of brain recording translation, especially the structures proposed in the past year, and their performance on various datasets are shown in Table~\ref{BRT}. The word sequence decoded from the non-invasive brain signal shows great disparity with the original textual signal, as reflected in the high WER, while they are consistent with semantic correlation, achieving a promising BERTScore. Considering the promotion prospects of non-invasive signal acquisition equipment, this is a feasible BCI design, which does not require accurate decoding of text information but focuses more on semantics reconstruction.



\section{Speech Neuroprosthesis}\label{sec5}

Neurological diseases that result in the loss of communication abilities are devastating. Many patients rely on BCIs to select letters to spell words \cite{metzger2022generalizable}, moving computer cursor \cite{leuthardt2011using} or direct handwriting \cite{willett2021high}. Although these systems can improve the quality of life for patients, communication efficiency is a concern. A major challenge is to overcome the limitations of current spelling-based methods to achieve higher and more natural rates of communication. The goal of speech neuroprosthesis (SN) is to directly decode the words or speech waves the experiment participants intend to speak from their brain signals (Fig.~\ref{fig:fig5}). The participant usually loses the ability to speak recognizably due to damage to their vocal organs or neurological impairments but has an intact brain. This represents a hopeful path for creating devices that assist in voice communication.

\subsection{Inner Speech Recognition}

The inner speech was first called imagined speech in a two-phoneme classification task \cite{brigham2010imagined}. The subjects have typically lost their ability to produce recognizable sounds, and the brain signals are recorded as they try to speak. In some experiments, the brain signals during vocal speech are also collected. Unlike brain recording translation, inner speech recognition demands high-quality brain waves, as high-resolution neural recordings improve the accuracy of speech decoding \cite{duraivel2023high}. This task is highly correlated with ASR, for they both: 1) model the relation between diverse temporal features and deterministic textual information; 2) correlate with pronunciation and acoustics; 3) aim to generate language-compliant text. Compared with other tasks, the neural activity corresponding to pronunciation has a strong correlation with the movement of the articulators. A recent study shows that even at the level of a single neuron, there are significant neural representations related to inner and vocalized speech that are sufficient to discriminate between words from a small vocabulary \cite{Wandelt2024RepresentationOI}.

Phonemes, recognized as the foundational elements of speech pronunciation, have historically been the focus of initial studies aiming to decipher human articulatory patterns through brain activities. Previous studies have provided evidence for the neural representation of phonemes and other acoustic features during the perception of speech \cite{chang2010categorical, mesgarani2014phonetic}. The pioneer attempted to apply instance-based matching algorithms and demonstrated the feasibility of text decoding from brain recordings even without learning for features \cite{suppes1997brain, suppes1998brain, d2009toward}. The following research concentrated on identifying these phonetic units, framing the task similarly to classification due to the relatively narrow scope of phoneme varieties. Experiments have been conducted using linear classifiers \cite{tankus2012structured}, support vector machine (SVM) \cite{dasalla2009single, wang2013analysis, stavisky2018decoding}, naive Bayes classifier \cite{pei2011decoding}, k-nearest neighbor classifier \cite{brigham2010imagined}, linear discriminant analysis (LDA) classifier \cite{deng2010eeg, kim2014eeg, moses2016neural}, flexible discriminant analysis (FDA) \cite{brumberg2011classification} and based on brain recording features after principal component analysis (PCA) \cite{kellis2010decoding}. The above work was conducted with few phoneme candidates with clear acoustic boundaries. Following this, the researchers achieved full-set phoneme decoding of American English \cite{mugler2014direct}, and implemented a similar approach with brainwave recorded by mobile EEG devices \cite{clayton2020decoding}.

\begin{figure}[htp]
\centering
\includegraphics[width=1\textwidth]{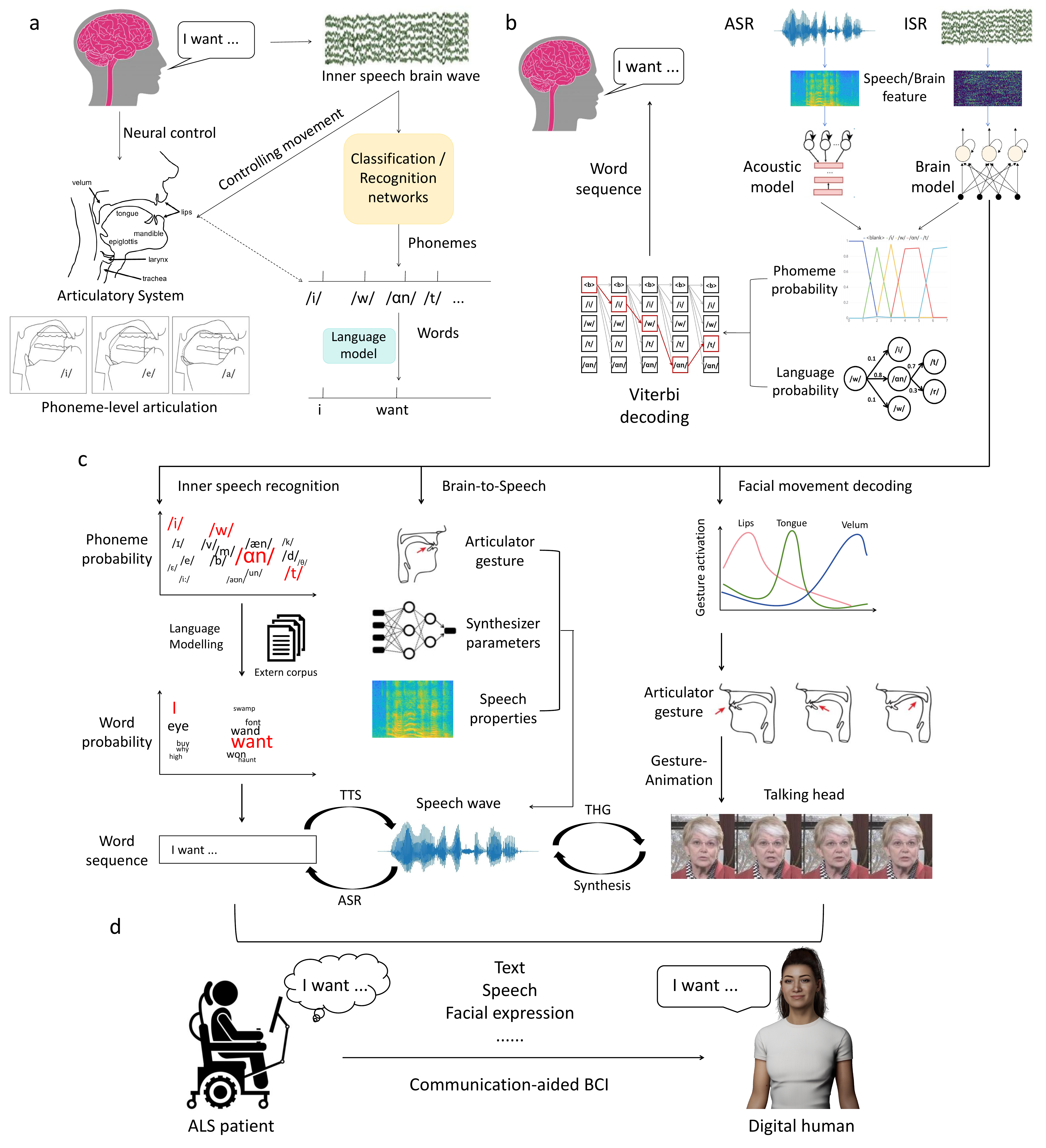}
\caption*{}\label{fig5}
\end{figure}

\clearpage

\begin{center}
    \captionof{figure}{\textbf{Overview of speech neuroprosthesis.} a. The experiment setting for inner speech recognition. The subjects attempt to speak without making a sound (inner speech) or try their best to pronounce (overt speech), while their active brain signals are collected. From the perspective of neurological, the brain controls the movement of the articulatory system to complete the pronunciation of each phoneme in series, thereby producing recognizable speech, indicating the mapping from evoked brain signals to movements of the articulators to phonemes. The classification and the recognition module are adopted to generate the corresponding phoneme sequences before leveraging the language model to form word sequences.b. The comparison between ASR and inner speech recognition (ISR). The raw time-series signals are processed for feature extraction and then fed into acoustic and brain models respectively. Both models aim to bridge the relationship between learnable features related to acoustics and phoneme sequences. The Viterbi decoding algorithm is performed on the sum of the phoneme probability from the acoustic/brain model and the language probability derived from a language model trained on an extensive corpus to generate the decoded word sequences. c. The brain model can be implemented to decode various modalities. For inner speech recognition, the phoneme and word sequences are decoded with the aim of language models. For brain-to-speech decoding, the speech waves are synthesized according to the articulator gestures, synthesizer parameters or speech properties. By modeling the articulator gesture probability and adopting a gesture-animation system, the talking head can be generated. Different modalities are associated through TTS, ASR, talking head generation (THG) and synthesis methods. d. The acoustic-related brain activities show the potential to develop communication-aided BCI for ALS patients considering the decoding feasibility of text, speech and facial expression. }
    \label{fig:fig5}
\end{center}


Progressing from phonemes, researchers achieved advancements toward decoding brain signals into words within a modest vocabulary range. Many investigations were conducted in severely restricted word sets with clearly distinguishable pronunciations. Due to the small vocabulary, such approaches typically employ classifiers for the selection of candidate words. In \cite{mohanchandra2016communication} the researchers introduced a human-defined lexicon-"water", "help", "thanks", "food" and "stop". The multiclass SVM was used for intended speech decoding. The following work implemented the same architecture to accomplish individual word classification \cite{martin2016word}, while others adopted the relevance vector machine, a probabilistic model incorporating a Bayesian inference framework \cite{nguyen2017inferring}. Another work based on classification distinguished five words in Spanish and focused on the multiple-modality fusion of text, sound and EEG brain wave \cite{gonzalez2017sonification}. The most recent achievements conducted the illustration of speech-related representation on single neuron level recognition \cite{Wandelt2024RepresentationOI}. The LDA classifier was adopted to distinguish six words and two pseudowords. Deep learning methods have also been applied to the recognition of imagined speech. The premier attempt implemented several networks to classify imagined words "yes" and "no" \cite{salama2014recognition}, followed by research utilizing deep belief neural networks for brain activity feature extraction as well as phoneme and word recognition \cite{zhao2015classifying}. The cascade approaches divided the pipeline into convolutional-based cascade modules, including an MFCC prediction module and a word classification model \cite{petrosyan2021compact}. Network structures with larger parameters are suitable for more complex recognition units, for instance, conducting long word recognition using a mixed ANN approach containing CNNs and RNNs \cite{saha2019hierarchical}. To test the recognition performance of the network model on longer units, the researchers investigated the decoding performance of five imagined and spoken phrases with FCs and CNNs \cite{dash2020decoding}.

The challenge of low SNR in brain signal recordings, primarily from non-invasive techniques, is a significant obstacle to expanding the decoding repertoire \cite{ball2009signal}. It is posited that invasive techniques might be indispensable for more detailed decoding tasks encompassing a broader vocabulary. In \cite{makin2020machine}, the authors achieved word sequence decoding on a vocabulary of 250 words using RNN-based encoder-decoder architecture with invasive ECoG recordings. The most promising approach to generating sentences originates from ASR. Specifically, the hybrid model ASR includes an acoustic model, a language model, and a lexicon. The acoustic model calculates the scores of recognition units and then added to the language model scores followed by a search algorithm to generate the decoding hypothesis. The cascade speech neuroprosthesis replaces the acoustic model with a brain model and decodes the corresponding phoneme or small-vocabulary word hypothesis before generating the sentences \cite{herff2015brain, moses2021neuroprosthesis}. These works typically adopted the Gaussian mixture model (GMM) to fit the data distribution of invasive brain activities. Such approaches did not make a groundbreaking impact until the replacement from GMM to ANNs contributed to the improvement that could rival the effectiveness of ASR systems \cite{willett2023high, metzger2023high}. This groundbreaking work used RNNs to model the mapping relationship between invasive brain activity and phonemes. The phoneme scores, in conjunction with an n-gram language model trained on a large amount of external text with Kaldi \cite{povey2011kaldi}, worked together through the Viterbi search algorithm to decode sentence hypotheses, and a lexicon established the connection from phonemes to words. Through this work, researchers achieved a 25.8\% WER on a vocabulary of 125,000 words within the acceptable bound of performance \cite{munteanu2006measuring}, with a recognition rate of 62 words per minute. A similar previous work was proposed \cite{sun2020brain2char}, in which an encoder-decoder architecture with a feature regularization module was used to decode character sequences from ECoG recordings. However, the regularization process consumed acoustic and articulatory kinematic features, which are unavailable for ALS patients. The continuous speech decoding has extended to logosyllabic languages like Mandarin Chinese, designing three CNNs to predict the initials, tones and finals of Pinyin, a phonetic text input system based on the Latin alphabet, followed by a language model to convert Pinyin sequences to Chinese sentences \cite{feng2023high}. The prediction of initials was based on the articulatory feature, including the place and manner of articulation and whether voiced or aspirated. A more convincing result appeared in multilingual recognition, where the participant was presented with the target phrases either in English or Spanish \cite{silva2024bilingual}. The evoked ECoG brain signals were sent to serial RNNs and dense layers to produce the possible probabilities across both languages before being processed by language models on each language separately. The end-to-end (E2E) system is parallel to the hybrid model, which directly models the series signals and text information, avoiding the optimization of individual modules. In \cite{yuan2024improving}, encoder-decoder RNNs were implemented to recognize the vocal speech using invasive ECoG recordings, where the representations generated by revised wav2vec \cite{schneider2019wav2vec} yielded superior decoding performance to the original ECoG data. Another recent approach introduced an E2E framework with pre-trained LLMs for decoding invasive brain signals, leveraging the comprehensive reasoning capability of GPT-2, OPT and Llama2 \cite{radford2019language, zhang2022opt, touvron2023llama, feng2024towards}. As an initial attempt, the E2E model achieved comparable performance to the cascade model, demonstrating a promising avenue. 

Since ANN-based cascade inner speech recognition achieved efficient and accurate performance, breakthroughs in this field have accelerated. However, invasive data collection introduces medical risks which makes it difficult to promote among patient groups. Additionally, it has been verified that brain patterns vary over time and in subjects \cite{willett2023high}. We believe that inner speech recognition is the most promising solution for communication-aided BCIs, but there's still a distance from a high-security, high-quality, and low-latency strategy.



\subsection{Brain-to-Speech}

Another challenging approach is to directly synthesize speech waves from brain signals. In practical applications, these two paradigms are essentially equivalent, given the maturity of text-to-speech (TTS) technology. Neuroprostheses using speech synthesis employ deep learning models to convert brain activity records sequentially into synthesizer commands \cite{guenther2009wireless, chen2024neural}, kinematic features (e.g., amplitude envelope), or acoustic features (e.g., pitches, MFCC) \cite{anumanchipalli2019speech, krishna2020speech}, thereby reconstructing the original speech signal. For instance, a study implemented the DenseNet regression model \cite{huang2017densely} to map ECoG features to the logMel spectrogram \cite{angrick2019speech}. Articulatory-based speech synthesizers generate intelligible speech signals from primary speech articulators using articulator representations \cite{bouchard2013functional} or electromagnetic articulography (EMA) \cite{bocquelet2014robust, bocquelet2016real}. EMA measures the position of mouth articulators: the tongue, lips, velum, jaw, and larynx. This method is based on the finding that during speech production, activity in the brain's sensorimotor cortex closely aligns with articulatory characteristics \cite{cheung2016auditory}. Additionally, various features related to synthesized speech, such as vocal pitch \cite{dichter2018control}, articulatory kinematic trajectories \cite{chartier2018encoding, mugler2018differential}, and speech energy \cite{angrick2021real}, can be identified based on brain activity. Speech synthesis without relying on deep learning, such as unit selection, has also been extensively studied \cite{herff2019generating}. Besides synthesizing intelligible waves, researchers are also focusing on generating spontaneous speech, including speech with accurate lexical tones. A feasible approach involves constructing specific neural networks to separately decode the neural activities of tones and syllables, then using the combined decoded features to synthesize tonal speech \cite{liu2023decoding}.

In addition to speech synthesis, information related to other modalities can be obtained through invasive brain signals. The most intuitive attempt is to leverage articulator gestures for facial movement synthesis \cite{metzger2023high}, which can be achieved by decoding orofacial representations in the speech motor cortex \cite{bouchard2013functional, chartier2018encoding}. It has been verified that facial movement could be generated using an avatar-animation system, and the progress on talking head generation inspired restoring the patient's own face \cite{sun2023vividtalk}. In theory, multiple elements of building a digital human can be obtained from invasive brain activity, including sentences of inner speech, speech waves, facial movements, as well as body movements not related to language \cite{song2024continuous, wang2024neural}. This may be the future development direction of communication-aided BCIs, which can restore the patient's dignity to the greatest extent possible and communicate with the outside world through a virtual image that is the same as a normal person's (Fig.~\ref{fig:fig5}d). For patients who are confined to bed and unable to move, especially ALS patients, this can greatly improve their quality of life.

\section{Future Trends}\label{sec6}

\begin{figure}[htp]
\centering
\includegraphics[width=1\textwidth]{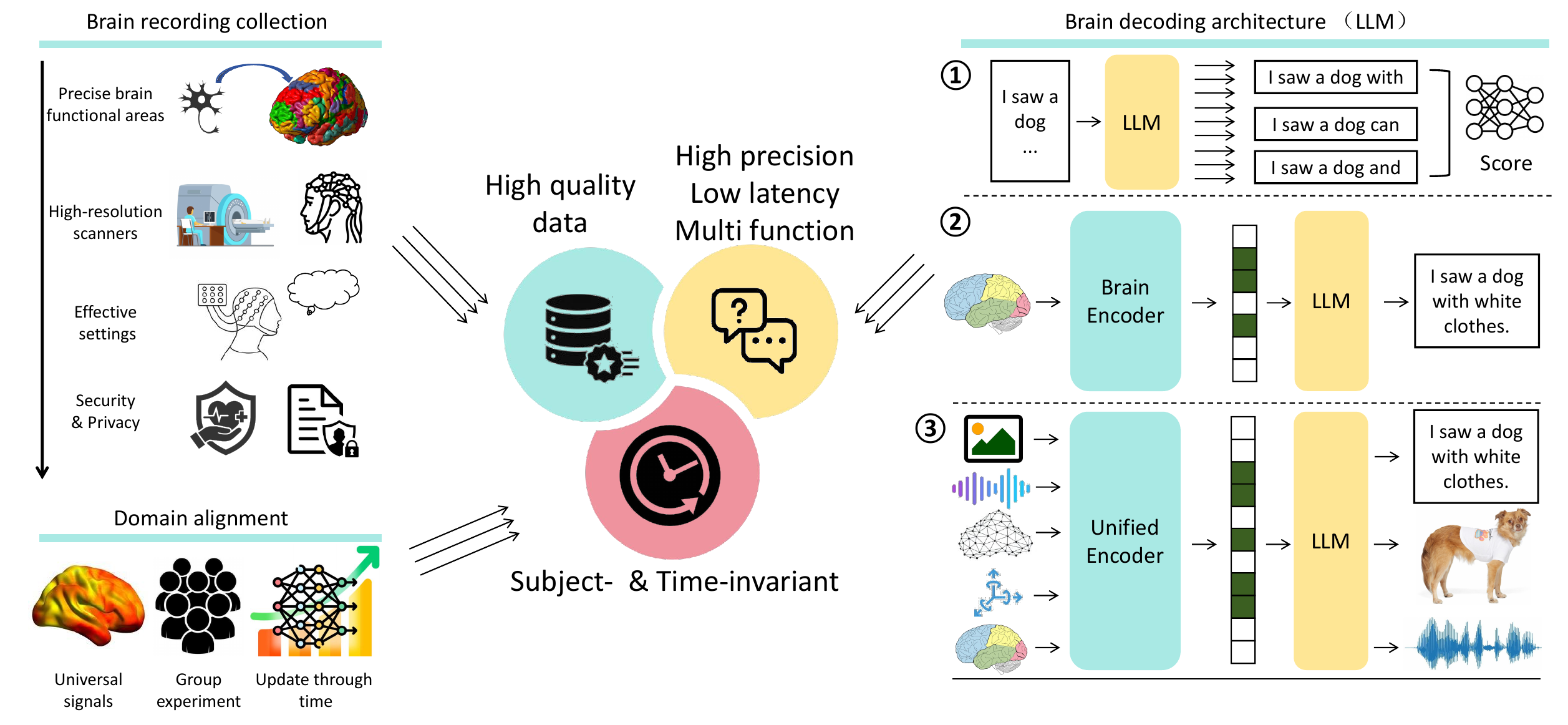}
\caption{\textbf{Characteristics of an ideal BCI system for communication its achieving solutions.} The BCI system requires high-quality brain recordings and addresses the problem of individual and time differences through strategies such as domain alignment. Additionally, the reform of network structure, especially the application of LLM, provides ideas for high-precision, low-latency, and multi-functional interactions.   }\label{fig6}
\end{figure}

Language is the primary means of human communication, and decoding linguistic information from brain activity is crucial for the development of future BCIs. Despite its promising potential, brain decoding faces several challenges, including the invasiveness of data collection, subject specificity, and limitations in accuracy and efficiency. An ideal BCI system for communication should possess the following characteristics (Fig.~\ref{fig6}):

\begin{itemize}
    \item Consume non-invasive, high-quality data: Even though the invasive recording outperforms with its superior qualities of brain imaging, the necessary surgery and unbearable medical risks prevent its spread in patients. The collection of high-quality non-invasive data is a prerequisite for word-level fine-grained sequential decoding.
    
    \item Subject- and time-invariant: For the same neural stimulation, brain activity varies across subjects and acquisition time \cite{willett2023high}. On a small vocabulary, a 3-month clinical trial in an ALS patient showed that speech commands could be accurately detected and decoded without recalibrating or retraining the model \cite{Luo2023StableDF}, and another study showed that the developed decoding system worked successfully in two human patients \cite{Wandelt2024RepresentationOI}. Future research must focus on developing universal decoding schemas that are robust across different subjects and temporal variations. 
    \item High precision, low latency and multi-function: The upper bound of speech-related BCIs can be viewed as a corresponding ASR system, considering the unified backend of the neural networks and the superposition of noise from the brain response to speech. On the Gigaspeech benchmark \cite{chen2021gigaspeech} close to daily scenery, the ASR system can achieve a WER of around 10\%, which can be considered a decent decoding performance. The development of more sophisticated and responsive BCIs could revolutionize how we interact with machines, offering applications in medical rehabilitation, verbal communication and even entertainment. Furthermore, integrating multiple modalities—such as visual and auditory inputs—can enhance the functionality of BCIs, enabling more comprehensive communication solutions. 
\end{itemize}

Current research is conducted in multiple directions. In neuroscience, a pressing issue is the precise collection of neural recordings related to language processing, including acoustic and phonological aspects. This requires identifying specific neuronal populations and brain areas involved in language functions \cite{leonard2024large, khanna2024single, tankus2024machine}. High-resolution scanners, wearable neurotechnology devices and advanced equipment are also necessary \cite{feinberg2023next}, and more reasonable experimental settings need to be explored. Additionally, ethical considerations regarding the security and privacy of neural data, especially from neurologically impaired patients, must be addressed. To overcome the variability across subjects and over time, a universal form of brain signals needs to be explored, and group patient involvement is crucial. Alternatively, continuously updating model parameters over time may also mitigate performance degradation.


From a machine learning perspective, brain decoding is developing towards the subfield of NLP and is closely related to ASR and MT. Inspired by these areas, incorporating multi-modality to improve the decoding performance is widely used \cite{ikegawa2024text}. The most important tendency is the incorporation of inferring and reasoning capabilities of LLMs, which have created disruptive breakthroughs in brain decoding and even deep learning in the past few years \cite{feng2023aligning, feng2024towards, chen2024openvocabulary}. LLMs have been widely used in predicting brain activity and decoding stimuli based on evoked brain response, demonstrating that LLMs trained with tremendous text data can understand limited brain features. As shown in Fig.~\ref{fig6}, in the initial attempts, the LLMs were adopted to generate hypothesis candidates with a separate module score for each potential sentence \cite{ antonello2024many}. A more promising approach treats the LLM as the inferring core to generate correlated textual information \cite{feng2024towards}, and gradually evolves into a unified decoding system with multi-modality inputs and user-specified output \cite{han2024onellm}. We believe that the update and iteration of LLMs will promote qualitative changes in brain decoding, thereby achieving application levels in the near future.

So far, brain decoding has been used for neuron studies and medical concerns, especially for linguistic stimuli reconstruction and real-time communication for ALS patients. Deeper research indicates a wider range of applications. Progress in decoding complex neural signals has shown the potential to reconstruct thoughts, images, or even dreams. This could have profound implications for understanding consciousness and developing communication methods. The interaction between the brain and the environment is bidirectional, but we have only achieved decoding from evoked brain responses to textual information. Stimuli encoding, by performing tiny simulated currents on the cortex to generate evoked brain activity, might be a solution for sensory loss such as blindness and deafness. It is worth mentioning that synthesized brain stimuli carry significant medical risks and ethical controversies, and the black-box nature of neural networks brings unknown dangers to the subject. Guiding brain cognition through artificial stimulation, commonly known as deep brain stimulation (DBS), is a promising direction for disease treatment and has emerged as an effective treatment for neurological conditions such as Alzheimer’s \cite{Tatulian2022ChallengesAH} and Parkinson's disease \cite{Bucur2022DeepBS}. Another important question is whether BCIs can improve the efficiency of information transmission. Information interaction via voice or visual text is limited by the rate of speech flow and vision refresh, while the brain's information reception rate may far exceed both thresholds. When machine operating efficiency reaches a certain level, a large-scale industrial revolution may come from a leap in information transmission efficiency. In general, brain linguistic decoding is a cross-disciplinary collaboration. We expect a further revolution from strengthened cooperation between neuroscience, engineering, psychology, and machine intelligence to promote innovation and accelerate the development of brain signal recording technology and its applications.


\bibliography{sn-article}


\begin{thebibliography}{164}
\ifx \bisbn   \undefined \def \bisbn  #1{ISBN #1}\fi
\ifx \binits  \undefined \def \binits#1{#1}\fi
\ifx \bauthor  \undefined \def \bauthor#1{#1}\fi
\ifx \batitle  \undefined \def \batitle#1{#1}\fi
\ifx \bjtitle  \undefined \def \bjtitle#1{#1}\fi
\ifx \bvolume  \undefined \def \bvolume#1{\textbf{#1}}\fi
\ifx \byear  \undefined \def \byear#1{#1}\fi
\ifx \bissue  \undefined \def \bissue#1{#1}\fi
\ifx \bfpage  \undefined \def \bfpage#1{#1}\fi
\ifx \blpage  \undefined \def \blpage #1{#1}\fi
\ifx \burl  \undefined \def \burl#1{\textsf{#1}}\fi
\ifx \doiurl  \undefined \def \doiurl#1{\url{https://doi.org/#1}}\fi
\ifx \betal  \undefined \def \betal{\textit{et al.}}\fi
\ifx \binstitute  \undefined \def \binstitute#1{#1}\fi
\ifx \binstitutionaled  \undefined \def \binstitutionaled#1{#1}\fi
\ifx \bctitle  \undefined \def \bctitle#1{#1}\fi
\ifx \beditor  \undefined \def \beditor#1{#1}\fi
\ifx \bpublisher  \undefined \def \bpublisher#1{#1}\fi
\ifx \bbtitle  \undefined \def \bbtitle#1{#1}\fi
\ifx \bedition  \undefined \def \bedition#1{#1}\fi
\ifx \bseriesno  \undefined \def \bseriesno#1{#1}\fi
\ifx \blocation  \undefined \def \blocation#1{#1}\fi
\ifx \bsertitle  \undefined \def \bsertitle#1{#1}\fi
\ifx \bsnm \undefined \def \bsnm#1{#1}\fi
\ifx \bsuffix \undefined \def \bsuffix#1{#1}\fi
\ifx \bparticle \undefined \def \bparticle#1{#1}\fi
\ifx \barticle \undefined \def \barticle#1{#1}\fi
\bibcommenthead
\ifx \bconfdate \undefined \def \bconfdate #1{#1}\fi
\ifx \botherref \undefined \def \botherref #1{#1}\fi
\ifx \url \undefined \def \url#1{\textsf{#1}}\fi
\ifx \bchapter \undefined \def \bchapter#1{#1}\fi
\ifx \bbook \undefined \def \bbook#1{#1}\fi
\ifx \bcomment \undefined \def \bcomment#1{#1}\fi
\ifx \oauthor \undefined \def \oauthor#1{#1}\fi
\ifx \citeauthoryear \undefined \def \citeauthoryear#1{#1}\fi
\ifx \endbibitem  \undefined \def \endbibitem {}\fi
\ifx \bconflocation  \undefined \def \bconflocation#1{#1}\fi
\ifx \arxivurl  \undefined \def \arxivurl#1{\textsf{#1}}\fi
\csname PreBibitemsHook\endcsname

\bibitem[\protect\citeauthoryear{Ahissar et~al.}{2001}]{ahissar2001speech}
\begin{barticle}
\bauthor{\bsnm{Ahissar}, \binits{E.}},
\bauthor{\bsnm{Nagarajan}, \binits{S.}},
\bauthor{\bsnm{Ahissar}, \binits{M.}},
\bauthor{\bsnm{Protopapas}, \binits{A.}},
\bauthor{\bsnm{Mahncke}, \binits{H.}},
\bauthor{\bsnm{Merzenich}, \binits{M.M.}}:
\batitle{Speech comprehension is correlated with temporal response patterns recorded from auditory cortex}.
\bjtitle{Proceedings of the National Academy of Sciences}
\bvolume{98}(\bissue{23}),
\bfpage{13367}--\blpage{13372}
(\byear{2001})
\end{barticle}
\endbibitem

\bibitem[\protect\citeauthoryear{Donhauser and Baillet}{2020}]{donhauser2020two}
\begin{barticle}
\bauthor{\bsnm{Donhauser}, \binits{P.W.}},
\bauthor{\bsnm{Baillet}, \binits{S.}}:
\batitle{Two distinct neural timescales for predictive speech processing}.
\bjtitle{Neuron}
\bvolume{105}(\bissue{2}),
\bfpage{385}--\blpage{393}
(\byear{2020})
\end{barticle}
\endbibitem

\bibitem[\protect\citeauthoryear{Gillis et~al.}{2021}]{gillis2021neural}
\begin{barticle}
\bauthor{\bsnm{Gillis}, \binits{M.}},
\bauthor{\bsnm{Vanthornhout}, \binits{J.}},
\bauthor{\bsnm{Simon}, \binits{J.Z.}},
\bauthor{\bsnm{Francart}, \binits{T.}},
\bauthor{\bsnm{Brodbeck}, \binits{C.}}:
\batitle{Neural markers of speech comprehension: measuring eeg tracking of linguistic speech representations, controlling the speech acoustics}.
\bjtitle{Journal of Neuroscience}
\bvolume{41}(\bissue{50}),
\bfpage{10316}--\blpage{10329}
(\byear{2021})
\end{barticle}
\endbibitem

\bibitem[\protect\citeauthoryear{Brodbeck et~al.}{2018}]{brodbeck2018rapid}
\begin{barticle}
\bauthor{\bsnm{Brodbeck}, \binits{C.}},
\bauthor{\bsnm{Hong}, \binits{L.E.}},
\bauthor{\bsnm{Simon}, \binits{J.Z.}}:
\batitle{Rapid transformation from auditory to linguistic representations of continuous speech}.
\bjtitle{Current Biology}
\bvolume{28}(\bissue{24}),
\bfpage{3976}--\blpage{3983}
(\byear{2018})
\end{barticle}
\endbibitem

\bibitem[\protect\citeauthoryear{Koskinen et~al.}{2020}]{koskinen2020brain}
\begin{barticle}
\bauthor{\bsnm{Koskinen}, \binits{M.}},
\bauthor{\bsnm{Kurimo}, \binits{M.}},
\bauthor{\bsnm{Gross}, \binits{J.}},
\bauthor{\bsnm{Hyv{\"a}rinen}, \binits{A.}},
\bauthor{\bsnm{Hari}, \binits{R.}}:
\batitle{Brain activity reflects the predictability of word sequences in listened continuous speech}.
\bjtitle{NeuroImage}
\bvolume{219},
\bfpage{116936}
(\byear{2020})
\end{barticle}
\endbibitem

\bibitem[\protect\citeauthoryear{Leonard et~al.}{2024}]{leonard2024large}
\begin{barticle}
\bauthor{\bsnm{Leonard}, \binits{M.K.}},
\bauthor{\bsnm{Gwilliams}, \binits{L.}},
\bauthor{\bsnm{Sellers}, \binits{K.K.}},
\bauthor{\bsnm{Chung}, \binits{J.E.}},
\bauthor{\bsnm{Xu}, \binits{D.}},
\bauthor{\bsnm{Mischler}, \binits{G.}},
\bauthor{\bsnm{Mesgarani}, \binits{N.}},
\bauthor{\bsnm{Welkenhuysen}, \binits{M.}},
\bauthor{\bsnm{Dutta}, \binits{B.}},
\bauthor{\bsnm{Chang}, \binits{E.F.}}:
\batitle{Large-scale single-neuron speech sound encoding across the depth of human cortex}.
\bjtitle{Nature}
\bvolume{626}(\bissue{7999}),
\bfpage{593}--\blpage{602}
(\byear{2024})
\end{barticle}
\endbibitem

\bibitem[\protect\citeauthoryear{Khanna et~al.}{2024}]{khanna2024single}
\begin{barticle}
\bauthor{\bsnm{Khanna}, \binits{A.R.}},
\bauthor{\bsnm{Mu{\~n}oz}, \binits{W.}},
\bauthor{\bsnm{Kim}, \binits{Y.J.}},
\bauthor{\bsnm{Kfir}, \binits{Y.}},
\bauthor{\bsnm{Paulk}, \binits{A.C.}},
\bauthor{\bsnm{Jamali}, \binits{M.}},
\bauthor{\bsnm{Cai}, \binits{J.}},
\bauthor{\bsnm{Mustroph}, \binits{M.L.}},
\bauthor{\bsnm{Caprara}, \binits{I.}},
\bauthor{\bsnm{Hardstone}, \binits{R.}}, \betal:
\batitle{Single-neuronal elements of speech production in humans}.
\bjtitle{Nature}
\bvolume{626}(\bissue{7999}),
\bfpage{603}--\blpage{610}
(\byear{2024})
\end{barticle}
\endbibitem

\bibitem[\protect\citeauthoryear{Clark}{2013}]{clark2013whatever}
\begin{barticle}
\bauthor{\bsnm{Clark}, \binits{A.}}:
\batitle{Whatever next? predictive brains, situated agents, and the future of cognitive science}.
\bjtitle{Behavioral and brain sciences}
\bvolume{36}(\bissue{3}),
\bfpage{181}--\blpage{204}
(\byear{2013})
\end{barticle}
\endbibitem

\bibitem[\protect\citeauthoryear{Schrimpf et~al.}{2021}]{schrimpf2021neural}
\begin{barticle}
\bauthor{\bsnm{Schrimpf}, \binits{M.}},
\bauthor{\bsnm{Blank}, \binits{I.A.}},
\bauthor{\bsnm{Tuckute}, \binits{G.}},
\bauthor{\bsnm{Kauf}, \binits{C.}},
\bauthor{\bsnm{Hosseini}, \binits{E.A.}},
\bauthor{\bsnm{Kanwisher}, \binits{N.}},
\bauthor{\bsnm{Tenenbaum}, \binits{J.B.}},
\bauthor{\bsnm{Fedorenko}, \binits{E.}}:
\batitle{The neural architecture of language: Integrative modeling converges on predictive processing}.
\bjtitle{Proceedings of the National Academy of Sciences}
\bvolume{118}(\bissue{45}),
\bfpage{2105646118}
(\byear{2021})
\end{barticle}
\endbibitem

\bibitem[\protect\citeauthoryear{Heilbron et~al.}{2022}]{heilbron2022hierarchy}
\begin{barticle}
\bauthor{\bsnm{Heilbron}, \binits{M.}},
\bauthor{\bsnm{Armeni}, \binits{K.}},
\bauthor{\bsnm{Schoffelen}, \binits{J.-M.}},
\bauthor{\bsnm{Hagoort}, \binits{P.}},
\bauthor{\bsnm{De~Lange}, \binits{F.P.}}:
\batitle{A hierarchy of linguistic predictions during natural language comprehension}.
\bjtitle{Proceedings of the National Academy of Sciences}
\bvolume{119}(\bissue{32}),
\bfpage{2201968119}
(\byear{2022})
\end{barticle}
\endbibitem

\bibitem[\protect\citeauthoryear{Caucheteux et~al.}{2021}]{caucheteux2021disentangling}
\begin{bchapter}
\bauthor{\bsnm{Caucheteux}, \binits{C.}},
\bauthor{\bsnm{Gramfort}, \binits{A.}},
\bauthor{\bsnm{King}, \binits{J.-R.}}:
\bctitle{Disentangling syntax and semantics in the brain with deep networks}.
In: \bbtitle{International Conference on Machine Learning},
pp. \bfpage{1336}--\blpage{1348}
(\byear{2021}).
\bcomment{PMLR}
\end{bchapter}
\endbibitem

\bibitem[\protect\citeauthoryear{Toneva et~al.}{2022}]{toneva2022combining}
\begin{barticle}
\bauthor{\bsnm{Toneva}, \binits{M.}},
\bauthor{\bsnm{Mitchell}, \binits{T.M.}},
\bauthor{\bsnm{Wehbe}, \binits{L.}}:
\batitle{Combining computational controls with natural text reveals aspects of meaning composition}.
\bjtitle{Nature computational science}
\bvolume{2}(\bissue{11}),
\bfpage{745}--\blpage{757}
(\byear{2022})
\end{barticle}
\endbibitem

\bibitem[\protect\citeauthoryear{Just et~al.}{2010}]{just2010neurosemantic}
\begin{barticle}
\bauthor{\bsnm{Just}, \binits{M.A.}},
\bauthor{\bsnm{Cherkassky}, \binits{V.L.}},
\bauthor{\bsnm{Aryal}, \binits{S.}},
\bauthor{\bsnm{Mitchell}, \binits{T.M.}}:
\batitle{A neurosemantic theory of concrete noun representation based on the underlying brain codes}.
\bjtitle{PloS one}
\bvolume{5}(\bissue{1}),
\bfpage{8622}
(\byear{2010})
\end{barticle}
\endbibitem

\bibitem[\protect\citeauthoryear{Sudre et~al.}{2012}]{sudre2012tracking}
\begin{barticle}
\bauthor{\bsnm{Sudre}, \binits{G.}},
\bauthor{\bsnm{Pomerleau}, \binits{D.}},
\bauthor{\bsnm{Palatucci}, \binits{M.}},
\bauthor{\bsnm{Wehbe}, \binits{L.}},
\bauthor{\bsnm{Fyshe}, \binits{A.}},
\bauthor{\bsnm{Salmelin}, \binits{R.}},
\bauthor{\bsnm{Mitchell}, \binits{T.}}:
\batitle{Tracking neural coding of perceptual and semantic features of concrete nouns}.
\bjtitle{NeuroImage}
\bvolume{62}(\bissue{1}),
\bfpage{451}--\blpage{463}
(\byear{2012})
\end{barticle}
\endbibitem

\bibitem[\protect\citeauthoryear{Anderson et~al.}{2017}]{anderson2017visually}
\begin{barticle}
\bauthor{\bsnm{Anderson}, \binits{A.J.}},
\bauthor{\bsnm{Kiela}, \binits{D.}},
\bauthor{\bsnm{Clark}, \binits{S.}},
\bauthor{\bsnm{Poesio}, \binits{M.}}:
\batitle{Visually grounded and textual semantic models differentially decode brain activity associated with concrete and abstract nouns}.
\bjtitle{Transactions of the Association for Computational Linguistics}
\bvolume{5},
\bfpage{17}--\blpage{30}
(\byear{2017})
\end{barticle}
\endbibitem

\bibitem[\protect\citeauthoryear{Abnar et~al.}{2018}]{abnar2018experiential}
\begin{bchapter}
\bauthor{\bsnm{Abnar}, \binits{S.}},
\bauthor{\bsnm{Ahmed}, \binits{R.}},
\bauthor{\bsnm{Mijnheer}, \binits{M.}},
\bauthor{\bsnm{Zuidema}, \binits{W.}}:
\bctitle{Experiential, distributional and dependency-based word embeddings have complementary roles in decoding brain activity}.
In: \bbtitle{Proceedings of the 8th Workshop on Cognitive Modeling and Computational Linguistics (CMCL 2018)},
pp. \bfpage{57}--\blpage{66}
(\byear{2018})
\end{bchapter}
\endbibitem

\bibitem[\protect\citeauthoryear{Pereira et~al.}{2018}]{pereira2018toward}
\begin{barticle}
\bauthor{\bsnm{Pereira}, \binits{F.}},
\bauthor{\bsnm{Lou}, \binits{B.}},
\bauthor{\bsnm{Pritchett}, \binits{B.}},
\bauthor{\bsnm{Ritter}, \binits{S.}},
\bauthor{\bsnm{Gershman}, \binits{S.J.}},
\bauthor{\bsnm{Kanwisher}, \binits{N.}},
\bauthor{\bsnm{Botvinick}, \binits{M.}},
\bauthor{\bsnm{Fedorenko}, \binits{E.}}:
\batitle{Toward a universal decoder of linguistic meaning from brain activation}.
\bjtitle{Nature communications}
\bvolume{9}(\bissue{1}),
\bfpage{963}
(\byear{2018})
\end{barticle}
\endbibitem

\bibitem[\protect\citeauthoryear{Anderson et~al.}{2017}]{anderson2017predicting}
\begin{barticle}
\bauthor{\bsnm{Anderson}, \binits{A.J.}},
\bauthor{\bsnm{Binder}, \binits{J.R.}},
\bauthor{\bsnm{Fernandino}, \binits{L.}},
\bauthor{\bsnm{Humphries}, \binits{C.J.}},
\bauthor{\bsnm{Conant}, \binits{L.L.}},
\bauthor{\bsnm{Aguilar}, \binits{M.}},
\bauthor{\bsnm{Wang}, \binits{X.}},
\bauthor{\bsnm{Doko}, \binits{D.}},
\bauthor{\bsnm{Raizada}, \binits{R.D.}}:
\batitle{Predicting neural activity patterns associated with sentences using a neurobiologically motivated model of semantic representation}.
\bjtitle{Cerebral Cortex}
\bvolume{27}(\bissue{9}),
\bfpage{4379}--\blpage{4395}
(\byear{2017})
\end{barticle}
\endbibitem

\bibitem[\protect\citeauthoryear{Wang et~al.}{2017}]{wang2017predicting}
\begin{barticle}
\bauthor{\bsnm{Wang}, \binits{J.}},
\bauthor{\bsnm{Cherkassky}, \binits{V.L.}},
\bauthor{\bsnm{Just}, \binits{M.A.}}:
\batitle{Predicting the brain activation pattern associated with the propositional content of a sentence: modeling neural representations of events and states}.
\bjtitle{Human brain mapping}
\bvolume{38}(\bissue{10}),
\bfpage{4865}--\blpage{4881}
(\byear{2017})
\end{barticle}
\endbibitem

\bibitem[\protect\citeauthoryear{Anderson et~al.}{2019}]{anderson2019integrated}
\begin{barticle}
\bauthor{\bsnm{Anderson}, \binits{A.J.}},
\bauthor{\bsnm{Binder}, \binits{J.R.}},
\bauthor{\bsnm{Fernandino}, \binits{L.}},
\bauthor{\bsnm{Humphries}, \binits{C.J.}},
\bauthor{\bsnm{Conant}, \binits{L.L.}},
\bauthor{\bsnm{Raizada}, \binits{R.D.}},
\bauthor{\bsnm{Lin}, \binits{F.}},
\bauthor{\bsnm{Lalor}, \binits{E.C.}}:
\batitle{An integrated neural decoder of linguistic and experiential meaning}.
\bjtitle{Journal of Neuroscience}
\bvolume{39}(\bissue{45}),
\bfpage{8969}--\blpage{8987}
(\byear{2019})
\end{barticle}
\endbibitem

\bibitem[\protect\citeauthoryear{Sun et~al.}{2019}]{sun2019towards}
\begin{bchapter}
\bauthor{\bsnm{Sun}, \binits{J.}},
\bauthor{\bsnm{Wang}, \binits{S.}},
\bauthor{\bsnm{Zhang}, \binits{J.}},
\bauthor{\bsnm{Zong}, \binits{C.}}:
\bctitle{Towards sentence-level brain decoding with distributed representations}.
In: \bbtitle{Proceedings of the AAAI Conference on Artificial Intelligence},
vol. \bseriesno{33},
pp. \bfpage{7047}--\blpage{7054}
(\byear{2019})
\end{bchapter}
\endbibitem

\bibitem[\protect\citeauthoryear{Gauthier and Levy}{2019}]{gauthier2019linking}
\begin{bchapter}
\bauthor{\bsnm{Gauthier}, \binits{J.}},
\bauthor{\bsnm{Levy}, \binits{R.}}:
\bctitle{Linking artificial and human neural representations of language}.
In: \bbtitle{Proceedings of the 2019 Conference on Empirical Methods in Natural Language Processing and the 9th International Joint Conference on Natural Language Processing (EMNLP-IJCNLP)},
pp. \bfpage{529}--\blpage{539}
(\byear{2019})
\end{bchapter}
\endbibitem

\bibitem[\protect\citeauthoryear{Jat et~al.}{2019}]{jat2019relating}
\begin{bchapter}
\bauthor{\bsnm{Jat}, \binits{S.}},
\bauthor{\bsnm{Tang}, \binits{H.}},
\bauthor{\bsnm{Talukdar}, \binits{P.}},
\bauthor{\bsnm{Mitchell}, \binits{T.}}:
\bctitle{Relating simple sentence representations in deep neural networks and the brain}.
In: \bbtitle{Proceedings of the 57th Annual Meeting of the Association for Computational Linguistics},
pp. \bfpage{5137}--\blpage{5154}
(\byear{2019})
\end{bchapter}
\endbibitem

\bibitem[\protect\citeauthoryear{Sun et~al.}{2020}]{sun2020neural}
\begin{barticle}
\bauthor{\bsnm{Sun}, \binits{J.}},
\bauthor{\bsnm{Wang}, \binits{S.}},
\bauthor{\bsnm{Zhang}, \binits{J.}},
\bauthor{\bsnm{Zong}, \binits{C.}}:
\batitle{Neural encoding and decoding with distributed sentence representations}.
\bjtitle{IEEE Transactions on Neural Networks and Learning Systems}
\bvolume{32}(\bissue{2}),
\bfpage{589}--\blpage{603}
(\byear{2020})
\end{barticle}
\endbibitem

\bibitem[\protect\citeauthoryear{Affolter et~al.}{2020}]{affolter2020brain2word}
\begin{botherref}
\oauthor{\bsnm{Affolter}, \binits{N.}},
\oauthor{\bsnm{Egressy}, \binits{B.}},
\oauthor{\bsnm{Pascual}, \binits{D.}},
\oauthor{\bsnm{Wattenhofer}, \binits{R.}}:
Brain2word: decoding brain activity for language generation.
arXiv preprint arXiv:2009.04765
(2020)
\end{botherref}
\endbibitem

\bibitem[\protect\citeauthoryear{Zou et~al.}{2021}]{zou2021towards}
\begin{bchapter}
\bauthor{\bsnm{Zou}, \binits{S.}},
\bauthor{\bsnm{Wang}, \binits{S.}},
\bauthor{\bsnm{Zhang}, \binits{J.}},
\bauthor{\bsnm{Zong}, \binits{C.}}:
\bctitle{Towards brain-to-text generation: Neural decoding with pre-trained encoder-decoder models}.
In: \bbtitle{NeurIPS 2021 AI for Science Workshop}
(\byear{2021})
\end{bchapter}
\endbibitem

\bibitem[\protect\citeauthoryear{Liu and Ayaz}{2018}]{liu2018speech}
\begin{barticle}
\bauthor{\bsnm{Liu}, \binits{Y.}},
\bauthor{\bsnm{Ayaz}, \binits{H.}}:
\batitle{Speech recognition via fnirs based brain signals}.
\bjtitle{Frontiers in neuroscience}
\bvolume{12},
\bfpage{395799}
(\byear{2018})
\end{barticle}
\endbibitem

\bibitem[\protect\citeauthoryear{Moses et~al.}{2019}]{moses2019real}
\begin{barticle}
\bauthor{\bsnm{Moses}, \binits{D.A.}},
\bauthor{\bsnm{Leonard}, \binits{M.K.}},
\bauthor{\bsnm{Makin}, \binits{J.G.}},
\bauthor{\bsnm{Chang}, \binits{E.F.}}:
\batitle{Real-time decoding of question-and-answer speech dialogue using human cortical activity}.
\bjtitle{Nature communications}
\bvolume{10}(\bissue{1}),
\bfpage{3096}
(\byear{2019})
\end{barticle}
\endbibitem

\bibitem[\protect\citeauthoryear{Bollens et~al.}{2022}]{bollens2022learning}
\begin{bchapter}
\bauthor{\bsnm{Bollens}, \binits{L.}},
\bauthor{\bsnm{Francart}, \binits{T.}},
\bauthor{\bsnm{Van~Hamme}, \binits{H.}}:
\bctitle{Learning subject-invariant representations from speech-evoked eeg using variational autoencoders}.
In: \bbtitle{ICASSP 2022-2022 IEEE International Conference on Acoustics, Speech and Signal Processing (ICASSP)},
pp. \bfpage{1256}--\blpage{1260}
(\byear{2022}).
\bcomment{IEEE}
\end{bchapter}
\endbibitem

\bibitem[\protect\citeauthoryear{D{\'e}fossez et~al.}{2023}]{defossez2023decoding}
\begin{barticle}
\bauthor{\bsnm{D{\'e}fossez}, \binits{A.}},
\bauthor{\bsnm{Caucheteux}, \binits{C.}},
\bauthor{\bsnm{Rapin}, \binits{J.}},
\bauthor{\bsnm{Kabeli}, \binits{O.}},
\bauthor{\bsnm{King}, \binits{J.-R.}}:
\batitle{Decoding speech perception from non-invasive brain recordings}.
\bjtitle{Nature Machine Intelligence}
\bvolume{5}(\bissue{10}),
\bfpage{1097}--\blpage{1107}
(\byear{2023})
\end{barticle}
\endbibitem

\bibitem[\protect\citeauthoryear{Vanthornhout et~al.}{2018}]{vanthornhout2018speech}
\begin{barticle}
\bauthor{\bsnm{Vanthornhout}, \binits{J.}},
\bauthor{\bsnm{Decruy}, \binits{L.}},
\bauthor{\bsnm{Wouters}, \binits{J.}},
\bauthor{\bsnm{Simon}, \binits{J.Z.}},
\bauthor{\bsnm{Francart}, \binits{T.}}:
\batitle{Speech intelligibility predicted from neural entrainment of the speech envelope}.
\bjtitle{Journal of the Association for Research in Otolaryngology}
\bvolume{19},
\bfpage{181}--\blpage{191}
(\byear{2018})
\end{barticle}
\endbibitem

\bibitem[\protect\citeauthoryear{Crosse et~al.}{2016}]{crosse2016multivariate}
\begin{barticle}
\bauthor{\bsnm{Crosse}, \binits{M.J.}},
\bauthor{\bsnm{Di~Liberto}, \binits{G.M.}},
\bauthor{\bsnm{Bednar}, \binits{A.}},
\bauthor{\bsnm{Lalor}, \binits{E.C.}}:
\batitle{The multivariate temporal response function (mtrf) toolbox: a matlab toolbox for relating neural signals to continuous stimuli}.
\bjtitle{Frontiers in human neuroscience}
\bvolume{10},
\bfpage{604}
(\byear{2016})
\end{barticle}
\endbibitem

\bibitem[\protect\citeauthoryear{Accou et~al.}{2021}]{accou2021modeling}
\begin{bchapter}
\bauthor{\bsnm{Accou}, \binits{B.}},
\bauthor{\bsnm{Monesi}, \binits{M.J.}},
\bauthor{\bsnm{Montoya}, \binits{J.}},
\bauthor{\bsnm{Francart}, \binits{T.}}, \betal:
\bctitle{Modeling the relationship between acoustic stimulus and eeg with a dilated convolutional neural network}.
In: \bbtitle{2020 28th European Signal Processing Conference (EUSIPCO)},
pp. \bfpage{1175}--\blpage{1179}
(\byear{2021}).
\bcomment{IEEE}
\end{bchapter}
\endbibitem

\bibitem[\protect\citeauthoryear{Thornton et~al.}{2022}]{thornton2022robust}
\begin{barticle}
\bauthor{\bsnm{Thornton}, \binits{M.}},
\bauthor{\bsnm{Mandic}, \binits{D.}},
\bauthor{\bsnm{Reichenbach}, \binits{T.}}:
\batitle{Robust decoding of the speech envelope from eeg recordings through deep neural networks}.
\bjtitle{Journal of neural engineering}
\bvolume{19}(\bissue{4}),
\bfpage{046007}
(\byear{2022})
\end{barticle}
\endbibitem

\bibitem[\protect\citeauthoryear{Xu et~al.}{2022}]{xu2022decoding}
\begin{barticle}
\bauthor{\bsnm{Xu}, \binits{Z.}},
\bauthor{\bsnm{Bai}, \binits{Y.}},
\bauthor{\bsnm{Zhao}, \binits{R.}},
\bauthor{\bsnm{Hu}, \binits{H.}},
\bauthor{\bsnm{Ni}, \binits{G.}},
\bauthor{\bsnm{Ming}, \binits{D.}}:
\batitle{Decoding selective auditory attention with eeg using a transformer model}.
\bjtitle{Methods}
\bvolume{204},
\bfpage{410}--\blpage{417}
(\byear{2022})
\end{barticle}
\endbibitem

\bibitem[\protect\citeauthoryear{Accou et~al.}{2023}]{accou2023decoding}
\begin{barticle}
\bauthor{\bsnm{Accou}, \binits{B.}},
\bauthor{\bsnm{Vanthornhout}, \binits{J.}},
\bauthor{\bsnm{hamme}, \binits{H.V.}},
\bauthor{\bsnm{Francart}, \binits{T.}}:
\batitle{Decoding of the speech envelope from eeg using the vlaai deep neural network}.
\bjtitle{Scientific Reports}
\bvolume{13}(\bissue{1}),
\bfpage{812}
(\byear{2023})
\end{barticle}
\endbibitem

\bibitem[\protect\citeauthoryear{De~Cheveign{\'e} et~al.}{2018}]{de2018decoding}
\begin{barticle}
\bauthor{\bsnm{De~Cheveign{\'e}}, \binits{A.}},
\bauthor{\bsnm{Wong}, \binits{D.D.}},
\bauthor{\bsnm{Di~Liberto}, \binits{G.M.}},
\bauthor{\bsnm{Hjortkj{\ae}r}, \binits{J.}},
\bauthor{\bsnm{Slaney}, \binits{M.}},
\bauthor{\bsnm{Lalor}, \binits{E.}}:
\batitle{Decoding the auditory brain with canonical component analysis}.
\bjtitle{NeuroImage}
\bvolume{172},
\bfpage{206}--\blpage{216}
(\byear{2018})
\end{barticle}
\endbibitem

\bibitem[\protect\citeauthoryear{Zhu et~al.}{2023}]{zhu2023eeg2vec}
\begin{botherref}
\oauthor{\bsnm{Zhu}, \binits{Q.}},
\oauthor{\bsnm{Zhao}, \binits{X.}},
\oauthor{\bsnm{Zhang}, \binits{J.}},
\oauthor{\bsnm{Gu}, \binits{Y.}},
\oauthor{\bsnm{Weng}, \binits{C.}},
\oauthor{\bsnm{Hu}, \binits{Y.}}:
Eeg2vec: Self-supervised electroencephalographic representation learning.
arXiv preprint arXiv:2305.13957
(2023)
\end{botherref}
\endbibitem

\bibitem[\protect\citeauthoryear{de~Taillez et~al.}{2020}]{de2020machine}
\begin{barticle}
\bauthor{\bsnm{Taillez}, \binits{T.}},
\bauthor{\bsnm{Kollmeier}, \binits{B.}},
\bauthor{\bsnm{Meyer}, \binits{B.T.}}:
\batitle{Machine learning for decoding listeners’ attention from electroencephalography evoked by continuous speech}.
\bjtitle{European Journal of Neuroscience}
\bvolume{51}(\bissue{5}),
\bfpage{1234}--\blpage{1241}
(\byear{2020})
\end{barticle}
\endbibitem

\bibitem[\protect\citeauthoryear{Krishna et~al.}{2019}]{krishna2019state}
\begin{botherref}
\oauthor{\bsnm{Krishna}, \binits{G.}},
\oauthor{\bsnm{Han}, \binits{Y.}},
\oauthor{\bsnm{Tran}, \binits{C.}},
\oauthor{\bsnm{Carnahan}, \binits{M.}},
\oauthor{\bsnm{Tewfik}, \binits{A.H.}}:
State-of-the-art speech recognition using eeg and towards decoding of speech spectrum from eeg.
arXiv preprint arXiv:1908.05743
(2019)
\end{botherref}
\endbibitem

\bibitem[\protect\citeauthoryear{Krishna et~al.}{2021}]{krishna2021advancing}
\begin{bchapter}
\bauthor{\bsnm{Krishna}, \binits{G.}},
\bauthor{\bsnm{Tran}, \binits{C.}},
\bauthor{\bsnm{Carnahan}, \binits{M.}},
\bauthor{\bsnm{Tewfik}, \binits{A.H.}}:
\bctitle{Advancing speech synthesis using eeg}.
In: \bbtitle{2021 10th International IEEE/EMBS Conference on Neural Engineering (NER)},
pp. \bfpage{199}--\blpage{204}
(\byear{2021}).
\bcomment{IEEE}
\end{bchapter}
\endbibitem

\bibitem[\protect\citeauthoryear{Petrosyan et~al.}{2021}]{petrosyan2021compact}
\begin{bchapter}
\bauthor{\bsnm{Petrosyan}, \binits{A.}},
\bauthor{\bsnm{Voskoboynikov}, \binits{A.}},
\bauthor{\bsnm{Ossadtchi}, \binits{A.}}:
\bctitle{Compact and interpretable architecture for speech decoding from stereotactic eeg}.
In: \bbtitle{2021 Third International Conference Neurotechnologies and Neurointerfaces (CNN)},
pp. \bfpage{79}--\blpage{82}
(\byear{2021}).
\bcomment{IEEE}
\end{bchapter}
\endbibitem

\bibitem[\protect\citeauthoryear{Wang et~al.}{2020}]{wang2020stimulus}
\begin{bchapter}
\bauthor{\bsnm{Wang}, \binits{R.}},
\bauthor{\bsnm{Chen}, \binits{X.}},
\bauthor{\bsnm{Khalilian-Gourtani}, \binits{A.}},
\bauthor{\bsnm{Chen}, \binits{Z.}},
\bauthor{\bsnm{Yu}, \binits{L.}},
\bauthor{\bsnm{Flinker}, \binits{A.}},
\bauthor{\bsnm{Wang}, \binits{Y.}}:
\bctitle{Stimulus speech decoding from human cortex with generative adversarial network transfer learning}.
In: \bbtitle{2020 IEEE 17th International Symposium on Biomedical Imaging (ISBI)},
pp. \bfpage{390}--\blpage{394}
(\byear{2020}).
\bcomment{IEEE}
\end{bchapter}
\endbibitem

\bibitem[\protect\citeauthoryear{Guo et~al.}{2023}]{guo2023end}
\begin{barticle}
\bauthor{\bsnm{Guo}, \binits{Y.}},
\bauthor{\bsnm{Liu}, \binits{T.}},
\bauthor{\bsnm{Zhang}, \binits{X.}},
\bauthor{\bsnm{Wang}, \binits{A.}},
\bauthor{\bsnm{Wang}, \binits{W.}}:
\batitle{End-to-end translation of human neural activity to speech with a dual--dual generative adversarial network}.
\bjtitle{Knowledge-based systems}
\bvolume{277},
\bfpage{110837}
(\byear{2023})
\end{barticle}
\endbibitem

\bibitem[\protect\citeauthoryear{Senda et~al.}{2024}]{senda2024auditory}
\begin{barticle}
\bauthor{\bsnm{Senda}, \binits{J.}},
\bauthor{\bsnm{Tanaka}, \binits{M.}},
\bauthor{\bsnm{Iijima}, \binits{K.}},
\bauthor{\bsnm{Sugino}, \binits{M.}},
\bauthor{\bsnm{Mori}, \binits{F.}},
\bauthor{\bsnm{Jimbo}, \binits{Y.}},
\bauthor{\bsnm{Iwasaki}, \binits{M.}},
\bauthor{\bsnm{Kotani}, \binits{K.}}:
\batitle{Auditory stimulus reconstruction from ecog with dnn and self-attention modules}.
\bjtitle{Biomedical Signal Processing and Control}
\bvolume{89},
\bfpage{105761}
(\byear{2024})
\end{barticle}
\endbibitem

\bibitem[\protect\citeauthoryear{Akbari et~al.}{2019}]{akbari2019towards}
\begin{barticle}
\bauthor{\bsnm{Akbari}, \binits{H.}},
\bauthor{\bsnm{Khalighinejad}, \binits{B.}},
\bauthor{\bsnm{Herrero}, \binits{J.L.}},
\bauthor{\bsnm{Mehta}, \binits{A.D.}},
\bauthor{\bsnm{Mesgarani}, \binits{N.}}:
\batitle{Towards reconstructing intelligible speech from the human auditory cortex}.
\bjtitle{Scientific reports}
\bvolume{9}(\bissue{1}),
\bfpage{874}
(\byear{2019})
\end{barticle}
\endbibitem

\bibitem[\protect\citeauthoryear{Makin et~al.}{2020}]{makin2020machine}
\begin{barticle}
\bauthor{\bsnm{Makin}, \binits{J.G.}},
\bauthor{\bsnm{Moses}, \binits{D.A.}},
\bauthor{\bsnm{Chang}, \binits{E.F.}}:
\batitle{Machine translation of cortical activity to text with an encoder--decoder framework}.
\bjtitle{Nature neuroscience}
\bvolume{23}(\bissue{4}),
\bfpage{575}--\blpage{582}
(\byear{2020})
\end{barticle}
\endbibitem

\bibitem[\protect\citeauthoryear{Wang and Ji}{2022}]{wang2022open}
\begin{bchapter}
\bauthor{\bsnm{Wang}, \binits{Z.}},
\bauthor{\bsnm{Ji}, \binits{H.}}:
\bctitle{Open vocabulary electroencephalography-to-text decoding and zero-shot sentiment classification}.
In: \bbtitle{Proceedings of the AAAI Conference on Artificial Intelligence},
vol. \bseriesno{36},
pp. \bfpage{5350}--\blpage{5358}
(\byear{2022})
\end{bchapter}
\endbibitem

\bibitem[\protect\citeauthoryear{Xi et~al.}{2023}]{xi2023unicorn}
\begin{bchapter}
\bauthor{\bsnm{Xi}, \binits{N.}},
\bauthor{\bsnm{Zhao}, \binits{S.}},
\bauthor{\bsnm{Wang}, \binits{H.}},
\bauthor{\bsnm{Liu}, \binits{C.}},
\bauthor{\bsnm{Qin}, \binits{B.}},
\bauthor{\bsnm{Liu}, \binits{T.}}:
\bctitle{Unicorn: Unified cognitive signal reconstruction bridging cognitive signals and human language}.
In: \bbtitle{Proceedings of the 61st Annual Meeting of the Association for Computational Linguistics (Volume 1: Long Papers)},
pp. \bfpage{13277}--\blpage{13291}
(\byear{2023})
\end{bchapter}
\endbibitem

\bibitem[\protect\citeauthoryear{Duan et~al.}{2024}]{duan2024dewave}
\begin{botherref}
\oauthor{\bsnm{Duan}, \binits{Y.}},
\oauthor{\bsnm{Chau}, \binits{C.}},
\oauthor{\bsnm{Wang}, \binits{Z.}},
\oauthor{\bsnm{Wang}, \binits{Y.-K.}},
\oauthor{\bsnm{Lin}, \binits{C.-t.}}:
Dewave: Discrete encoding of eeg waves for eeg to text translation.
Advances in Neural Information Processing Systems
\textbf{36}
(2024)
\end{botherref}
\endbibitem

\bibitem[\protect\citeauthoryear{Feng et~al.}{2023}]{feng2023aligning}
\begin{botherref}
\oauthor{\bsnm{Feng}, \binits{X.}},
\oauthor{\bsnm{Feng}, \binits{X.}},
\oauthor{\bsnm{Qin}, \binits{B.}},
\oauthor{\bsnm{Liu}, \binits{T.}}:
Aligning semantic in brain and language: A curriculum contrastive method for electroencephalography-to-text generation.
IEEE Transactions on Neural Systems and Rehabilitation Engineering
(2023)
\end{botherref}
\endbibitem

\bibitem[\protect\citeauthoryear{Tang et~al.}{2023}]{tang2023semantic}
\begin{barticle}
\bauthor{\bsnm{Tang}, \binits{J.}},
\bauthor{\bsnm{LeBel}, \binits{A.}},
\bauthor{\bsnm{Jain}, \binits{S.}},
\bauthor{\bsnm{Huth}, \binits{A.G.}}:
\batitle{Semantic reconstruction of continuous language from non-invasive brain recordings}.
\bjtitle{Nature Neuroscience}
\bvolume{26}(\bissue{5}),
\bfpage{858}--\blpage{866}
(\byear{2023})
\end{barticle}
\endbibitem

\bibitem[\protect\citeauthoryear{Chen et~al.}{2024}]{chen2024openvocabulary}
\begin{botherref}
\oauthor{\bsnm{Chen}, \binits{X.}},
\oauthor{\bsnm{Du}, \binits{C.}},
\oauthor{\bsnm{Liu}, \binits{C.}},
\oauthor{\bsnm{Wang}, \binits{Y.}},
\oauthor{\bsnm{He}, \binits{H.}}:
Open-vocabulary Auditory Neural Decoding Using fMRI-prompted LLM
(2024)
\end{botherref}
\endbibitem

\bibitem[\protect\citeauthoryear{Yang et~al.}{2024a}]{yang2024decode}
\begin{botherref}
\oauthor{\bsnm{Yang}, \binits{Y.}},
\oauthor{\bsnm{Duan}, \binits{Y.}},
\oauthor{\bsnm{Zhang}, \binits{Q.}},
\oauthor{\bsnm{Xu}, \binits{R.}},
\oauthor{\bsnm{Xiong}, \binits{H.}}:
Decode neural signal as speech.
arXiv preprint arXiv:2403.01748
(2024)
\end{botherref}
\endbibitem

\bibitem[\protect\citeauthoryear{Yang et~al.}{2024b}]{yang2024mad}
\begin{botherref}
\oauthor{\bsnm{Yang}, \binits{Y.}},
\oauthor{\bsnm{Jo}, \binits{H.}},
\oauthor{\bsnm{Duan}, \binits{Y.}},
\oauthor{\bsnm{Zhang}, \binits{Q.}},
\oauthor{\bsnm{Zhou}, \binits{J.}},
\oauthor{\bsnm{Lee}, \binits{W.H.}},
\oauthor{\bsnm{Xu}, \binits{R.}},
\oauthor{\bsnm{Xiong}, \binits{H.}}:
MAD: Multi-Alignment MEG-to-Text Decoding
(2024)
\end{botherref}
\endbibitem

\bibitem[\protect\citeauthoryear{Antonello et~al.}{2024}]{antonello2024many}
\begin{botherref}
\oauthor{\bsnm{Antonello}, \binits{R.}},
\oauthor{\bsnm{Sarma}, \binits{N.}},
\oauthor{\bsnm{Tang}, \binits{J.}},
\oauthor{\bsnm{Song}, \binits{J.}},
\oauthor{\bsnm{Huth}, \binits{A.}}:
How many bytes can you take out of brain-to-text decoding?
arXiv preprint arXiv:2405.14055
(2024)
\end{botherref}
\endbibitem

\bibitem[\protect\citeauthoryear{Suppes et~al.}{1997}]{suppes1997brain}
\begin{barticle}
\bauthor{\bsnm{Suppes}, \binits{P.}},
\bauthor{\bsnm{Lu}, \binits{Z.-L.}},
\bauthor{\bsnm{Han}, \binits{B.}}:
\batitle{Brain wave recognition of words}.
\bjtitle{Proceedings of the National Academy of Sciences}
\bvolume{94}(\bissue{26}),
\bfpage{14965}--\blpage{14969}
(\byear{1997})
\end{barticle}
\endbibitem

\bibitem[\protect\citeauthoryear{Suppes et~al.}{1998}]{suppes1998brain}
\begin{barticle}
\bauthor{\bsnm{Suppes}, \binits{P.}},
\bauthor{\bsnm{Han}, \binits{B.}},
\bauthor{\bsnm{Lu}, \binits{Z.-L.}}:
\batitle{Brain-wave recognition of sentences}.
\bjtitle{Proceedings of the National Academy of Sciences}
\bvolume{95}(\bissue{26}),
\bfpage{15861}--\blpage{15866}
(\byear{1998})
\end{barticle}
\endbibitem

\bibitem[\protect\citeauthoryear{D’Zmura et~al.}{2009}]{d2009toward}
\begin{bchapter}
\bauthor{\bsnm{D’Zmura}, \binits{M.}},
\bauthor{\bsnm{Deng}, \binits{S.}},
\bauthor{\bsnm{Lappas}, \binits{T.}},
\bauthor{\bsnm{Thorpe}, \binits{S.}},
\bauthor{\bsnm{Srinivasan}, \binits{R.}}:
\bctitle{Toward eeg sensing of imagined speech}.
In: \bbtitle{Human-Computer Interaction. New Trends: 13th International Conference, HCI International 2009, San Diego, CA, USA, July 19-24, 2009, Proceedings, Part I 13},
pp. \bfpage{40}--\blpage{48}
(\byear{2009}).
\bcomment{Springer}
\end{bchapter}
\endbibitem

\bibitem[\protect\citeauthoryear{Tankus et~al.}{2012}]{tankus2012structured}
\begin{barticle}
\bauthor{\bsnm{Tankus}, \binits{A.}},
\bauthor{\bsnm{Fried}, \binits{I.}},
\bauthor{\bsnm{Shoham}, \binits{S.}}:
\batitle{Structured neuronal encoding and decoding of human speech features}.
\bjtitle{Nature communications}
\bvolume{3}(\bissue{1}),
\bfpage{1015}
(\byear{2012})
\end{barticle}
\endbibitem

\bibitem[\protect\citeauthoryear{DaSalla et~al.}{2009}]{dasalla2009single}
\begin{barticle}
\bauthor{\bsnm{DaSalla}, \binits{C.S.}},
\bauthor{\bsnm{Kambara}, \binits{H.}},
\bauthor{\bsnm{Sato}, \binits{M.}},
\bauthor{\bsnm{Koike}, \binits{Y.}}:
\batitle{Single-trial classification of vowel speech imagery using common spatial patterns}.
\bjtitle{Neural networks}
\bvolume{22}(\bissue{9}),
\bfpage{1334}--\blpage{1339}
(\byear{2009})
\end{barticle}
\endbibitem

\bibitem[\protect\citeauthoryear{Wang et~al.}{2013}]{wang2013analysis}
\begin{barticle}
\bauthor{\bsnm{Wang}, \binits{L.}},
\bauthor{\bsnm{Zhang}, \binits{X.}},
\bauthor{\bsnm{Zhong}, \binits{X.}},
\bauthor{\bsnm{Zhang}, \binits{Y.}}:
\batitle{Analysis and classification of speech imagery eeg for bci}.
\bjtitle{Biomedical signal processing and control}
\bvolume{8}(\bissue{6}),
\bfpage{901}--\blpage{908}
(\byear{2013})
\end{barticle}
\endbibitem

\bibitem[\protect\citeauthoryear{Stavisky et~al.}{2018}]{stavisky2018decoding}
\begin{bchapter}
\bauthor{\bsnm{Stavisky}, \binits{S.D.}},
\bauthor{\bsnm{Rezaii}, \binits{P.}},
\bauthor{\bsnm{Willett}, \binits{F.R.}},
\bauthor{\bsnm{Hochberg}, \binits{L.R.}},
\bauthor{\bsnm{Shenoy}, \binits{K.V.}},
\bauthor{\bsnm{Henderson}, \binits{J.M.}}:
\bctitle{Decoding speech from intracortical multielectrode arrays in dorsal “arm/hand areas” of human motor cortex}.
In: \bbtitle{2018 40th Annual International Conference of the IEEE Engineering in Medicine and Biology Society (EMBC)},
pp. \bfpage{93}--\blpage{97}
(\byear{2018}).
\bcomment{IEEE}
\end{bchapter}
\endbibitem

\bibitem[\protect\citeauthoryear{Pei et~al.}{2011}]{pei2011decoding}
\begin{barticle}
\bauthor{\bsnm{Pei}, \binits{X.}},
\bauthor{\bsnm{Barbour}, \binits{D.L.}},
\bauthor{\bsnm{Leuthardt}, \binits{E.C.}},
\bauthor{\bsnm{Schalk}, \binits{G.}}:
\batitle{Decoding vowels and consonants in spoken and imagined words using electrocorticographic signals in humans}.
\bjtitle{Journal of neural engineering}
\bvolume{8}(\bissue{4}),
\bfpage{046028}
(\byear{2011})
\end{barticle}
\endbibitem

\bibitem[\protect\citeauthoryear{Kellis et~al.}{2010}]{kellis2010decoding}
\begin{barticle}
\bauthor{\bsnm{Kellis}, \binits{S.}},
\bauthor{\bsnm{Miller}, \binits{K.}},
\bauthor{\bsnm{Thomson}, \binits{K.}},
\bauthor{\bsnm{Brown}, \binits{R.}},
\bauthor{\bsnm{House}, \binits{P.}},
\bauthor{\bsnm{Greger}, \binits{B.}}:
\batitle{Decoding spoken words using local field potentials recorded from the cortical surface}.
\bjtitle{Journal of neural engineering}
\bvolume{7}(\bissue{5}),
\bfpage{056007}
(\byear{2010})
\end{barticle}
\endbibitem

\bibitem[\protect\citeauthoryear{Brumberg et~al.}{2011}]{brumberg2011classification}
\begin{barticle}
\bauthor{\bsnm{Brumberg}, \binits{J.S.}},
\bauthor{\bsnm{Wright}, \binits{E.J.}},
\bauthor{\bsnm{Guenther}, \binits{F.H.}},
\bauthor{\bsnm{Kennedy}, \binits{P.R.}}:
\batitle{Classification of intended phoneme production from chronic intracortical microelectrode recordings in speech motor cortex}.
\bjtitle{Frontiers in neuroscience}
\bvolume{5},
\bfpage{7880}
(\byear{2011})
\end{barticle}
\endbibitem

\bibitem[\protect\citeauthoryear{Deng et~al.}{2010}]{deng2010eeg}
\begin{barticle}
\bauthor{\bsnm{Deng}, \binits{S.}},
\bauthor{\bsnm{Srinivasan}, \binits{R.}},
\bauthor{\bsnm{Lappas}, \binits{T.}},
\bauthor{\bsnm{D'Zmura}, \binits{M.}}:
\batitle{Eeg classification of imagined syllable rhythm using hilbert spectrum methods}.
\bjtitle{Journal of neural engineering}
\bvolume{7}(\bissue{4}),
\bfpage{046006}
(\byear{2010})
\end{barticle}
\endbibitem

\bibitem[\protect\citeauthoryear{Kim et~al.}{2014}]{kim2014eeg}
\begin{barticle}
\bauthor{\bsnm{Kim}, \binits{J.}},
\bauthor{\bsnm{Lee}, \binits{S.-K.}},
\bauthor{\bsnm{Lee}, \binits{B.}}:
\batitle{Eeg classification in a single-trial basis for vowel speech perception using multivariate empirical mode decomposition}.
\bjtitle{Journal of neural engineering}
\bvolume{11}(\bissue{3}),
\bfpage{036010}
(\byear{2014})
\end{barticle}
\endbibitem

\bibitem[\protect\citeauthoryear{Mugler et~al.}{2014}]{mugler2014direct}
\begin{barticle}
\bauthor{\bsnm{Mugler}, \binits{E.M.}},
\bauthor{\bsnm{Patton}, \binits{J.L.}},
\bauthor{\bsnm{Flint}, \binits{R.D.}},
\bauthor{\bsnm{Wright}, \binits{Z.A.}},
\bauthor{\bsnm{Schuele}, \binits{S.U.}},
\bauthor{\bsnm{Rosenow}, \binits{J.}},
\bauthor{\bsnm{Shih}, \binits{J.J.}},
\bauthor{\bsnm{Krusienski}, \binits{D.J.}},
\bauthor{\bsnm{Slutzky}, \binits{M.W.}}:
\batitle{Direct classification of all american english phonemes using signals from functional speech motor cortex}.
\bjtitle{Journal of neural engineering}
\bvolume{11}(\bissue{3}),
\bfpage{035015}
(\byear{2014})
\end{barticle}
\endbibitem

\bibitem[\protect\citeauthoryear{Mohanchandra and Saha}{2016}]{mohanchandra2016communication}
\begin{barticle}
\bauthor{\bsnm{Mohanchandra}, \binits{K.}},
\bauthor{\bsnm{Saha}, \binits{S.}}:
\batitle{A communication paradigm using subvocalized speech: translating brain signals into speech}.
\bjtitle{Augmented Human Research}
\bvolume{1}(\bissue{1}),
\bfpage{3}
(\byear{2016})
\end{barticle}
\endbibitem

\bibitem[\protect\citeauthoryear{Martin et~al.}{2016}]{martin2016word}
\begin{barticle}
\bauthor{\bsnm{Martin}, \binits{S.}},
\bauthor{\bsnm{Brunner}, \binits{P.}},
\bauthor{\bsnm{Iturrate}, \binits{I.}},
\bauthor{\bsnm{Mill{\'a}n}, \binits{J.d.R.}},
\bauthor{\bsnm{Schalk}, \binits{G.}},
\bauthor{\bsnm{Knight}, \binits{R.T.}},
\bauthor{\bsnm{Pasley}, \binits{B.N.}}:
\batitle{Word pair classification during imagined speech using direct brain recordings}.
\bjtitle{Scientific reports}
\bvolume{6}(\bissue{1}),
\bfpage{25803}
(\byear{2016})
\end{barticle}
\endbibitem

\bibitem[\protect\citeauthoryear{Nguyen et~al.}{2017}]{nguyen2017inferring}
\begin{barticle}
\bauthor{\bsnm{Nguyen}, \binits{C.H.}},
\bauthor{\bsnm{Karavas}, \binits{G.K.}},
\bauthor{\bsnm{Artemiadis}, \binits{P.}}:
\batitle{Inferring imagined speech using eeg signals: a new approach using riemannian manifold features}.
\bjtitle{Journal of neural engineering}
\bvolume{15}(\bissue{1}),
\bfpage{016002}
(\byear{2017})
\end{barticle}
\endbibitem

\bibitem[\protect\citeauthoryear{Gonz{\'a}lez-Casta{\~n}eda et~al.}{2017}]{gonzalez2017sonification}
\begin{barticle}
\bauthor{\bsnm{Gonz{\'a}lez-Casta{\~n}eda}, \binits{E.F.}},
\bauthor{\bsnm{Torres-Garc{\'\i}a}, \binits{A.A.}},
\bauthor{\bsnm{Reyes-Garc{\'\i}a}, \binits{C.A.}},
\bauthor{\bsnm{Villase{\~n}or-Pineda}, \binits{L.}}:
\batitle{Sonification and textification: Proposing methods for classifying unspoken words from eeg signals}.
\bjtitle{Biomedical Signal Processing and Control}
\bvolume{37},
\bfpage{82}--\blpage{91}
(\byear{2017})
\end{barticle}
\endbibitem

\bibitem[\protect\citeauthoryear{Salama et~al.}{2014}]{salama2014recognition}
\begin{bchapter}
\bauthor{\bsnm{Salama}, \binits{M.}},
\bauthor{\bsnm{ElSherif}, \binits{L.}},
\bauthor{\bsnm{Lashin}, \binits{H.}},
\bauthor{\bsnm{Gamal}, \binits{T.}}:
\bctitle{Recognition of unspoken words using electrode electroencephalograhic signals}.
In: \bbtitle{The Sixth International Conference on Advanced Cognitive Technologies and Applications},
pp. \bfpage{51}--\blpage{5}
(\byear{2014})
\end{bchapter}
\endbibitem

\bibitem[\protect\citeauthoryear{Zhao and Rudzicz}{2015}]{zhao2015classifying}
\begin{bchapter}
\bauthor{\bsnm{Zhao}, \binits{S.}},
\bauthor{\bsnm{Rudzicz}, \binits{F.}}:
\bctitle{Classifying phonological categories in imagined and articulated speech}.
In: \bbtitle{2015 IEEE International Conference on Acoustics, Speech and Signal Processing (ICASSP)},
pp. \bfpage{992}--\blpage{996}
(\byear{2015}).
\bcomment{IEEE}
\end{bchapter}
\endbibitem

\bibitem[\protect\citeauthoryear{Saha and Fels}{2019}]{saha2019hierarchical}
\begin{bchapter}
\bauthor{\bsnm{Saha}, \binits{P.}},
\bauthor{\bsnm{Fels}, \binits{S.}}:
\bctitle{Hierarchical deep feature learning for decoding imagined speech from eeg}.
In: \bbtitle{Proceedings of the AAAI Conference on Artificial Intelligence},
vol. \bseriesno{33},
pp. \bfpage{10019}--\blpage{10020}
(\byear{2019})
\end{bchapter}
\endbibitem

\bibitem[\protect\citeauthoryear{Dash et~al.}{2020}]{dash2020decoding}
\begin{barticle}
\bauthor{\bsnm{Dash}, \binits{D.}},
\bauthor{\bsnm{Ferrari}, \binits{P.}},
\bauthor{\bsnm{Wang}, \binits{J.}}:
\batitle{Decoding imagined and spoken phrases from non-invasive neural (meg) signals}.
\bjtitle{Frontiers in neuroscience}
\bvolume{14},
\bfpage{490970}
(\byear{2020})
\end{barticle}
\endbibitem

\bibitem[\protect\citeauthoryear{Herff et~al.}{2015}]{herff2015brain}
\begin{barticle}
\bauthor{\bsnm{Herff}, \binits{C.}},
\bauthor{\bsnm{Heger}, \binits{D.}},
\bauthor{\bsnm{De~Pesters}, \binits{A.}},
\bauthor{\bsnm{Telaar}, \binits{D.}},
\bauthor{\bsnm{Brunner}, \binits{P.}},
\bauthor{\bsnm{Schalk}, \binits{G.}},
\bauthor{\bsnm{Schultz}, \binits{T.}}:
\batitle{Brain-to-text: decoding spoken phrases from phone representations in the brain}.
\bjtitle{Frontiers in neuroscience}
\bvolume{8},
\bfpage{141498}
(\byear{2015})
\end{barticle}
\endbibitem

\bibitem[\protect\citeauthoryear{Moses et~al.}{2016}]{moses2016neural}
\begin{barticle}
\bauthor{\bsnm{Moses}, \binits{D.A.}},
\bauthor{\bsnm{Mesgarani}, \binits{N.}},
\bauthor{\bsnm{Leonard}, \binits{M.K.}},
\bauthor{\bsnm{Chang}, \binits{E.F.}}:
\batitle{Neural speech recognition: continuous phoneme decoding using spatiotemporal representations of human cortical activity}.
\bjtitle{Journal of neural engineering}
\bvolume{13}(\bissue{5}),
\bfpage{056004}
(\byear{2016})
\end{barticle}
\endbibitem

\bibitem[\protect\citeauthoryear{Willett et~al.}{2023}]{willett2023high}
\begin{barticle}
\bauthor{\bsnm{Willett}, \binits{F.R.}},
\bauthor{\bsnm{Kunz}, \binits{E.M.}},
\bauthor{\bsnm{Fan}, \binits{C.}},
\bauthor{\bsnm{Avansino}, \binits{D.T.}},
\bauthor{\bsnm{Wilson}, \binits{G.H.}},
\bauthor{\bsnm{Choi}, \binits{E.Y.}},
\bauthor{\bsnm{Kamdar}, \binits{F.}},
\bauthor{\bsnm{Glasser}, \binits{M.F.}},
\bauthor{\bsnm{Hochberg}, \binits{L.R.}},
\bauthor{\bsnm{Druckmann}, \binits{S.}}, \betal:
\batitle{A high-performance speech neuroprosthesis}.
\bjtitle{Nature}
\bvolume{620}(\bissue{7976}),
\bfpage{1031}--\blpage{1036}
(\byear{2023})
\end{barticle}
\endbibitem

\bibitem[\protect\citeauthoryear{Metzger et~al.}{2023}]{metzger2023high}
\begin{barticle}
\bauthor{\bsnm{Metzger}, \binits{S.L.}},
\bauthor{\bsnm{Littlejohn}, \binits{K.T.}},
\bauthor{\bsnm{Silva}, \binits{A.B.}},
\bauthor{\bsnm{Moses}, \binits{D.A.}},
\bauthor{\bsnm{Seaton}, \binits{M.P.}},
\bauthor{\bsnm{Wang}, \binits{R.}},
\bauthor{\bsnm{Dougherty}, \binits{M.E.}},
\bauthor{\bsnm{Liu}, \binits{J.R.}},
\bauthor{\bsnm{Wu}, \binits{P.}},
\bauthor{\bsnm{Berger}, \binits{M.A.}}, \betal:
\batitle{A high-performance neuroprosthesis for speech decoding and avatar control}.
\bjtitle{Nature}
\bvolume{620}(\bissue{7976}),
\bfpage{1037}--\blpage{1046}
(\byear{2023})
\end{barticle}
\endbibitem

\bibitem[\protect\citeauthoryear{Feng et~al.}{2024}]{feng2024towards}
\begin{botherref}
\oauthor{\bsnm{Feng}, \binits{S.}},
\oauthor{\bsnm{Liu}, \binits{H.}},
\oauthor{\bsnm{Wang}, \binits{Y.}},
\oauthor{\bsnm{Wang}, \binits{Y.}}:
Towards an end-to-end framework for invasive brain signal decoding with large language models.
arXiv preprint arXiv:2406.11568
(2024)
\end{botherref}
\endbibitem

\bibitem[\protect\citeauthoryear{Yuan and Makin}{2024}]{yuan2024improving}
\begin{botherref}
\oauthor{\bsnm{Yuan}, \binits{B.A.}},
\oauthor{\bsnm{Makin}, \binits{J.G.}}:
Improving speech decoding from ecog with self-supervised pretraining.
arXiv preprint arXiv:2405.18639
(2024)
\end{botherref}
\endbibitem

\bibitem[\protect\citeauthoryear{Herff et~al.}{2019}]{herff2019generating}
\begin{barticle}
\bauthor{\bsnm{Herff}, \binits{C.}},
\bauthor{\bsnm{Diener}, \binits{L.}},
\bauthor{\bsnm{Angrick}, \binits{M.}},
\bauthor{\bsnm{Mugler}, \binits{E.}},
\bauthor{\bsnm{Tate}, \binits{M.C.}},
\bauthor{\bsnm{Goldrick}, \binits{M.A.}},
\bauthor{\bsnm{Krusienski}, \binits{D.J.}},
\bauthor{\bsnm{Slutzky}, \binits{M.W.}},
\bauthor{\bsnm{Schultz}, \binits{T.}}:
\batitle{Generating natural, intelligible speech from brain activity in motor, premotor, and inferior frontal cortices}.
\bjtitle{Frontiers in neuroscience}
\bvolume{13},
\bfpage{469935}
(\byear{2019})
\end{barticle}
\endbibitem

\bibitem[\protect\citeauthoryear{Guenther et~al.}{2009}]{guenther2009wireless}
\begin{barticle}
\bauthor{\bsnm{Guenther}, \binits{F.H.}},
\bauthor{\bsnm{Brumberg}, \binits{J.S.}},
\bauthor{\bsnm{Wright}, \binits{E.J.}},
\bauthor{\bsnm{Nieto-Castanon}, \binits{A.}},
\bauthor{\bsnm{Tourville}, \binits{J.A.}},
\bauthor{\bsnm{Panko}, \binits{M.}},
\bauthor{\bsnm{Law}, \binits{R.}},
\bauthor{\bsnm{Siebert}, \binits{S.A.}},
\bauthor{\bsnm{Bartels}, \binits{J.L.}},
\bauthor{\bsnm{Andreasen}, \binits{D.S.}}, \betal:
\batitle{A wireless brain-machine interface for real-time speech synthesis}.
\bjtitle{PloS one}
\bvolume{4}(\bissue{12}),
\bfpage{8218}
(\byear{2009})
\end{barticle}
\endbibitem

\bibitem[\protect\citeauthoryear{Angrick et~al.}{2019}]{angrick2019speech}
\begin{barticle}
\bauthor{\bsnm{Angrick}, \binits{M.}},
\bauthor{\bsnm{Herff}, \binits{C.}},
\bauthor{\bsnm{Mugler}, \binits{E.}},
\bauthor{\bsnm{Tate}, \binits{M.C.}},
\bauthor{\bsnm{Slutzky}, \binits{M.W.}},
\bauthor{\bsnm{Krusienski}, \binits{D.J.}},
\bauthor{\bsnm{Schultz}, \binits{T.}}:
\batitle{Speech synthesis from ecog using densely connected 3d convolutional neural networks}.
\bjtitle{Journal of neural engineering}
\bvolume{16}(\bissue{3}),
\bfpage{036019}
(\byear{2019})
\end{barticle}
\endbibitem

\bibitem[\protect\citeauthoryear{Chen et~al.}{2024}]{chen2024neural}
\begin{botherref}
\oauthor{\bsnm{Chen}, \binits{X.}},
\oauthor{\bsnm{Wang}, \binits{R.}},
\oauthor{\bsnm{Khalilian-Gourtani}, \binits{A.}},
\oauthor{\bsnm{Yu}, \binits{L.}},
\oauthor{\bsnm{Dugan}, \binits{P.}},
\oauthor{\bsnm{Friedman}, \binits{D.}},
\oauthor{\bsnm{Doyle}, \binits{W.}},
\oauthor{\bsnm{Devinsky}, \binits{O.}},
\oauthor{\bsnm{Wang}, \binits{Y.}},
\oauthor{\bsnm{Flinker}, \binits{A.}}:
A neural speech decoding framework leveraging deep learning and speech synthesis.
Nature Machine Intelligence,
1--14
(2024)
\end{botherref}
\endbibitem

\bibitem[\protect\citeauthoryear{Bouchard et~al.}{2013}]{bouchard2013functional}
\begin{barticle}
\bauthor{\bsnm{Bouchard}, \binits{K.E.}},
\bauthor{\bsnm{Mesgarani}, \binits{N.}},
\bauthor{\bsnm{Johnson}, \binits{K.}},
\bauthor{\bsnm{Chang}, \binits{E.F.}}:
\batitle{Functional organization of human sensorimotor cortex for speech articulation}.
\bjtitle{Nature}
\bvolume{495}(\bissue{7441}),
\bfpage{327}--\blpage{332}
(\byear{2013})
\end{barticle}
\endbibitem

\bibitem[\protect\citeauthoryear{Bocquelet et~al.}{2014}]{bocquelet2014robust}
\begin{bchapter}
\bauthor{\bsnm{Bocquelet}, \binits{F.}},
\bauthor{\bsnm{Hueber}, \binits{T.}},
\bauthor{\bsnm{Girin}, \binits{L.}},
\bauthor{\bsnm{Badin}, \binits{P.}},
\bauthor{\bsnm{Yvert}, \binits{B.}}:
\bctitle{Robust articulatory speech synthesis using deep neural networks for bci applications}.
In: \bbtitle{Interspeech 2014-15th Annual Conference of the International Speech Communication Association}
(\byear{2014})
\end{bchapter}
\endbibitem

\bibitem[\protect\citeauthoryear{Abnar et~al.}{2019}]{abnar2019blackbox}
\begin{bchapter}
\bauthor{\bsnm{Abnar}, \binits{S.}},
\bauthor{\bsnm{Beinborn}, \binits{L.}},
\bauthor{\bsnm{Choenni}, \binits{R.}},
\bauthor{\bsnm{Zuidema}, \binits{W.}}:
\bctitle{Blackbox meets blackbox: Representational similarity \& stability analysis of neural language models and brains}.
In: \bbtitle{Proceedings of the 2019 ACL Workshop BlackboxNLP: Analyzing and Interpreting Neural Networks for NLP},
pp. \bfpage{191}--\blpage{203}
(\byear{2019})
\end{bchapter}
\endbibitem

\bibitem[\protect\citeauthoryear{Goutman et~al.}{2023}]{goutman2023amyotrophic}
\begin{barticle}
\bauthor{\bsnm{Goutman}, \binits{S.A.}},
\bauthor{\bsnm{Savelieff}, \binits{M.G.}},
\bauthor{\bsnm{Jang}, \binits{D.-G.}},
\bauthor{\bsnm{Hur}, \binits{J.}},
\bauthor{\bsnm{Feldman}, \binits{E.L.}}:
\batitle{The amyotrophic lateral sclerosis exposome: Recent advances and future directions}.
\bjtitle{Nature Reviews Neurology}
\bvolume{19}(\bissue{10}),
\bfpage{617}--\blpage{634}
(\byear{2023})
\end{barticle}
\endbibitem

\bibitem[\protect\citeauthoryear{Di~Liberto et~al.}{2023}]{di2023emergence}
\begin{barticle}
\bauthor{\bsnm{Di~Liberto}, \binits{G.M.}},
\bauthor{\bsnm{Attaheri}, \binits{A.}},
\bauthor{\bsnm{Cantisani}, \binits{G.}},
\bauthor{\bsnm{Reilly}, \binits{R.B.}},
\bauthor{\bsnm{N{\'\i}~Choisdealbha}, \binits{{\'A}.}},
\bauthor{\bsnm{Rocha}, \binits{S.}},
\bauthor{\bsnm{Brusini}, \binits{P.}},
\bauthor{\bsnm{Goswami}, \binits{U.}}:
\batitle{Emergence of the cortical encoding of phonetic features in the first year of life}.
\bjtitle{Nature communications}
\bvolume{14}(\bissue{1}),
\bfpage{7789}
(\byear{2023})
\end{barticle}
\endbibitem

\bibitem[\protect\citeauthoryear{Broderick et~al.}{2018}]{broderick2018electrophysiological}
\begin{barticle}
\bauthor{\bsnm{Broderick}, \binits{M.P.}},
\bauthor{\bsnm{Anderson}, \binits{A.J.}},
\bauthor{\bsnm{Di~Liberto}, \binits{G.M.}},
\bauthor{\bsnm{Crosse}, \binits{M.J.}},
\bauthor{\bsnm{Lalor}, \binits{E.C.}}:
\batitle{Electrophysiological correlates of semantic dissimilarity reflect the comprehension of natural, narrative speech}.
\bjtitle{Current Biology}
\bvolume{28}(\bissue{5}),
\bfpage{803}--\blpage{809}
(\byear{2018})
\end{barticle}
\endbibitem

\bibitem[\protect\citeauthoryear{Antonello et~al.}{2024}]{antonello2024scaling}
\begin{botherref}
\oauthor{\bsnm{Antonello}, \binits{R.}},
\oauthor{\bsnm{Vaidya}, \binits{A.}},
\oauthor{\bsnm{Huth}, \binits{A.}}:
Scaling laws for language encoding models in fmri.
Advances in Neural Information Processing Systems
\textbf{36}
(2024)
\end{botherref}
\endbibitem

\bibitem[\protect\citeauthoryear{Lin et~al.}{2024}]{lin2024selecting}
\begin{botherref}
\oauthor{\bsnm{Lin}, \binits{H.}},
\oauthor{\bsnm{Huang}, \binits{B.}},
\oauthor{\bsnm{Ye}, \binits{H.}},
\oauthor{\bsnm{Chen}, \binits{Q.}},
\oauthor{\bsnm{Wang}, \binits{Z.}},
\oauthor{\bsnm{Li}, \binits{S.}},
\oauthor{\bsnm{Ma}, \binits{J.}},
\oauthor{\bsnm{Wan}, \binits{X.}},
\oauthor{\bsnm{Zou}, \binits{J.}},
\oauthor{\bsnm{Liang}, \binits{Y.}}:
Selecting large language model to fine-tune via rectified scaling law.
arXiv preprint arXiv:2402.02314
(2024)
\end{botherref}
\endbibitem

\bibitem[\protect\citeauthoryear{Whittington et~al.}{2021}]{whittington2021relating}
\begin{bchapter}
\bauthor{\bsnm{Whittington}, \binits{J.C.}},
\bauthor{\bsnm{Warren}, \binits{J.}},
\bauthor{\bsnm{Behrens}, \binits{T.E.}}:
\bctitle{Relating transformers to models and neural representations of the hippocampal formation}.
In: \bbtitle{International Conference on Learning Representations}
(\byear{2021})
\end{bchapter}
\endbibitem

\bibitem[\protect\citeauthoryear{Liu et~al.}{2023}]{liu2023coupling}
\begin{bchapter}
\bauthor{\bsnm{Liu}, \binits{X.}},
\bauthor{\bsnm{Zhou}, \binits{M.}},
\bauthor{\bsnm{Shi}, \binits{G.}},
\bauthor{\bsnm{Du}, \binits{Y.}},
\bauthor{\bsnm{Zhao}, \binits{L.}},
\bauthor{\bsnm{Wu}, \binits{Z.}},
\bauthor{\bsnm{Liu}, \binits{D.}},
\bauthor{\bsnm{Liu}, \binits{T.}},
\bauthor{\bsnm{Hu}, \binits{X.}}:
\bctitle{Coupling artificial neurons in bert and biological neurons in the human brain}.
In: \bbtitle{Proceedings of the AAAI Conference on Artificial Intelligence},
vol. \bseriesno{37},
pp. \bfpage{8888}--\blpage{8896}
(\byear{2023})
\end{bchapter}
\endbibitem

\bibitem[\protect\citeauthoryear{Papineni et~al.}{2002}]{papineni2002bleu}
\begin{bchapter}
\bauthor{\bsnm{Papineni}, \binits{K.}},
\bauthor{\bsnm{Roukos}, \binits{S.}},
\bauthor{\bsnm{Ward}, \binits{T.}},
\bauthor{\bsnm{Zhu}, \binits{W.-J.}}:
\bctitle{Bleu: a method for automatic evaluation of machine translation}.
In: \bbtitle{Proceedings of the 40th Annual Meeting of the Association for Computational Linguistics},
pp. \bfpage{311}--\blpage{318}
(\byear{2002})
\end{bchapter}
\endbibitem

\bibitem[\protect\citeauthoryear{Zhang et~al.}{2019}]{zhang2019bertscore}
\begin{bchapter}
\bauthor{\bsnm{Zhang}, \binits{T.}},
\bauthor{\bsnm{Kishore}, \binits{V.}},
\bauthor{\bsnm{Wu}, \binits{F.}},
\bauthor{\bsnm{Weinberger}, \binits{K.Q.}},
\bauthor{\bsnm{Artzi}, \binits{Y.}}:
\bctitle{Bertscore: Evaluating text generation with bert}.
In: \bbtitle{International Conference on Learning Representations}
(\byear{2019})
\end{bchapter}
\endbibitem

\bibitem[\protect\citeauthoryear{Devlin et~al.}{2018}]{devlin2018bert}
\begin{botherref}
\oauthor{\bsnm{Devlin}, \binits{J.}},
\oauthor{\bsnm{Chang}, \binits{M.-W.}},
\oauthor{\bsnm{Lee}, \binits{K.}},
\oauthor{\bsnm{Toutanova}, \binits{K.}}:
Bert: Pre-training of deep bidirectional transformers for language understanding.
arXiv preprint arXiv:1810.04805
(2018)
\end{botherref}
\endbibitem

\bibitem[\protect\citeauthoryear{Taal et~al.}{2010}]{taal2010short}
\begin{bchapter}
\bauthor{\bsnm{Taal}, \binits{C.H.}},
\bauthor{\bsnm{Hendriks}, \binits{R.C.}},
\bauthor{\bsnm{Heusdens}, \binits{R.}},
\bauthor{\bsnm{Jensen}, \binits{J.}}:
\bctitle{A short-time objective intelligibility measure for time-frequency weighted noisy speech}.
In: \bbtitle{2010 IEEE International Conference on Acoustics, Speech and Signal Processing},
pp. \bfpage{4214}--\blpage{4217}
(\byear{2010}).
\bcomment{IEEE}
\end{bchapter}
\endbibitem

\bibitem[\protect\citeauthoryear{Kubichek}{1993}]{kubichek1993mel}
\begin{bchapter}
\bauthor{\bsnm{Kubichek}, \binits{R.}}:
\bctitle{Mel-cepstral distance measure for objective speech quality assessment}.
In: \bbtitle{Proceedings of IEEE Pacific Rim Conference on Communications Computers and Signal Processing},
vol. \bseriesno{1},
pp. \bfpage{125}--\blpage{128}
(\byear{1993}).
\bcomment{IEEE}
\end{bchapter}
\endbibitem

\bibitem[\protect\citeauthoryear{Wehbe et~al.}{2014}]{wehbe2014simultaneously}
\begin{barticle}
\bauthor{\bsnm{Wehbe}, \binits{L.}},
\bauthor{\bsnm{Murphy}, \binits{B.}},
\bauthor{\bsnm{Talukdar}, \binits{P.}},
\bauthor{\bsnm{Fyshe}, \binits{A.}},
\bauthor{\bsnm{Ramdas}, \binits{A.}},
\bauthor{\bsnm{Mitchell}, \binits{T.}}:
\batitle{Simultaneously uncovering the patterns of brain regions involved in different story reading subprocesses}.
\bjtitle{PloS one}
\bvolume{9}(\bissue{11}),
\bfpage{112575}
(\byear{2014})
\end{barticle}
\endbibitem

\bibitem[\protect\citeauthoryear{Li et~al.}{2023}]{li2023dissecting}
\begin{barticle}
\bauthor{\bsnm{Li}, \binits{Y.}},
\bauthor{\bsnm{Anumanchipalli}, \binits{G.K.}},
\bauthor{\bsnm{Mohamed}, \binits{A.}},
\bauthor{\bsnm{Chen}, \binits{P.}},
\bauthor{\bsnm{Carney}, \binits{L.H.}},
\bauthor{\bsnm{Lu}, \binits{J.}},
\bauthor{\bsnm{Wu}, \binits{J.}},
\bauthor{\bsnm{Chang}, \binits{E.F.}}:
\batitle{Dissecting neural computations in the human auditory pathway using deep neural networks for speech}.
\bjtitle{Nature Neuroscience}
\bvolume{26}(\bissue{12}),
\bfpage{2213}--\blpage{2225}
(\byear{2023})
\end{barticle}
\endbibitem

\bibitem[\protect\citeauthoryear{Radford et~al.}{2021}]{radford2021learning}
\begin{bchapter}
\bauthor{\bsnm{Radford}, \binits{A.}},
\bauthor{\bsnm{Kim}, \binits{J.W.}},
\bauthor{\bsnm{Hallacy}, \binits{C.}},
\bauthor{\bsnm{Ramesh}, \binits{A.}},
\bauthor{\bsnm{Goh}, \binits{G.}},
\bauthor{\bsnm{Agarwal}, \binits{S.}},
\bauthor{\bsnm{Sastry}, \binits{G.}},
\bauthor{\bsnm{Askell}, \binits{A.}},
\bauthor{\bsnm{Mishkin}, \binits{P.}},
\bauthor{\bsnm{Clark}, \binits{J.}}, \betal:
\bctitle{Learning transferable visual models from natural language supervision}.
In: \bbtitle{International Conference on Machine Learning},
pp. \bfpage{8748}--\blpage{8763}
(\byear{2021}).
\bcomment{PMLR}
\end{bchapter}
\endbibitem

\bibitem[\protect\citeauthoryear{Baevski et~al.}{2020}]{baevski2020wav2vec}
\begin{barticle}
\bauthor{\bsnm{Baevski}, \binits{A.}},
\bauthor{\bsnm{Zhou}, \binits{Y.}},
\bauthor{\bsnm{Mohamed}, \binits{A.}},
\bauthor{\bsnm{Auli}, \binits{M.}}:
\batitle{wav2vec 2.0: A framework for self-supervised learning of speech representations}.
\bjtitle{Advances in neural information processing systems}
\bvolume{33},
\bfpage{12449}--\blpage{12460}
(\byear{2020})
\end{barticle}
\endbibitem

\bibitem[\protect\citeauthoryear{Aiken and Picton}{2008}]{aiken2008human}
\begin{barticle}
\bauthor{\bsnm{Aiken}, \binits{S.J.}},
\bauthor{\bsnm{Picton}, \binits{T.W.}}:
\batitle{Human cortical responses to the speech envelope}.
\bjtitle{Ear and hearing}
\bvolume{29}(\bissue{2}),
\bfpage{139}--\blpage{157}
(\byear{2008})
\end{barticle}
\endbibitem

\bibitem[\protect\citeauthoryear{Ding and Simon}{2014}]{ding2014cortical}
\begin{barticle}
\bauthor{\bsnm{Ding}, \binits{N.}},
\bauthor{\bsnm{Simon}, \binits{J.Z.}}:
\batitle{Cortical entrainment to continuous speech: functional roles and interpretations}.
\bjtitle{Frontiers in human neuroscience}
\bvolume{8},
\bfpage{311}
(\byear{2014})
\end{barticle}
\endbibitem

\bibitem[\protect\citeauthoryear{De~Clercq et~al.}{2023}]{de2023beyond}
\begin{barticle}
\bauthor{\bsnm{De~Clercq}, \binits{P.}},
\bauthor{\bsnm{Vanthornhout}, \binits{J.}},
\bauthor{\bsnm{Vandermosten}, \binits{M.}},
\bauthor{\bsnm{Francart}, \binits{T.}}:
\batitle{Beyond linear neural envelope tracking: a mutual information approach}.
\bjtitle{Journal of Neural Engineering}
\bvolume{20}(\bissue{2}),
\bfpage{026007}
(\byear{2023})
\end{barticle}
\endbibitem

\bibitem[\protect\citeauthoryear{Pasley et~al.}{2012}]{pasley2012reconstructing}
\begin{barticle}
\bauthor{\bsnm{Pasley}, \binits{B.N.}},
\bauthor{\bsnm{David}, \binits{S.V.}},
\bauthor{\bsnm{Mesgarani}, \binits{N.}},
\bauthor{\bsnm{Flinker}, \binits{A.}},
\bauthor{\bsnm{Shamma}, \binits{S.A.}},
\bauthor{\bsnm{Crone}, \binits{N.E.}},
\bauthor{\bsnm{Knight}, \binits{R.T.}},
\bauthor{\bsnm{Chang}, \binits{E.F.}}:
\batitle{Reconstructing speech from human auditory cortex}.
\bjtitle{PLoS biology}
\bvolume{10}(\bissue{1}),
\bfpage{1001251}
(\byear{2012})
\end{barticle}
\endbibitem

\bibitem[\protect\citeauthoryear{Arjovsky et~al.}{2017}]{arjovsky2017wasserstein}
\begin{bchapter}
\bauthor{\bsnm{Arjovsky}, \binits{M.}},
\bauthor{\bsnm{Chintala}, \binits{S.}},
\bauthor{\bsnm{Bottou}, \binits{L.}}:
\bctitle{Wasserstein generative adversarial networks}.
In: \bbtitle{International Conference on Machine Learning},
pp. \bfpage{214}--\blpage{223}
(\byear{2017}).
\bcomment{PMLR}
\end{bchapter}
\endbibitem

\bibitem[\protect\citeauthoryear{Yi et~al.}{2017}]{yi2017dualgan}
\begin{bchapter}
\bauthor{\bsnm{Yi}, \binits{Z.}},
\bauthor{\bsnm{Zhang}, \binits{H.}},
\bauthor{\bsnm{Tan}, \binits{P.}},
\bauthor{\bsnm{Gong}, \binits{M.}}:
\bctitle{Dualgan: Unsupervised dual learning for image-to-image translation}.
In: \bbtitle{Proceedings of the IEEE International Conference on Computer Vision},
pp. \bfpage{2849}--\blpage{2857}
(\byear{2017})
\end{bchapter}
\endbibitem

\bibitem[\protect\citeauthoryear{Lewis et~al.}{2019}]{lewis2019bart}
\begin{botherref}
\oauthor{\bsnm{Lewis}, \binits{M.}},
\oauthor{\bsnm{Liu}, \binits{Y.}},
\oauthor{\bsnm{Goyal}, \binits{N.}},
\oauthor{\bsnm{Ghazvininejad}, \binits{M.}},
\oauthor{\bsnm{Mohamed}, \binits{A.}},
\oauthor{\bsnm{Levy}, \binits{O.}},
\oauthor{\bsnm{Stoyanov}, \binits{V.}},
\oauthor{\bsnm{Zettlemoyer}, \binits{L.}}:
Bart: Denoising sequence-to-sequence pre-training for natural language generation, translation, and comprehension.
arXiv preprint arXiv:1910.13461
(2019)
\end{botherref}
\endbibitem

\bibitem[\protect\citeauthoryear{Van Den~Oord et~al.}{2017}]{van2017neural}
\begin{botherref}
\oauthor{\bsnm{Van Den~Oord}, \binits{A.}},
\oauthor{\bsnm{Vinyals}, \binits{O.}}, et al.:
Neural discrete representation learning.
Advances in neural information processing systems
\textbf{30}
(2017)
\end{botherref}
\endbibitem

\bibitem[\protect\citeauthoryear{Jo et~al.}{2024}]{jo2024eegtotext}
\begin{botherref}
\oauthor{\bsnm{Jo}, \binits{H.}},
\oauthor{\bsnm{Yang}, \binits{Y.}},
\oauthor{\bsnm{Han}, \binits{J.}},
\oauthor{\bsnm{Duan}, \binits{Y.}},
\oauthor{\bsnm{Xiong}, \binits{H.}},
\oauthor{\bsnm{Lee}, \binits{W.H.}}:
Are EEG-to-Text Models Working?
(2024)
\end{botherref}
\endbibitem

\bibitem[\protect\citeauthoryear{Radford et~al.}{2023}]{radford2023robust}
\begin{bchapter}
\bauthor{\bsnm{Radford}, \binits{A.}},
\bauthor{\bsnm{Kim}, \binits{J.W.}},
\bauthor{\bsnm{Xu}, \binits{T.}},
\bauthor{\bsnm{Brockman}, \binits{G.}},
\bauthor{\bsnm{McLeavey}, \binits{C.}},
\bauthor{\bsnm{Sutskever}, \binits{I.}}:
\bctitle{Robust speech recognition via large-scale weak supervision}.
In: \bbtitle{International Conference on Machine Learning},
pp. \bfpage{28492}--\blpage{28518}
(\byear{2023}).
\bcomment{PMLR}
\end{bchapter}
\endbibitem

\bibitem[\protect\citeauthoryear{Hollenstein et~al.}{2018}]{hollenstein2018zuco}
\begin{barticle}
\bauthor{\bsnm{Hollenstein}, \binits{N.}},
\bauthor{\bsnm{Rotsztejn}, \binits{J.}},
\bauthor{\bsnm{Troendle}, \binits{M.}},
\bauthor{\bsnm{Pedroni}, \binits{A.}},
\bauthor{\bsnm{Zhang}, \binits{C.}},
\bauthor{\bsnm{Langer}, \binits{N.}}:
\batitle{Zuco, a simultaneous eeg and eye-tracking resource for natural sentence reading}.
\bjtitle{Scientific data}
\bvolume{5}(\bissue{1}),
\bfpage{1}--\blpage{13}
(\byear{2018})
\end{barticle}
\endbibitem

\bibitem[\protect\citeauthoryear{Hollenstein et~al.}{2019}]{hollenstein2019zuco}
\begin{botherref}
\oauthor{\bsnm{Hollenstein}, \binits{N.}},
\oauthor{\bsnm{Troendle}, \binits{M.}},
\oauthor{\bsnm{Zhang}, \binits{C.}},
\oauthor{\bsnm{Langer}, \binits{N.}}:
Zuco 2.0: A dataset of physiological recordings during natural reading and annotation.
arXiv preprint arXiv:1912.00903
(2019)
\end{botherref}
\endbibitem

\bibitem[\protect\citeauthoryear{Nastase et~al.}{2021}]{nastase2021narratives}
\begin{barticle}
\bauthor{\bsnm{Nastase}, \binits{S.A.}},
\bauthor{\bsnm{Liu}, \binits{Y.-F.}},
\bauthor{\bsnm{Hillman}, \binits{H.}},
\bauthor{\bsnm{Zadbood}, \binits{A.}},
\bauthor{\bsnm{Hasenfratz}, \binits{L.}},
\bauthor{\bsnm{Keshavarzian}, \binits{N.}},
\bauthor{\bsnm{Chen}, \binits{J.}},
\bauthor{\bsnm{Honey}, \binits{C.J.}},
\bauthor{\bsnm{Yeshurun}, \binits{Y.}},
\bauthor{\bsnm{Regev}, \binits{M.}}, \betal:
\batitle{The “narratives” fmri dataset for evaluating models of naturalistic language comprehension}.
\bjtitle{Scientific data}
\bvolume{8}(\bissue{1}),
\bfpage{250}
(\byear{2021})
\end{barticle}
\endbibitem

\bibitem[\protect\citeauthoryear{Gwilliams et~al.}{2023}]{gwilliams2023introducing}
\begin{barticle}
\bauthor{\bsnm{Gwilliams}, \binits{L.}},
\bauthor{\bsnm{Flick}, \binits{G.}},
\bauthor{\bsnm{Marantz}, \binits{A.}},
\bauthor{\bsnm{Pylkk{\"a}nen}, \binits{L.}},
\bauthor{\bsnm{Poeppel}, \binits{D.}},
\bauthor{\bsnm{King}, \binits{J.-R.}}:
\batitle{Introducing meg-masc a high-quality magneto-encephalography dataset for evaluating natural speech processing}.
\bjtitle{Scientific data}
\bvolume{10}(\bissue{1}),
\bfpage{862}
(\byear{2023})
\end{barticle}
\endbibitem

\bibitem[\protect\citeauthoryear{Schoffelen et~al.}{2019}]{schoffelen2019204}
\begin{barticle}
\bauthor{\bsnm{Schoffelen}, \binits{J.-M.}},
\bauthor{\bsnm{Oostenveld}, \binits{R.}},
\bauthor{\bsnm{Lam}, \binits{N.H.}},
\bauthor{\bsnm{Udd{\'e}n}, \binits{J.}},
\bauthor{\bsnm{Hult{\'e}n}, \binits{A.}},
\bauthor{\bsnm{Hagoort}, \binits{P.}}:
\batitle{A 204-subject multimodal neuroimaging dataset to study language processing}.
\bjtitle{Scientific data}
\bvolume{6}(\bissue{1}),
\bfpage{17}
(\byear{2019})
\end{barticle}
\endbibitem

\bibitem[\protect\citeauthoryear{LeBel et~al.}{2023}]{lebel2023natural}
\begin{barticle}
\bauthor{\bsnm{LeBel}, \binits{A.}},
\bauthor{\bsnm{Wagner}, \binits{L.}},
\bauthor{\bsnm{Jain}, \binits{S.}},
\bauthor{\bsnm{Adhikari-Desai}, \binits{A.}},
\bauthor{\bsnm{Gupta}, \binits{B.}},
\bauthor{\bsnm{Morgenthal}, \binits{A.}},
\bauthor{\bsnm{Tang}, \binits{J.}},
\bauthor{\bsnm{Xu}, \binits{L.}},
\bauthor{\bsnm{Huth}, \binits{A.G.}}:
\batitle{A natural language fmri dataset for voxelwise encoding models}.
\bjtitle{Scientific Data}
\bvolume{10}(\bissue{1}),
\bfpage{555}
(\byear{2023})
\end{barticle}
\endbibitem

\bibitem[\protect\citeauthoryear{Mikolov et~al.}{2013}]{mikolov2013efficient}
\begin{botherref}
\oauthor{\bsnm{Mikolov}, \binits{T.}},
\oauthor{\bsnm{Chen}, \binits{K.}},
\oauthor{\bsnm{Corrado}, \binits{G.}},
\oauthor{\bsnm{Dean}, \binits{J.}}:
Efficient estimation of word representations in vector space.
arXiv preprint arXiv:1301.3781
(2013)
\end{botherref}
\endbibitem

\bibitem[\protect\citeauthoryear{Metzger et~al.}{2022}]{metzger2022generalizable}
\begin{barticle}
\bauthor{\bsnm{Metzger}, \binits{S.L.}},
\bauthor{\bsnm{Liu}, \binits{J.R.}},
\bauthor{\bsnm{Moses}, \binits{D.A.}},
\bauthor{\bsnm{Dougherty}, \binits{M.E.}},
\bauthor{\bsnm{Seaton}, \binits{M.P.}},
\bauthor{\bsnm{Littlejohn}, \binits{K.T.}},
\bauthor{\bsnm{Chartier}, \binits{J.}},
\bauthor{\bsnm{Anumanchipalli}, \binits{G.K.}},
\bauthor{\bsnm{Tu-Chan}, \binits{A.}},
\bauthor{\bsnm{Ganguly}, \binits{K.}}, \betal:
\batitle{Generalizable spelling using a speech neuroprosthesis in an individual with severe limb and vocal paralysis}.
\bjtitle{Nature communications}
\bvolume{13}(\bissue{1}),
\bfpage{6510}
(\byear{2022})
\end{barticle}
\endbibitem

\bibitem[\protect\citeauthoryear{Leuthardt et~al.}{2011}]{leuthardt2011using}
\begin{barticle}
\bauthor{\bsnm{Leuthardt}, \binits{E.C.}},
\bauthor{\bsnm{Gaona}, \binits{C.}},
\bauthor{\bsnm{Sharma}, \binits{M.}},
\bauthor{\bsnm{Szrama}, \binits{N.}},
\bauthor{\bsnm{Roland}, \binits{J.}},
\bauthor{\bsnm{Freudenberg}, \binits{Z.}},
\bauthor{\bsnm{Solis}, \binits{J.}},
\bauthor{\bsnm{Breshears}, \binits{J.}},
\bauthor{\bsnm{Schalk}, \binits{G.}}:
\batitle{Using the electrocorticographic speech network to control a brain--computer interface in humans}.
\bjtitle{Journal of neural engineering}
\bvolume{8}(\bissue{3}),
\bfpage{036004}
(\byear{2011})
\end{barticle}
\endbibitem

\bibitem[\protect\citeauthoryear{Willett et~al.}{2021}]{willett2021high}
\begin{barticle}
\bauthor{\bsnm{Willett}, \binits{F.R.}},
\bauthor{\bsnm{Avansino}, \binits{D.T.}},
\bauthor{\bsnm{Hochberg}, \binits{L.R.}},
\bauthor{\bsnm{Henderson}, \binits{J.M.}},
\bauthor{\bsnm{Shenoy}, \binits{K.V.}}:
\batitle{High-performance brain-to-text communication via handwriting}.
\bjtitle{Nature}
\bvolume{593}(\bissue{7858}),
\bfpage{249}--\blpage{254}
(\byear{2021})
\end{barticle}
\endbibitem

\bibitem[\protect\citeauthoryear{Brigham and Kumar}{2010}]{brigham2010imagined}
\begin{bchapter}
\bauthor{\bsnm{Brigham}, \binits{K.}},
\bauthor{\bsnm{Kumar}, \binits{B.V.}}:
\bctitle{Imagined speech classification with eeg signals for silent communication: a preliminary investigation into synthetic telepathy}.
In: \bbtitle{2010 4th International Conference on Bioinformatics and Biomedical Engineering},
pp. \bfpage{1}--\blpage{4}
(\byear{2010}).
\bcomment{IEEE}
\end{bchapter}
\endbibitem

\bibitem[\protect\citeauthoryear{Duraivel et~al.}{2023}]{duraivel2023high}
\begin{barticle}
\bauthor{\bsnm{Duraivel}, \binits{S.}},
\bauthor{\bsnm{Rahimpour}, \binits{S.}},
\bauthor{\bsnm{Chiang}, \binits{C.-H.}},
\bauthor{\bsnm{Trumpis}, \binits{M.}},
\bauthor{\bsnm{Wang}, \binits{C.}},
\bauthor{\bsnm{Barth}, \binits{K.}},
\bauthor{\bsnm{Harward}, \binits{S.C.}},
\bauthor{\bsnm{Lad}, \binits{S.P.}},
\bauthor{\bsnm{Friedman}, \binits{A.H.}},
\bauthor{\bsnm{Southwell}, \binits{D.G.}}, \betal:
\batitle{High-resolution neural recordings improve the accuracy of speech decoding}.
\bjtitle{Nature communications}
\bvolume{14}(\bissue{1}),
\bfpage{6938}
(\byear{2023})
\end{barticle}
\endbibitem

\bibitem[\protect\citeauthoryear{Wandelt et~al.}{2024}]{Wandelt2024RepresentationOI}
\begin{botherref}
\oauthor{\bsnm{Wandelt}, \binits{S.K.}},
\oauthor{\bsnm{Bj{\aa}nes}, \binits{D.A.}},
\oauthor{\bsnm{Pejsa}, \binits{K.}},
\oauthor{\bsnm{Lee}, \binits{B.}},
\oauthor{\bsnm{Liu}, \binits{C.}},
\oauthor{\bsnm{Andersen}, \binits{R.A.}}:
Representation of internal speech by single neurons in human supramarginal gyrus.
Nature human behaviour
(2024)
\end{botherref}
\endbibitem

\bibitem[\protect\citeauthoryear{Chang et~al.}{2010}]{chang2010categorical}
\begin{barticle}
\bauthor{\bsnm{Chang}, \binits{E.F.}},
\bauthor{\bsnm{Rieger}, \binits{J.W.}},
\bauthor{\bsnm{Johnson}, \binits{K.}},
\bauthor{\bsnm{Berger}, \binits{M.S.}},
\bauthor{\bsnm{Barbaro}, \binits{N.M.}},
\bauthor{\bsnm{Knight}, \binits{R.T.}}:
\batitle{Categorical speech representation in human superior temporal gyrus}.
\bjtitle{Nature neuroscience}
\bvolume{13}(\bissue{11}),
\bfpage{1428}--\blpage{1432}
(\byear{2010})
\end{barticle}
\endbibitem

\bibitem[\protect\citeauthoryear{Mesgarani et~al.}{2014}]{mesgarani2014phonetic}
\begin{barticle}
\bauthor{\bsnm{Mesgarani}, \binits{N.}},
\bauthor{\bsnm{Cheung}, \binits{C.}},
\bauthor{\bsnm{Johnson}, \binits{K.}},
\bauthor{\bsnm{Chang}, \binits{E.F.}}:
\batitle{Phonetic feature encoding in human superior temporal gyrus}.
\bjtitle{Science}
\bvolume{343}(\bissue{6174}),
\bfpage{1006}--\blpage{1010}
(\byear{2014})
\end{barticle}
\endbibitem

\bibitem[\protect\citeauthoryear{Clayton et~al.}{2020}]{clayton2020decoding}
\begin{bchapter}
\bauthor{\bsnm{Clayton}, \binits{J.}},
\bauthor{\bsnm{Wellington}, \binits{S.}},
\bauthor{\bsnm{Valentini-Botinhao}, \binits{C.}},
\bauthor{\bsnm{Watts}, \binits{O.}}:
\bctitle{Decoding imagined, heard, and spoken speech: Classification and regression of eeg using a 14-channel dry-contact mobile headset.}
In: \bbtitle{INTERSPEECH},
pp. \bfpage{4886}--\blpage{4890}
(\byear{2020})
\end{bchapter}
\endbibitem

\bibitem[\protect\citeauthoryear{Ball et~al.}{2009}]{ball2009signal}
\begin{barticle}
\bauthor{\bsnm{Ball}, \binits{T.}},
\bauthor{\bsnm{Kern}, \binits{M.}},
\bauthor{\bsnm{Mutschler}, \binits{I.}},
\bauthor{\bsnm{Aertsen}, \binits{A.}},
\bauthor{\bsnm{Schulze-Bonhage}, \binits{A.}}:
\batitle{Signal quality of simultaneously recorded invasive and non-invasive eeg}.
\bjtitle{Neuroimage}
\bvolume{46}(\bissue{3}),
\bfpage{708}--\blpage{716}
(\byear{2009})
\end{barticle}
\endbibitem

\bibitem[\protect\citeauthoryear{Moses et~al.}{2021}]{moses2021neuroprosthesis}
\begin{barticle}
\bauthor{\bsnm{Moses}, \binits{D.A.}},
\bauthor{\bsnm{Metzger}, \binits{S.L.}},
\bauthor{\bsnm{Liu}, \binits{J.R.}},
\bauthor{\bsnm{Anumanchipalli}, \binits{G.K.}},
\bauthor{\bsnm{Makin}, \binits{J.G.}},
\bauthor{\bsnm{Sun}, \binits{P.F.}},
\bauthor{\bsnm{Chartier}, \binits{J.}},
\bauthor{\bsnm{Dougherty}, \binits{M.E.}},
\bauthor{\bsnm{Liu}, \binits{P.M.}},
\bauthor{\bsnm{Abrams}, \binits{G.M.}}, \betal:
\batitle{Neuroprosthesis for decoding speech in a paralyzed person with anarthria}.
\bjtitle{New England Journal of Medicine}
\bvolume{385}(\bissue{3}),
\bfpage{217}--\blpage{227}
(\byear{2021})
\end{barticle}
\endbibitem

\bibitem[\protect\citeauthoryear{Povey et~al.}{2011}]{povey2011kaldi}
\begin{bchapter}
\bauthor{\bsnm{Povey}, \binits{D.}},
\bauthor{\bsnm{Ghoshal}, \binits{A.}},
\bauthor{\bsnm{Boulianne}, \binits{G.}},
\bauthor{\bsnm{Burget}, \binits{L.}},
\bauthor{\bsnm{Glembek}, \binits{O.}},
\bauthor{\bsnm{Goel}, \binits{N.}},
\bauthor{\bsnm{Hannemann}, \binits{M.}},
\bauthor{\bsnm{Motlicek}, \binits{P.}},
\bauthor{\bsnm{Qian}, \binits{Y.}},
\bauthor{\bsnm{Schwarz}, \binits{P.}}, \betal:
\bctitle{The kaldi speech recognition toolkit}.
In: \bbtitle{IEEE 2011 Workshop on Automatic Speech Recognition and Understanding}
(\byear{2011}).
\bcomment{IEEE Signal Processing Society}
\end{bchapter}
\endbibitem

\bibitem[\protect\citeauthoryear{Munteanu et~al.}{2006}]{munteanu2006measuring}
\begin{bchapter}
\bauthor{\bsnm{Munteanu}, \binits{C.}},
\bauthor{\bsnm{Penn}, \binits{G.}},
\bauthor{\bsnm{Baecker}, \binits{R.}},
\bauthor{\bsnm{Toms}, \binits{E.}},
\bauthor{\bsnm{James}, \binits{D.}}:
\bctitle{Measuring the acceptable word error rate of machine-generated webcast transcripts}.
In: \bbtitle{Ninth International Conference on Spoken Language Processing}
(\byear{2006}).
\bcomment{Citeseer}
\end{bchapter}
\endbibitem

\bibitem[\protect\citeauthoryear{Sun et~al.}{2020}]{sun2020brain2char}
\begin{barticle}
\bauthor{\bsnm{Sun}, \binits{P.}},
\bauthor{\bsnm{Anumanchipalli}, \binits{G.K.}},
\bauthor{\bsnm{Chang}, \binits{E.F.}}:
\batitle{Brain2char: a deep architecture for decoding text from brain recordings}.
\bjtitle{Journal of neural engineering}
\bvolume{17}(\bissue{6}),
\bfpage{066015}
(\byear{2020})
\end{barticle}
\endbibitem

\bibitem[\protect\citeauthoryear{Feng et~al.}{2023}]{feng2023high}
\begin{botherref}
\oauthor{\bsnm{Feng}, \binits{C.}},
\oauthor{\bsnm{Cao}, \binits{L.}},
\oauthor{\bsnm{Wu}, \binits{D.}},
\oauthor{\bsnm{Zhang}, \binits{E.}},
\oauthor{\bsnm{Wang}, \binits{T.}},
\oauthor{\bsnm{Jiang}, \binits{X.}},
\oauthor{\bsnm{Ding}, \binits{H.}},
\oauthor{\bsnm{Zhou}, \binits{C.}},
\oauthor{\bsnm{Chen}, \binits{J.}},
\oauthor{\bsnm{Wu}, \binits{H.}}, et al.:
A high-performance brain-to-sentence decoder for logosyllabic language
(2023)
\end{botherref}
\endbibitem

\bibitem[\protect\citeauthoryear{Silva et~al.}{2024}]{silva2024bilingual}
\begin{botherref}
\oauthor{\bsnm{Silva}, \binits{A.B.}},
\oauthor{\bsnm{Liu}, \binits{J.R.}},
\oauthor{\bsnm{Metzger}, \binits{S.L.}},
\oauthor{\bsnm{Bhaya-Grossman}, \binits{I.}},
\oauthor{\bsnm{Dougherty}, \binits{M.E.}},
\oauthor{\bsnm{Seaton}, \binits{M.P.}},
\oauthor{\bsnm{Littlejohn}, \binits{K.T.}},
\oauthor{\bsnm{Tu-Chan}, \binits{A.}},
\oauthor{\bsnm{Ganguly}, \binits{K.}},
\oauthor{\bsnm{Moses}, \binits{D.A.}}, et al.:
A bilingual speech neuroprosthesis driven by cortical articulatory representations shared between languages.
Nature Biomedical Engineering,
1--15
(2024)
\end{botherref}
\endbibitem

\bibitem[\protect\citeauthoryear{Schneider et~al.}{2019}]{schneider2019wav2vec}
\begin{botherref}
\oauthor{\bsnm{Schneider}, \binits{S.}},
\oauthor{\bsnm{Baevski}, \binits{A.}},
\oauthor{\bsnm{Collobert}, \binits{R.}},
\oauthor{\bsnm{Auli}, \binits{M.}}:
wav2vec: Unsupervised pre-training for speech recognition.
arXiv preprint arXiv:1904.05862
(2019)
\end{botherref}
\endbibitem

\bibitem[\protect\citeauthoryear{Radford et~al.}{2019}]{radford2019language}
\begin{barticle}
\bauthor{\bsnm{Radford}, \binits{A.}},
\bauthor{\bsnm{Wu}, \binits{J.}},
\bauthor{\bsnm{Child}, \binits{R.}},
\bauthor{\bsnm{Luan}, \binits{D.}},
\bauthor{\bsnm{Amodei}, \binits{D.}},
\bauthor{\bsnm{Sutskever}, \binits{I.}}, \betal:
\batitle{Language models are unsupervised multitask learners}.
\bjtitle{OpenAI blog}
\bvolume{1}(\bissue{8}),
\bfpage{9}
(\byear{2019})
\end{barticle}
\endbibitem

\bibitem[\protect\citeauthoryear{Zhang et~al.}{2022}]{zhang2022opt}
\begin{botherref}
\oauthor{\bsnm{Zhang}, \binits{S.}},
\oauthor{\bsnm{Roller}, \binits{S.}},
\oauthor{\bsnm{Goyal}, \binits{N.}},
\oauthor{\bsnm{Artetxe}, \binits{M.}},
\oauthor{\bsnm{Chen}, \binits{M.}},
\oauthor{\bsnm{Chen}, \binits{S.}},
\oauthor{\bsnm{Dewan}, \binits{C.}},
\oauthor{\bsnm{Diab}, \binits{M.}},
\oauthor{\bsnm{Li}, \binits{X.}},
\oauthor{\bsnm{Lin}, \binits{X.V.}}, et al.:
Opt: Open pre-trained transformer language models.
arXiv preprint arXiv:2205.01068
(2022)
\end{botherref}
\endbibitem

\bibitem[\protect\citeauthoryear{Touvron et~al.}{2023}]{touvron2023llama}
\begin{botherref}
\oauthor{\bsnm{Touvron}, \binits{H.}},
\oauthor{\bsnm{Martin}, \binits{L.}},
\oauthor{\bsnm{Stone}, \binits{K.}},
\oauthor{\bsnm{Albert}, \binits{P.}},
\oauthor{\bsnm{Almahairi}, \binits{A.}},
\oauthor{\bsnm{Babaei}, \binits{Y.}},
\oauthor{\bsnm{Bashlykov}, \binits{N.}},
\oauthor{\bsnm{Batra}, \binits{S.}},
\oauthor{\bsnm{Bhargava}, \binits{P.}},
\oauthor{\bsnm{Bhosale}, \binits{S.}}, et al.:
Llama 2: Open foundation and fine-tuned chat models.
arXiv preprint arXiv:2307.09288
(2023)
\end{botherref}
\endbibitem

\bibitem[\protect\citeauthoryear{Anumanchipalli et~al.}{2019}]{anumanchipalli2019speech}
\begin{barticle}
\bauthor{\bsnm{Anumanchipalli}, \binits{G.K.}},
\bauthor{\bsnm{Chartier}, \binits{J.}},
\bauthor{\bsnm{Chang}, \binits{E.F.}}:
\batitle{Speech synthesis from neural decoding of spoken sentences}.
\bjtitle{Nature}
\bvolume{568}(\bissue{7753}),
\bfpage{493}--\blpage{498}
(\byear{2019})
\end{barticle}
\endbibitem

\bibitem[\protect\citeauthoryear{Krishna et~al.}{2020}]{krishna2020speech}
\begin{bchapter}
\bauthor{\bsnm{Krishna}, \binits{G.}},
\bauthor{\bsnm{Tran}, \binits{C.}},
\bauthor{\bsnm{Han}, \binits{Y.}},
\bauthor{\bsnm{Carnahan}, \binits{M.}},
\bauthor{\bsnm{Tewfik}, \binits{A.H.}}:
\bctitle{Speech synthesis using eeg}.
In: \bbtitle{ICASSP 2020-2020 IEEE International Conference on Acoustics, Speech and Signal Processing (ICASSP)},
pp. \bfpage{1235}--\blpage{1238}
(\byear{2020}).
\bcomment{IEEE}
\end{bchapter}
\endbibitem

\bibitem[\protect\citeauthoryear{Huang et~al.}{2017}]{huang2017densely}
\begin{bchapter}
\bauthor{\bsnm{Huang}, \binits{G.}},
\bauthor{\bsnm{Liu}, \binits{Z.}},
\bauthor{\bsnm{Van Der~Maaten}, \binits{L.}},
\bauthor{\bsnm{Weinberger}, \binits{K.Q.}}:
\bctitle{Densely connected convolutional networks}.
In: \bbtitle{Proceedings of the IEEE Conference on Computer Vision and Pattern Recognition},
pp. \bfpage{4700}--\blpage{4708}
(\byear{2017})
\end{bchapter}
\endbibitem

\bibitem[\protect\citeauthoryear{Bocquelet et~al.}{2016}]{bocquelet2016real}
\begin{barticle}
\bauthor{\bsnm{Bocquelet}, \binits{F.}},
\bauthor{\bsnm{Hueber}, \binits{T.}},
\bauthor{\bsnm{Girin}, \binits{L.}},
\bauthor{\bsnm{Savariaux}, \binits{C.}},
\bauthor{\bsnm{Yvert}, \binits{B.}}:
\batitle{Real-time control of an articulatory-based speech synthesizer for brain computer interfaces}.
\bjtitle{PLoS computational biology}
\bvolume{12}(\bissue{11}),
\bfpage{1005119}
(\byear{2016})
\end{barticle}
\endbibitem

\bibitem[\protect\citeauthoryear{Cheung et~al.}{2016}]{cheung2016auditory}
\begin{barticle}
\bauthor{\bsnm{Cheung}, \binits{C.}},
\bauthor{\bsnm{Hamilton}, \binits{L.S.}},
\bauthor{\bsnm{Johnson}, \binits{K.}},
\bauthor{\bsnm{Chang}, \binits{E.F.}}:
\batitle{The auditory representation of speech sounds in human motor cortex}.
\bjtitle{elife}
\bvolume{5},
\bfpage{12577}
(\byear{2016})
\end{barticle}
\endbibitem

\bibitem[\protect\citeauthoryear{Dichter et~al.}{2018}]{dichter2018control}
\begin{barticle}
\bauthor{\bsnm{Dichter}, \binits{B.K.}},
\bauthor{\bsnm{Breshears}, \binits{J.D.}},
\bauthor{\bsnm{Leonard}, \binits{M.K.}},
\bauthor{\bsnm{Chang}, \binits{E.F.}}:
\batitle{The control of vocal pitch in human laryngeal motor cortex}.
\bjtitle{Cell}
\bvolume{174}(\bissue{1}),
\bfpage{21}--\blpage{31}
(\byear{2018})
\end{barticle}
\endbibitem

\bibitem[\protect\citeauthoryear{Chartier et~al.}{2018}]{chartier2018encoding}
\begin{barticle}
\bauthor{\bsnm{Chartier}, \binits{J.}},
\bauthor{\bsnm{Anumanchipalli}, \binits{G.K.}},
\bauthor{\bsnm{Johnson}, \binits{K.}},
\bauthor{\bsnm{Chang}, \binits{E.F.}}:
\batitle{Encoding of articulatory kinematic trajectories in human speech sensorimotor cortex}.
\bjtitle{Neuron}
\bvolume{98}(\bissue{5}),
\bfpage{1042}--\blpage{1054}
(\byear{2018})
\end{barticle}
\endbibitem

\bibitem[\protect\citeauthoryear{Mugler et~al.}{2018}]{mugler2018differential}
\begin{barticle}
\bauthor{\bsnm{Mugler}, \binits{E.M.}},
\bauthor{\bsnm{Tate}, \binits{M.C.}},
\bauthor{\bsnm{Livescu}, \binits{K.}},
\bauthor{\bsnm{Templer}, \binits{J.W.}},
\bauthor{\bsnm{Goldrick}, \binits{M.A.}},
\bauthor{\bsnm{Slutzky}, \binits{M.W.}}:
\batitle{Differential representation of articulatory gestures and phonemes in precentral and inferior frontal gyri}.
\bjtitle{Journal of Neuroscience}
\bvolume{38}(\bissue{46}),
\bfpage{9803}--\blpage{9813}
(\byear{2018})
\end{barticle}
\endbibitem

\bibitem[\protect\citeauthoryear{Angrick et~al.}{2021}]{angrick2021real}
\begin{barticle}
\bauthor{\bsnm{Angrick}, \binits{M.}},
\bauthor{\bsnm{Ottenhoff}, \binits{M.C.}},
\bauthor{\bsnm{Diener}, \binits{L.}},
\bauthor{\bsnm{Ivucic}, \binits{D.}},
\bauthor{\bsnm{Ivucic}, \binits{G.}},
\bauthor{\bsnm{Goulis}, \binits{S.}},
\bauthor{\bsnm{Saal}, \binits{J.}},
\bauthor{\bsnm{Colon}, \binits{A.J.}},
\bauthor{\bsnm{Wagner}, \binits{L.}},
\bauthor{\bsnm{Krusienski}, \binits{D.J.}}, \betal:
\batitle{Real-time synthesis of imagined speech processes from minimally invasive recordings of neural activity}.
\bjtitle{Communications biology}
\bvolume{4}(\bissue{1}),
\bfpage{1055}
(\byear{2021})
\end{barticle}
\endbibitem

\bibitem[\protect\citeauthoryear{Liu et~al.}{2023}]{liu2023decoding}
\begin{barticle}
\bauthor{\bsnm{Liu}, \binits{Y.}},
\bauthor{\bsnm{Zhao}, \binits{Z.}},
\bauthor{\bsnm{Xu}, \binits{M.}},
\bauthor{\bsnm{Yu}, \binits{H.}},
\bauthor{\bsnm{Zhu}, \binits{Y.}},
\bauthor{\bsnm{Zhang}, \binits{J.}},
\bauthor{\bsnm{Bu}, \binits{L.}},
\bauthor{\bsnm{Zhang}, \binits{X.}},
\bauthor{\bsnm{Lu}, \binits{J.}},
\bauthor{\bsnm{Li}, \binits{Y.}}, \betal:
\batitle{Decoding and synthesizing tonal language speech from brain activity}.
\bjtitle{Science Advances}
\bvolume{9}(\bissue{23}),
\bfpage{0478}
(\byear{2023})
\end{barticle}
\endbibitem

\bibitem[\protect\citeauthoryear{Sun et~al.}{2023}]{sun2023vividtalk}
\begin{botherref}
\oauthor{\bsnm{Sun}, \binits{X.}},
\oauthor{\bsnm{Zhang}, \binits{L.}},
\oauthor{\bsnm{Zhu}, \binits{H.}},
\oauthor{\bsnm{Zhang}, \binits{P.}},
\oauthor{\bsnm{Zhang}, \binits{B.}},
\oauthor{\bsnm{Ji}, \binits{X.}},
\oauthor{\bsnm{Zhou}, \binits{K.}},
\oauthor{\bsnm{Gao}, \binits{D.}},
\oauthor{\bsnm{Bo}, \binits{L.}},
\oauthor{\bsnm{Cao}, \binits{X.}}:
Vividtalk: One-shot audio-driven talking head generation based on 3d hybrid prior.
arXiv preprint arXiv:2312.01841
(2023)
\end{botherref}
\endbibitem

\bibitem[\protect\citeauthoryear{Song et~al.}{2024}]{song2024continuous}
\begin{botherref}
\oauthor{\bsnm{Song}, \binits{H.}},
\oauthor{\bsnm{Hsieh}, \binits{T.-H.}},
\oauthor{\bsnm{Yeon}, \binits{S.H.}},
\oauthor{\bsnm{Shu}, \binits{T.}},
\oauthor{\bsnm{Nawrot}, \binits{M.}},
\oauthor{\bsnm{Landis}, \binits{C.F.}},
\oauthor{\bsnm{Friedman}, \binits{G.N.}},
\oauthor{\bsnm{Israel}, \binits{E.A.}},
\oauthor{\bsnm{Gutierrez-Arango}, \binits{S.}},
\oauthor{\bsnm{Carty}, \binits{M.J.}}, et al.:
Continuous neural control of a bionic limb restores biomimetic gait after amputation.
Nature Medicine,
1--10
(2024)
\end{botherref}
\endbibitem

\bibitem[\protect\citeauthoryear{Wang et~al.}{2024}]{wang2024neural}
\begin{botherref}
\oauthor{\bsnm{Wang}, \binits{J.}},
\oauthor{\bsnm{Bi}, \binits{L.}},
\oauthor{\bsnm{Fei}, \binits{W.}},
\oauthor{\bsnm{Xu}, \binits{X.}},
\oauthor{\bsnm{Liu}, \binits{A.}},
\oauthor{\bsnm{Mo}, \binits{L.}},
\oauthor{\bsnm{Feleke}, \binits{A.G.}}:
Neural correlate and movement decoding of simultaneous-and-sequential bimanual movements using eeg signals.
IEEE Transactions on Neural Systems and Rehabilitation Engineering
(2024)
\end{botherref}
\endbibitem

\bibitem[\protect\citeauthoryear{Luo et~al.}{2023}]{Luo2023StableDF}
\begin{botherref}
\oauthor{\bsnm{Luo}, \binits{S.}},
\oauthor{\bsnm{Angrick}, \binits{M.}},
\oauthor{\bsnm{Coogan}, \binits{C.}},
\oauthor{\bsnm{Candrea}, \binits{D.}},
\oauthor{\bsnm{Wyse-Sookoo}, \binits{K.}},
\oauthor{\bsnm{Shah}, \binits{S.}},
\oauthor{\bsnm{Rabbani}, \binits{Q.}},
\oauthor{\bsnm{Milsap}, \binits{G.W.}},
\oauthor{\bsnm{Weiss}, \binits{A.R.}},
\oauthor{\bsnm{Anderson}, \binits{W.S.}},
\oauthor{\bsnm{Tippett}, \binits{D.C.}},
\oauthor{\bsnm{Maragakis}, \binits{N.J.}},
\oauthor{\bsnm{Clawson}, \binits{L.}},
\oauthor{\bsnm{Vansteensel}, \binits{M.J.}},
\oauthor{\bsnm{Wester}, \binits{B.A.}},
\oauthor{\bsnm{Tenore}, \binits{F.V.}},
\oauthor{\bsnm{Hermansky}, \binits{H.}},
\oauthor{\bsnm{Fifer}, \binits{M.S.}},
\oauthor{\bsnm{Ramsey}, \binits{N.F.}},
\oauthor{\bsnm{Crone}, \binits{N.E.}}:
Stable decoding from a speech bci enables control for an individual with als without recalibration for 3 months.
Advanced Science
\textbf{10}
(2023)
\end{botherref}
\endbibitem

\bibitem[\protect\citeauthoryear{Chen et~al.}{2021}]{chen2021gigaspeech}
\begin{bchapter}
\bauthor{\bsnm{Chen}, \binits{G.}},
\bauthor{\bsnm{Chai}, \binits{S.}},
\bauthor{\bsnm{Wang}, \binits{G.}},
\bauthor{\bsnm{Du}, \binits{J.}},
\bauthor{\bsnm{Zhang}, \binits{W.Q.}},
\bauthor{\bsnm{Weng}, \binits{C.}},
\bauthor{\bsnm{Su}, \binits{D.}},
\bauthor{\bsnm{Povey}, \binits{D.}},
\bauthor{\bsnm{Trmal}, \binits{J.}},
\bauthor{\bsnm{Zhang}, \binits{J.}}, \betal:
\bctitle{Gigaspeech: An evolving, multi-domain asr corpus with 10,000 hours of transcribed audio}.
In: \bbtitle{22nd Annual Conference of the International Speech Communication Association, INTERSPEECH 2021},
pp. \bfpage{4376}--\blpage{4380}
(\byear{2021}).
\bcomment{International Speech Communication Association}
\end{bchapter}
\endbibitem

\bibitem[\protect\citeauthoryear{Tankus et~al.}{2024}]{tankus2024machine}
\begin{barticle}
\bauthor{\bsnm{Tankus}, \binits{A.}},
\bauthor{\bsnm{Rosenberg}, \binits{N.}},
\bauthor{\bsnm{Ben-Hamo}, \binits{O.}},
\bauthor{\bsnm{Stern}, \binits{E.}},
\bauthor{\bsnm{Strauss}, \binits{I.}}:
\batitle{Machine learning decoding of single neurons in the thalamus for speech brain-machine interfaces}.
\bjtitle{Journal of Neural Engineering}
\bvolume{21}(\bissue{3}),
\bfpage{036009}
(\byear{2024})
\end{barticle}
\endbibitem

\bibitem[\protect\citeauthoryear{Feinberg et~al.}{2023}]{feinberg2023next}
\begin{barticle}
\bauthor{\bsnm{Feinberg}, \binits{D.A.}},
\bauthor{\bsnm{Beckett}, \binits{A.J.}},
\bauthor{\bsnm{Vu}, \binits{A.T.}},
\bauthor{\bsnm{Stockmann}, \binits{J.}},
\bauthor{\bsnm{Huber}, \binits{L.}},
\bauthor{\bsnm{Ma}, \binits{S.}},
\bauthor{\bsnm{Ahn}, \binits{S.}},
\bauthor{\bsnm{Setsompop}, \binits{K.}},
\bauthor{\bsnm{Cao}, \binits{X.}},
\bauthor{\bsnm{Park}, \binits{S.}}, \betal:
\batitle{Next-generation mri scanner designed for ultra-high-resolution human brain imaging at 7 tesla}.
\bjtitle{Nature methods}
\bvolume{20}(\bissue{12}),
\bfpage{2048}--\blpage{2057}
(\byear{2023})
\end{barticle}
\endbibitem

\bibitem[\protect\citeauthoryear{Ikegawa et~al.}{2024}]{ikegawa2024text}
\begin{barticle}
\bauthor{\bsnm{Ikegawa}, \binits{Y.}},
\bauthor{\bsnm{Fukuma}, \binits{R.}},
\bauthor{\bsnm{Sugano}, \binits{H.}},
\bauthor{\bsnm{Oshino}, \binits{S.}},
\bauthor{\bsnm{Tani}, \binits{N.}},
\bauthor{\bsnm{Tamura}, \binits{K.}},
\bauthor{\bsnm{Iimura}, \binits{Y.}},
\bauthor{\bsnm{Suzuki}, \binits{H.}},
\bauthor{\bsnm{Yamamoto}, \binits{S.}},
\bauthor{\bsnm{Fujita}, \binits{Y.}}, \betal:
\batitle{Text and image generation from intracranial electroencephalography using an embedding space for text and images}.
\bjtitle{Journal of Neural Engineering}
\bvolume{21}(\bissue{3}),
\bfpage{036019}
(\byear{2024})
\end{barticle}
\endbibitem

\bibitem[\protect\citeauthoryear{Han et~al.}{2024}]{han2024onellm}
\begin{bchapter}
\bauthor{\bsnm{Han}, \binits{J.}},
\bauthor{\bsnm{Gong}, \binits{K.}},
\bauthor{\bsnm{Zhang}, \binits{Y.}},
\bauthor{\bsnm{Wang}, \binits{J.}},
\bauthor{\bsnm{Zhang}, \binits{K.}},
\bauthor{\bsnm{Lin}, \binits{D.}},
\bauthor{\bsnm{Qiao}, \binits{Y.}},
\bauthor{\bsnm{Gao}, \binits{P.}},
\bauthor{\bsnm{Yue}, \binits{X.}}:
\bctitle{Onellm: One framework to align all modalities with language}.
In: \bbtitle{Proceedings of the IEEE/CVF Conference on Computer Vision and Pattern Recognition},
pp. \bfpage{26584}--\blpage{26595}
(\byear{2024})
\end{bchapter}
\endbibitem

\bibitem[\protect\citeauthoryear{Tatulian}{2022}]{Tatulian2022ChallengesAH}
\begin{botherref}
\oauthor{\bsnm{Tatulian}, \binits{S.A.}}:
Challenges and hopes for alzheimer's disease.
Drug discovery today
(2022)
\end{botherref}
\endbibitem

\bibitem[\protect\citeauthoryear{Bucur and Papagno}{2022}]{Bucur2022DeepBS}
\begin{barticle}
\bauthor{\bsnm{Bucur}, \binits{M.}},
\bauthor{\bsnm{Papagno}, \binits{C.}}:
\batitle{Deep brain stimulation in parkinson disease: A meta-analysis of the long-term neuropsychological outcomes}.
\bjtitle{Neuropsychology Review}
\bvolume{33},
\bfpage{307}--\blpage{346}
(\byear{2022})
\end{barticle}
\endbibitem

\end{thebibliography}

\begin{appendices}

\section{Dataset Summary for Linguistic Brain Decoding}\label{secA1}

For the stimulus recognition paradigm, most work was conducted using self-collected brain recordings without the necessity of reuse and reproduction. As shown in Table~\ref{Dataset}, we summarize the typical datasets of sequence decoding, which mainly correspond to brain recording translation (BRT), inner speech recognition (ISR), speech stimuli reconstruction (SSR) and brain-to-speech (BTS).

\begin{table*}[htp!]
\caption{Representative datasets for linguistic sequence decoding of brain recordings. We annotate with the main author’s information for datasets without names.}
\begin{adjustbox}{center}
\centering
\label{Dataset}
\begin{tabular}{|c|c|c|c|c|c|c|c|}
\hline
Dataset & Task & Language & Type & Exp & Subjects & Vocabulary & Duration \\ \hline
ZuCo \cite{hollenstein2018zuco} & \multirow{7}{*}{BRT} & ENG & EEG & NR & 12 & - & 300  \\ \cline{1-1} \cline{3-8}
ZuCo 2.0 \cite{hollenstein2019zuco}&   & ENG & EEG & NR & 18 & - & 390 \\ \cline{1-1} \cline{3-8}
MOUS \cite{schoffelen2019204} &  & DUT & fMRI,MEG & NR,NL & 204 & - & 360\\ \cline{1-1} \cline{3-8}
Narratives \cite{nastase2021narratives}&  & ENG & fMRI & NL & 345 & - & 4.6h \\ \cline{1-1} \cline{3-8}
Tang's \cite{tang2023semantic} &  & ENG & fMRI & NL & 7 & - & 82$\times$(5-15m)\\ \cline{1-1} \cline{3-8}
MEG-MASC \cite{gwilliams2023introducing} &  & ENG & MEG & NL & 27 & - & 2h\\ \cline{1-1} \cline{3-8}
MothRadioHour \cite{lebel2023natural} &  & ENG & fMRI & NL & 8 & - & 320m\\ \hline
Herff's \cite{herff2015brain} & \multirow{8}{*}{ISR} & ENG & ECoG & VS & 7 & - & 548 phrases \\ \cline{1-1} \cline{3-8}
Sun's \cite{sun2020brain2char} &  & ENG & ECoG & VS & 4 & 1900 & - \\ \cline{1-1} \cline{3-8}
MOCHA \cite{makin2020machine} &  & ENG & ECoG & VS & 4 & 250 & 490 \\ \cline{1-1} \cline{3-8}
Moses's \cite{moses2021neuroprosthesis} &   & ENG & Invasive & VS & 1 & 50 & 50 \\\cline{1-1} \cline{3-8}
Willett's \cite{willett2023high} &  & ENG & Invasive & IS,VS & 1 & 125,000 & 10850 \\ \cline{1-1} \cline{3-8}
Metzger's \cite{metzger2023high} &  & ENG & ECoG & IS & 1 & 39,378 & 9655 \\ \cline{1-1} \cline{3-8}
Feng's \cite{feng2023high} &  & CHI & sEEG & VS & 4 & 84 & 100 \\ \cline{1-1} \cline{3-8}
Silva's \cite{silva2024bilingual} &  & ENG,SPA & ECoG & IS & 1 & $\sim$200 & - \\ \hline
Akbari's \cite{akbari2019towards} & \multirow{3}{*}{SSR} & ENG & ECoG & NL & 5 & - & 30m \\ \cline{1-1} \cline{3-8}
Guo's \cite{guo2023end} & & ENG & EEG & NL & 50 & - & 6300 \\ \cline{1-1} \cline{3-8}
Senda's \cite{senda2024auditory} & & JAN & ECoG,sEEG & NL & 5 & 
 & 96  \\ \hline
Anumanchipalli's \cite{anumanchipalli2019speech} & \multirow{3}{*}{BTS} & ENG & ECoG & VS & 5 & - & $\sim$2000 \\ \cline{1-1} \cline{3-8}
Angrick's \cite{angrick2019speech} & & ENG & ECoG & VS & 5 & - & 2000 \\ \cline{1-1} \cline{3-8}
Chen's \cite{chen2024neural} &  & ENG & ECoG & VS & 48 & 50 & - \\ \hline
\end{tabular}
\end{adjustbox}
    \begin{tablenotes}
        \item 1. Language: ENG (English), DUT (Dutch), CHI (Chinese), SPA (Spain), JAN (Japanese).
        \item 2. Type: sEEG (stereo-EEG).
        \item 3. Exp (Experiment setting): NR (natural reading), NL (natural listening), VS (vocal speech), IS (inner speech).
        \item 4. Duration: It represents the sum of sentences when contains only numbers, and the time duration ends with m (minutes) or h (hour).
        \item 5. The vocabulary refers to a specialized comprising pairs of initials and tones for Chinese.
    \end{tablenotes}
\end{table*}

\end{appendices}

\end{document}